\documentclass{ieeeaccess}
\makeatletter
\AtBeginDocument{\DeclareMathVersion{bold}
\SetSymbolFont{operators}{bold}{T1}{times}{b}{n}
\SetSymbolFont{NewLetters}{bold}{T1}{times}{b}{it}
\SetMathAlphabet{\mathrm}{bold}{T1}{times}{b}{n}
\SetMathAlphabet{\mathit}{bold}{T1}{times}{b}{it}
\SetMathAlphabet{\mathbf}{bold}{T1}{times}{b}{n}
\SetMathAlphabet{\mathtt}{bold}{OT1}{pcr}{b}{n}
\SetSymbolFont{symbols}{bold}{OMS}{cmsy}{b}{n}
\renewcommand\boldmath{\@nomath\boldmath\mathversion{bold}}}

\makeatother

\def\BibTeX{{\rm B\kern-.05em{\sc i\kern-.025em b}\kern-.08em
    T\kern-.1667em\lower.7ex\hbox{E}\kern-.125emX}}
\usepackage{cite}
\usepackage{amsmath,amssymb,amsfonts}
\usepackage{algorithmic}
\usepackage{graphicx}
\usepackage{textcomp}
\usepackage{bm}
\usepackage{lipsum}

\usepackage{longtable}
\usepackage{varwidth}
\usepackage{booktabs}
\usepackage{multirow}

\usepackage{setspace}
\usepackage{calligra}
\usepackage{mathtools}

\usepackage{amsthm}
\usepackage{tabularx}
\usepackage{rotating}
\usepackage{array}

\usepackage{comment}
\usepackage{listings}
\usepackage{enumitem}

\usepackage{blindtext}

\usepackage{acronym}

\usepackage{multicol}

\usepackage{float}

\usepackage{algorithm}
\usepackage[all=normal,paragraphs=tight, floats=tight,indent=tight]{savetrees}
\usepackage[T1]{fontenc}
\usepackage{lmodern}
\usepackage{soul}
\usepackage{framed}
\usepackage{csquotes}
\usepackage{multirow}
\usepackage{array}
\newcolumntype{L}[1]{>{\raggedright\let\newline\\\arraybackslash\hspace{0pt}}m{#1}}
\newcolumntype{C}[1]{>{\centering\let\newline\\\arraybackslash\hspace{0pt}}m{#1}}
\newcolumntype{R}[1]{>{\raggedleft\let\newline\\\arraybackslash\hspace{0pt}}m{#1}}
\newcolumntype{H}{>{\collectcell\lstinline}l<{\endcollectcell}}
\usepackage{url}
\MakeOuterQuote{"}
\usepackage[caption=false,font=footnotesize,labelformat=simple]{subfig}

\usepackage{pifont}

\newcommand{\tick}{\textcolor{green}{\ding{51}}}
\newcommand{\cross}{\textcolor{red}{\ding{55}}}
\begin{document}
\history{Date of publication xxxx 00, 0000, date of current version xxxx 00, 0000.}
\doi{10.1109/ACCESS.2024.0429000}

\title{Survey of different Large Language Model Architectures: Trends, Benchmarks, and Challenges}
\author{\uppercase{Minghao Shao}\authorrefmark{1},
\uppercase{Abdul Basit}\authorrefmark{2}, \uppercase{Ramesh Karri}\authorrefmark{1}, and  \uppercase{Muhammad Shafique}\authorrefmark{2}}

\address[1]{New York University, New York, USA}
\address[2]{New York University Abu Dhabi, Abu Dhabi, UAE}

\markboth
{Minghao \headeretal: Survey of different Large Language Model Architectures: Trends, Benchmarks, and Challenges}
{Minghao \headeretal: Survey of different Large Language Model Architectures: Trends, Benchmarks, and Challenges}

\corresp{Corresponding author: Minghao Shao (e-mail: shao.minghao@nyu.edu).}

\begin{abstract}
Large Language Models (LLMs) represent a class of deep learning models adept at understanding natural language and generating coherent responses to various prompts or queries. These models far exceed the complexity of conventional neural networks, often encompassing dozens of neural network layers and containing billions to trillions of parameters. They are typically trained on vast datasets, utilizing architectures based on transformer blocks. Present-day LLMs are multi-functional, capable of performing a range of tasks from text generation and language translation to question answering, as well as code generation and analysis. An advanced subset of these models, known as Multimodal Large Language Models (MLLMs), extends LLM  capabilities to process and interpret multiple data modalities, including images, audio, and video. This enhancement empowers MLLMs with capabilities like video editing, image comprehension, and captioning for visual content. This survey provides a comprehensive overview of the recent advancements in LLMs. We begin by tracing the evolution of LLMs and subsequently delve into the advent and nuances of MLLMs. We analyze emerging state-of-the-art MLLMs, exploring their technical features, strengths, and limitations. Additionally, we present a comparative analysis of these models and discuss their challenges, potential limitations, and prospects for future development.
\end{abstract}


\titlepgskip=-21pt

\maketitle

\vspace{-10pt}
\section{Introduction\label{sec:intro}}

The introduction of the Transformer architecture \cite{vaswani2017attention} in 2017 marked an inflection point in the trajectory of Natural Language Processing (NLP) technology. One notable derivative of this innovation is Large Language Models (LLMs). Demonstrating proficiency across multiple NLP tasks, LLMs have been integral for text generation, machine translation, and natural language understanding. Their evolution, spanning several years, has not only underlined their power in linguistic tasks but also showcased their versatility in handling diverse formats like images, videos, and robotic interfaces.

Notable tasks for which LLMs have been used include:
\begin{itemize}
\item \textbf{Text generation} of coherent text from structured input upon receiving pertinent instructions.
\item \textbf{Logical reasoning}: Analysis and inference based on logic intrinsic to a given scenario.
\item \textbf{Machine Translation} across linguistic frameworks.
\item \textbf{Summarization}: Contextual abridgment of content.
\item \textbf{Multimodal support}: Beyond textual content, LLMs also facilitate inputs and outputs in various formats, including images, videos, and interactions in robotic environments, leveraging multiple modalities.
\end{itemize}

The genesis of LLM development can be traced back to 2018 with the advent of GPT \cite{radford2018improving} and BERT \cite{devlin2018bert}. Each model, crafted with distinct architectures, catered to specific niches within the LLM spectrum. Contemporary LLMs primarily leverage the foundational Transformer architecture and can be grouped as:

\begin{itemize}
\item \textbf{Auto-encoding}: Primarily encoder-based, these models are tailored for contextual NLP tasks. E.g., BERT and its derivatives.
\item \textbf{Auto-regressive}: Decoder-centric, these models are optimized for generative tasks. E.g., GPT series.
\item \textbf{Encoder-Decoder}: Amalgamating both encoder and decoder structures, these LLMs harness the strengths of the preceding two types, albeit with certain trade-offs. E.g., the Pangu series, including Pangu-$\alpha$ and Pangu-$\Sigma$ \cite{zeng2021pangu,ren2023pangu}.
\end{itemize}

\begin{figure*}[ht]
    \centering
    \includegraphics[width=1\textwidth]{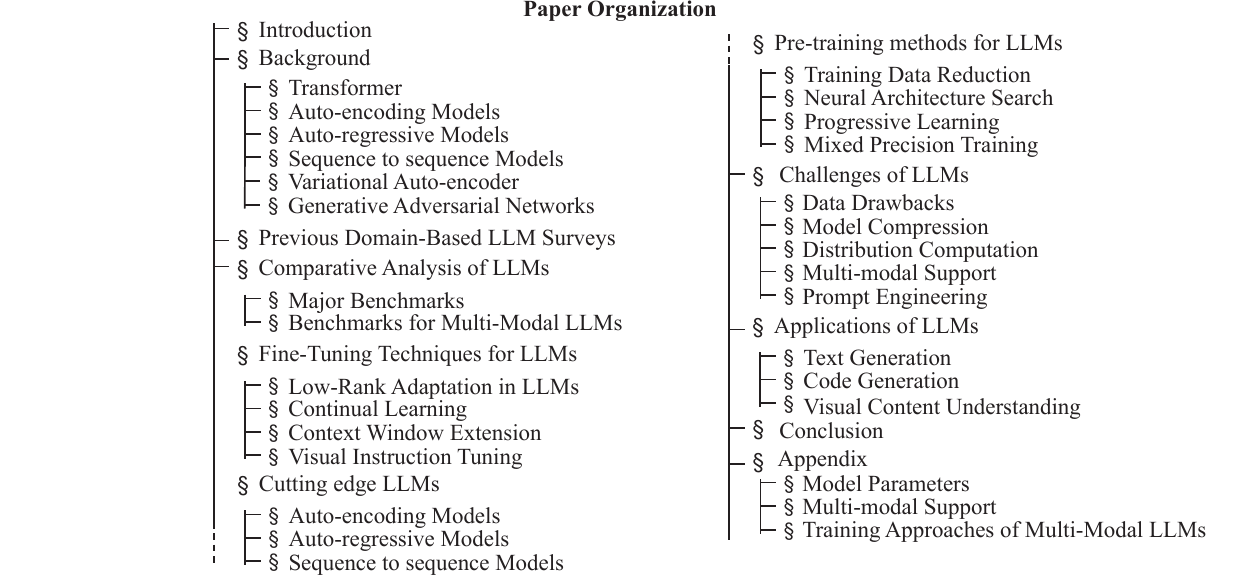} 
    \caption{Structured layout of the paper is presented, detailing the organization of sections including introduction, background and comparison, state-of-the-art LLMs and methodologies, challenges, and conclusion.}
    \label{fig:org}
\end{figure*}

    The structure of this paper is designed to furnish a concise yet thorough overview of LLMs. Initially, we delve into the essentials of LLMs, elucidating the underlying technologies and key terminology. We then provide a panoramic view of the LLM evolution, spotlighting influential contributors and institutions. Next, we explore avant-garde LLMs, pivotal in shaping the history of this domain. We culminate with a performance review of LLMs in their designated tasks, concluding with insights and reflections.
In essence, our contributions encompass:
\begin{enumerate}
\item Constructing a detailed timeline of seminal LLMs up to the release of this paper.
\item Distilling and collating pioneering technologies and strategies pivotal in the evolution of LLMs.
\item Undertaking a holistic comparison of LLMs across architectures and evaluating their performance metrics.
\item Assessing the overarching impacts and challenges posed by contemporary LLMs.
\end{enumerate}

\section{Background\label{sec:overview}}

LLMs predominantly encompass three architectural categorizations: encoder-only, decoder-only, and encoder-decoder. Each category has its unique strengths and constraints and finds relevance across various applications and contexts. This section explains the architecture behind modern LLMs, starting with the general transformer architecture, followed by an exploration of the three categories built upon this architecture.
\subsection{Transformer}
The contemporary landscape of LLMs predominantly employs the Transformer architecture, introduced by Vaswani et al. in 2017 \cite{vaswani2017attention}. This architecture represents a paradigm shift away from the recurrent sequence-to-sequence models, such as Long Short-Term Memory (LSTM) networks \cite{graves2012long} and Recurrent Neural Networks (RNNs) \cite{medsker2001recurrent}, which were conventionally used. The key innovation of the Transformer lies in its ability to process tokens in parallel, in contrast to the sequential processing constraint in LSTMs and RNNs, where the processing of each token depends on its predecessors. The Transformer achieves this through its multi-head self-attention mechanism, which allows for the parallelized training of models \cite{vaswani2017attention}.

\begin{figure}[ht]
    \centering
    \includegraphics[width=1\columnwidth]{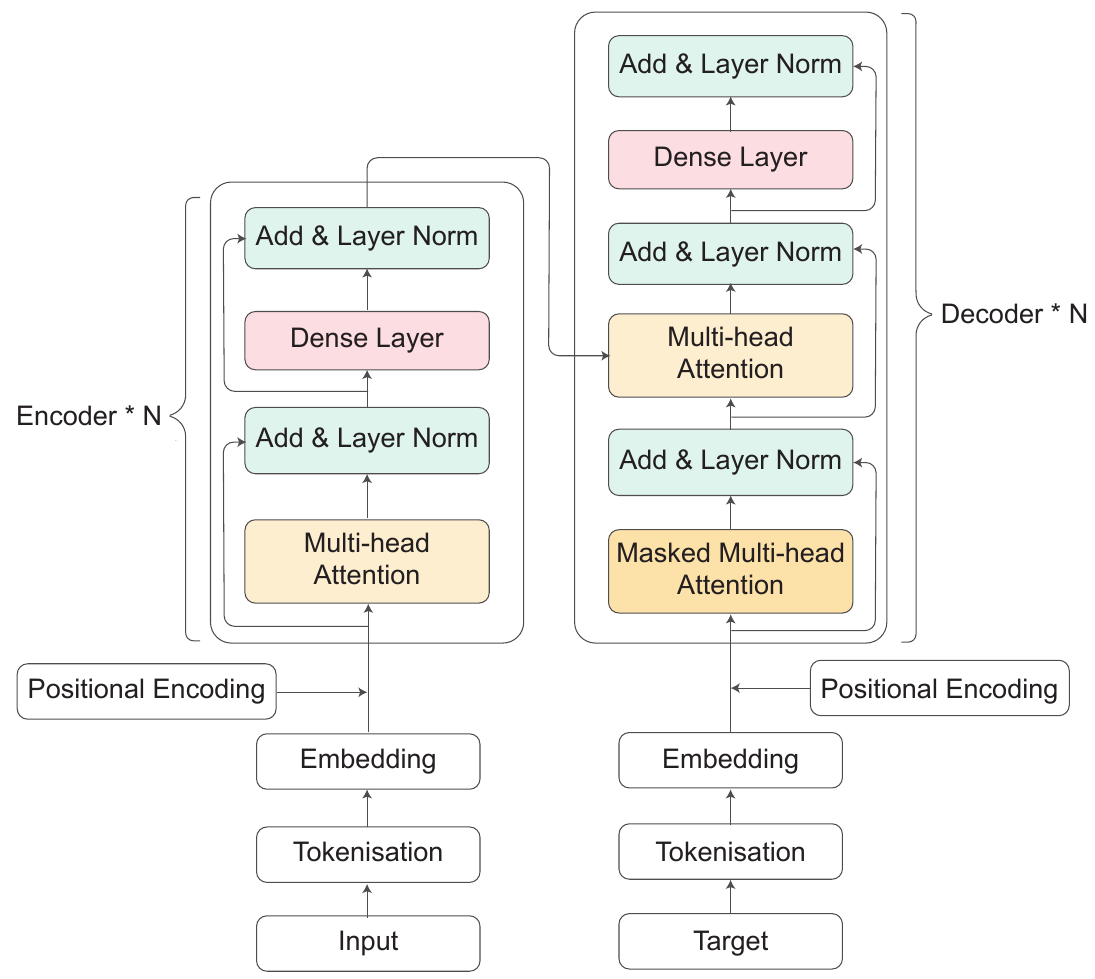} 
    \caption{The architecture of the Transformer model, which includes an encoder-decoder structure. Key components such as multi-head attention, positional encoding, and residual connections facilitate efficient learning and performance in tasks such as natural language processing and machine translation.}
    \label{fig:transformer}
\end{figure}

Conceptually, the Transformer architecture consists of encoder and decoder components. The encoder maps input sequences to a higher-dimensional embedding space, while the decoder generates output sequences from these embeddings. Typically, a Transformer model includes multiple layers of both encoders and decoders. Figure \ref{fig:transformer} provides an illustrative representation of the Transformer architecture.

\begin{table*}[!t]
\centering
\footnotesize
\begin{tabular}{|m{1.9cm}|m{4cm}|m{4.4cm}|m{5.8cm}|}
\hline
\textbf{Architecture} & \textbf{Advantage} & \textbf{Disadvantage} & \textbf{Example} \\
\hline
\textbf{Auto-Encoding} & Good at learning from context, efficient at representation learning & Not suitable for generating sequences & BERT family: BERT, ERNIE, ALBERT, RoBERTa \\
\hline
\textbf{Auto-Regressive} & Suited for generative tasks, effective at language modeling & Lacks context from future tokens during generation & LLaMA family: LLaMA, Alpaca, Vicuna; GPT family \\
\hline
\textbf{Sequence-to-Sequence} & Maps input sequences to word embeddings, conditional generation & High parameter count and complex training & Pangu family: Pangu-$\alpha$, Pangu-$\Sigma$ \\
\hline
\end{tabular}
\caption{Comparison of Auto-Encoding, Auto-Regressive, and Sequence-to-Sequence Models}
\label{table:encoder_decoder_comparison}
\vspace{-10pt}
\end{table*}



Unlike traditional models that process data sequentially, Transformers enable significantly faster and more efficient parallel processing by handling all parts of the input data simultaneously. To address the challenge of maintaining sequence information without inherent sequential processing, Transformers use a technique called \textbf{positional encoding}. This mechanism allows each token, such as a word in a sentence, to encode its relative position in the sequence. Positional encoding is essential; without it, the Transformer would treat a sentence as a bag of words, completely oblivious to the order of those words.

Positional encoding utilizes a specific mathematical formula involving sine and cosine functions. This formula ensures that each position in the sequence receives a unique encoding. By appending this encoding to the token's embedding, the model gains insight into the token's position within the sequence. The precise equations used are designed to provide a distinct positional signal for every possible position in the input sequence, thus enabling the model to interpret the order of words effectively, despite processing inputs in parallel.

A particular mathematical formula involving the sine and cosine functions is used in positional encoding. This formula ensures that a distinct encoding is assigned to every position in the sequence, enabling the model to be informed about the token's position by appending this encoding to the token's embedding. The precise equations employed are:

\begin{equation}
E(pos, 2i) = \sin\left(\frac{pos}{10000^{2i/\text{dim}}}\right)
\end{equation}
\begin{equation}
E(pos, 2i+1) = \cos\left(\frac{pos}{10000^{2i/\text{dim}}}\right)
\end{equation}

The token's position in the sequence is denoted by \( pos \), while \( i \) spans from 0 to half of the embedding dimension (\( \text{dim} \)), indexing even and odd positions respectively. The choice of sine and cosine functions is particularly advantageous because they provide a unique and consistent way to encode positional information across the embedding space. This setup not only simplifies the model's learning to attend based on relative positions but also enables generalization to sequence lengths beyond those encountered during training. The elegance of this method lies in its ability to imbue the model with the capacity to discern patterns in the data, enriched with positional context. This straightforward yet profound approach has been pivotal to the success of Transformer models in diverse tasks, ranging from text generation and language translation to applications beyond language, such as image recognition.

\subsection{Auto-encoding Models}
Primarily tailored for natural language processing tasks centered around comprehension, encoder-only models like BERT \cite{devlin2018bert}, ERNIE \cite{zhang2019ernie}, and ALBERT \cite{lan2019albert} have carved a niche for themselves. Training techniques such as bidirectional learning and masking enable them to excel in contextual understanding. However, they have certain limitations:
\begin{itemize}
    \item Constrained to fixed-length input sequences.
    \item Inherent context-dependency can be a hindrance for text generation.
    \item Given their composition lacks a decoder, downstream task adaptation necessitates fine-tuning.
\end{itemize}

\subsection{Auto-regressive Models}
These models, including renowned ones like GPT \cite{radford2018improving} and the LLaMA series \cite{touvron2023llama}, have gained prominence in recent times. Their auto-regressive design implies that token generation hinges on preceding tokens, rendering them apt for generation tasks. These models offer:
\begin{itemize}
    \item Flexibility in accepting varied input lengths, making them adept at extended data generation.
    \item Proficiency in few-shot or zero-shot tasks, circumventing the need for specific fine-tuning.
    \item However, their inability to capture overall context means they draw insights from antecedent tokens.
\end{itemize}


\subsection{Sequence-to-Sequence Models}
Models such as T5 \cite{raffel2020exploring} and GLM \cite{du2021glm},  harmonize strengths of the preceding two types. These models are adept at mapping input sequences to fixed-length embeddings, allowing the decoder to generate contextually relevant outputs. This makes them particularly effective for conditional generation tasks, such as summarization, translation, and question answering, where the output depends closely on the provided input.

The integration of encoder and decoder components enables Seq2Seq models to handle complex inputs but comes with drawbacks:
\begin{itemize}
    \item Amalgamation increases parameter count, potentially affecting efficiency.
    \item Training such models requires substantial computational resources due to the complexity of aligning the input and output sequences.
\end{itemize}

\subsection{Variational auto-encoder}
\begin{figure}[htbp]
    \centering
    \includegraphics[width=0.5\textwidth]{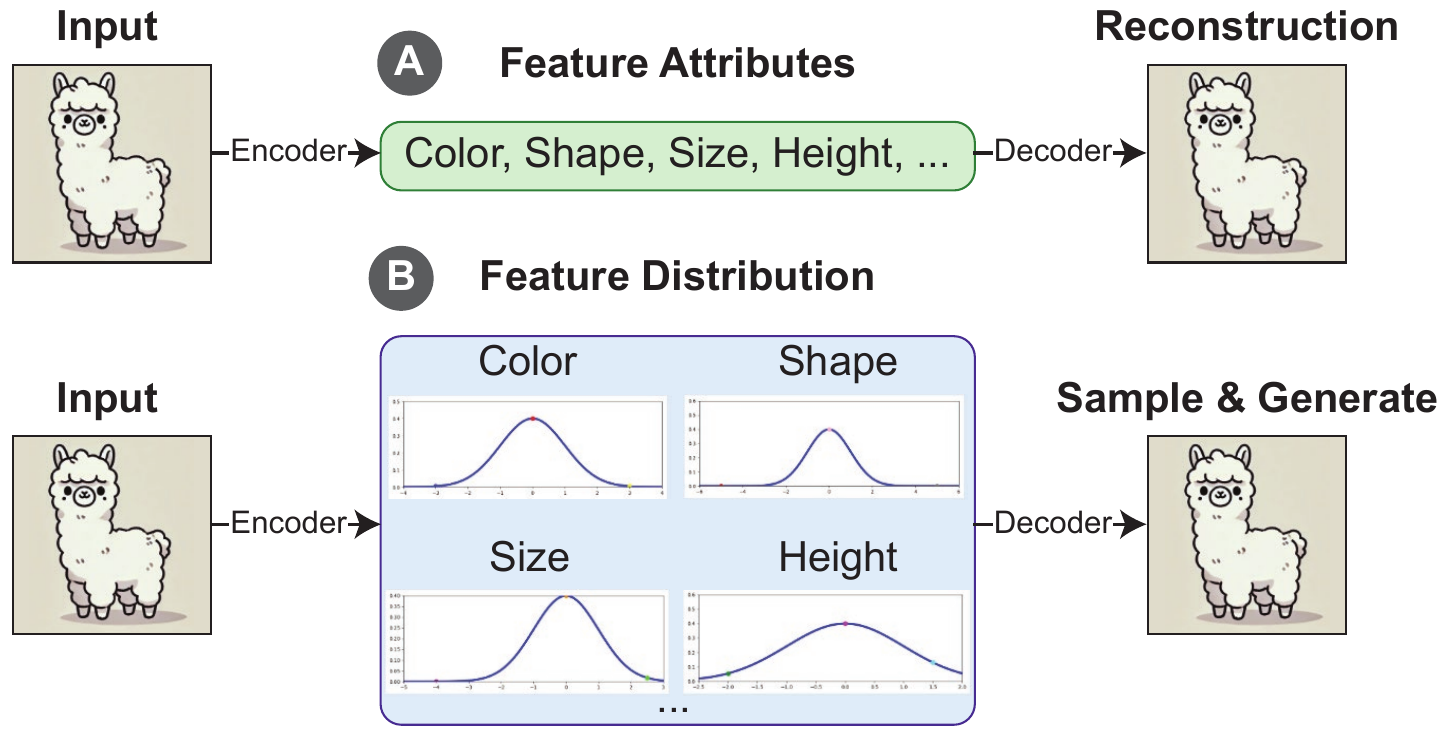} 
    \caption{(A) Workflow of Auto-encoder, auto-encoder encode the feature attribute directly. (B) Workflow of Variational Auto-encoder, different from auto-encoder, VAEs encode the feature distribution and reconstruct the image based on the sample of distribution, which give the VAEs the ability to generate new images.}
    \label{fig:ae-var}
\end{figure}

Variational Auto-encoder (VAE) \cite{kingma2013auto} is a sophisticated generative model that evolves from traditional auto-encoders (AE) by integrating probabilistic modeling to develop a meaningful and versatile latent space. Unlike standard AEs, which compress input data into a static representation that the decoder uses to reconstruct the original data, in VAE, the encoder produces a probability distribution defined by means and variances instead of a singular deterministic point. Figure \ref{fig:ae-var} shows the difference between of the principle of auto-encoders and variational auto-encoders. 

VAE utilizes probabilistic encoding to create a dynamic and adaptable latent space, allowing not only data reconstruction but also new data generation by sampling from learned probability distributions. This enhances model generalization and ensures smooth transitions in the latent space, crucial for tasks like data generation and augmentation. It leverages the reparameterization trick to keep gradients flowing through stochastic sampling processes during backpropagation, maintaining the differentiability of latent variables for conventional training. Their objective function balances reconstruction loss, assessing the accuracy of decoded samples against original inputs, with Kullback-Leibler (KL) divergence, promoting approximations of latent distributions to a standard Gaussian. This dual focus ensures both precise input reconstruction and a smooth, continuous latent space, making VAE powerful tools for applications in image generation, data augmentation, and anomaly detection. At the time this paper was released, VAEs had diversified into a wide array of variants. For a comprehensive overview and comparison of these different VAE variants, readers are encouraged to refer to \cite{zhai2018autoencoder} and \cite{wei2020variations} which provide extensive analyses and insights into the evolution and functionalities of these models.The training of VAE can be represented with following formula
$$
\mathcal{L}(\theta, \phi; x) = \mathbb{E}_{q_\phi(z|x)}[\log p_\theta(x|z)] - \mathrm{KL}[q_\phi(z|x) \| p(z)]
$$

Where $\mathbb{E}_{q_\phi(z|x)}[\cdot]$ represents the expectation under the distribution $q_\phi(z|x)$, $\log p_\theta(x|z)$ is the logarithm of the likelihood, which is how well the model can reconstruct the input from the latent variables, $\mathrm{KL}[q_\phi(z|x) \| p(z)]$ is the KL divergence between the approximate posterior $q_\phi(z|x)$ as a regularization by encouraging the posterior to be close to the prior with a Gaussian distribution.

\subsection{Generative Adversarial Network}
\begin{figure}[ht]
\centering
\includegraphics[width=\linewidth]{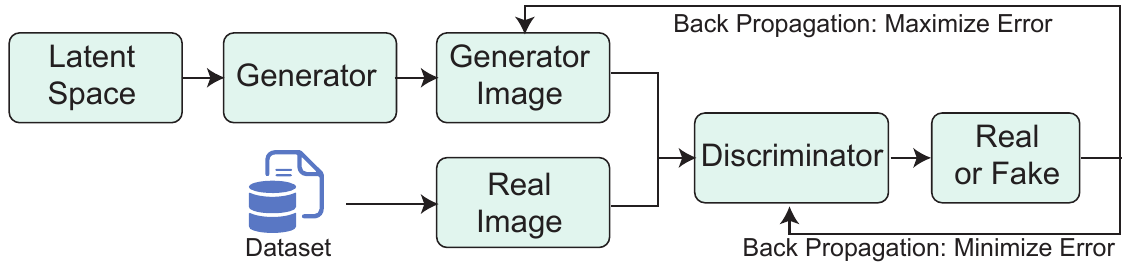}
\caption{Basic architecture of a Generative Adversarial Network (GAN). The generator creates synthetic images from a latent space, while the discriminator distinguishes between real images from the dataset and generated images. The generator is trained to maximize the discriminator’s error, while the discriminator is trained to minimize its error in distinguishing real from fake images, leading to adversarial learning between the two components.}
\label{fig:GAN}
\end{figure}

Generative Adversarial Networks (GANs) are a class of DL frameworks introduced by Goodfellow et al. \cite{Goodfellow2014}. GANs consist of two neural networks, a generator and a discriminator, which are trained simultaneously through adversarial processes. The generator aims to create synthetic data that resembles the real data, while the discriminator's role is to distinguish between real and synthetic data. Over time, as training progresses, the generator becomes better at creating realistic data, and the discriminator becomes better at differentiating real from fake data, illustrated in Figure \ref{fig:GAN}.

\begin{itemize}
    \item \textbf{Generator}: Takes a noise vector (randomly sampled) and maps it into a data space, aiming to produce samples that resemble the training data. The generator's goal is to generate data that is indistinguishable from the real data.
    \item \textbf{Discriminator}: Takes a sample (real or generated) and predicts whether it is real (from the training dataset) or fake (generated by the generator). It is trained to maximize the probability of correctly classifying real and generated samples.
\end{itemize}

The training process of GANs can be described as a min-max game where \( G \) is the Generator, \( D \) is the Discriminator, \( x \) represents real data samples, \( z \) is a noise vector, \( p_{\text{data}} \) denotes the data distribution, and \( p_z \) is the noise distribution.

\begingroup
\vspace{-10pt}
\setlength{\abovedisplayskip}{10pt} 
\setlength{\belowdisplayskip}{10pt} 
\footnotesize
\[
{ \min_G \max_D V(D, G) = \mathbb{E}_{x \sim p_{\text{data}}(x)}[\log D(x)] + \mathbb{E}_{z \sim p_z(z)}[\log (1 - D(G(z)))]
}
\]
\endgroup

\begin{table*}[ht]
\footnotesize
    \centering
    \begin{tabular} {|p{2.6cm}|p{0.6cm}|p{13.4cm}|}
        \hline
        \textbf{GAN Model} & \textbf{Year} & \textbf{Key Idea} \\ \hline
        Vanilla GAN \cite{Goodfellow2014} & 2014 & Introduced the fundamental adversarial training between the generator and discriminator. \\ \hline
        DCGAN \cite{Radford2015} & 2015 & Utilizes convolutional layers to enhance the performance and stability of GANs in image generation tasks. \\ \hline
        CGAN \cite{Mirza2014} & 2014 & Introduces conditioning variables (e.g., class labels) into both the generator and discriminator to control the output. \\ \hline
        WGAN \cite{Arjovsky2017} & 2017 & Replaces the original GAN loss with the Wasserstein distance to improve training stability and reduce mode collapse. \\ \hline
        WGAN-GP \cite{Gulrajani2017} & 2017 & Extends WGAN by adding a gradient penalty term to enforce the Lipschitz constraint more effectively. \\ \hline
        LSGAN \cite{Mao2016} & 2017 & Uses least-squares loss instead of the cross-entropy loss to address vanishing gradients and stabilize training. \\ \hline
        CycleGAN \cite{Zhu2017} & 2017 & Introduces cycle consistency loss to enable image-to-image translation without paired training data. \\ \hline
        StyleGAN \cite{Karras2018} & 2019 & Introduces a style-based generator architecture, allowing control over different aspects and details of generated images. \\ \hline
        BigGAN \cite{Zhang2018} & 2018 & Focuses on scaling up GANs using large batch sizes and deeper architectures to generate higher-quality images. \\ \hline
        SAGAN \cite{Brock2018} & 2018 & Incorporates self-attention mechanisms in GANs to capture long-range dependencies and generate detailed images. \\ \hline
        Progressive GAN \cite{Karras2017} & 2017 & Gradually increases the resolution of generated images during training to achieve more stable results. \\ \hline
        StarGAN \cite{Choi2017} & 2018 & Aims to perform multi-domain image-to-image translation using a single generator and discriminator. \\ \hline
    \end{tabular}
    \caption{Summary of Different GAN Models and Their Key Ideas}
    \label{tab:gan_variants}
    \vspace{-10pt}
\end{table*}

\subsubsection{Variants of GANs}

Since their inception, numerous variations of GANs have been proposed to address specific challenges such as mode collapse, training stability, and applicability to different types of data (e.g., images, text, or audio). Table \ref{tab:gan_variants} contains the summary of some of the most widely used GAN variants along with their unique characteristics.


Several surveys have extensively covered the advancements and applications of GANs. Wang et al. \cite{Wang2019} provide a detailed overview of GAN architectures and challenges like mode collapse. Gui et al. \cite{Gui2023} review methods to stabilize training and improve image quality, while Creswell et al. \cite{Creswell2017} focus on GANs' creative applications, including style transfer. Given the depth of these reviews, our paper will concentrate on Large Language Models (LLMs), exploring their generative capabilities and contributions to natural language understanding.


\begin{table*}[ht]
    \newcolumntype{D}{>{\centering\arraybackslash}m{1.2cm}}
    \newcolumntype{S}{>{\centering\arraybackslash}m{0.8cm}}
    \centering
    \caption{Survey Papers on Large Language Models: A comparative analysis}
    \scriptsize
    \begin{tabular}{|S|S|S|S|D|D|D|D|D|D|D|D|}
    \hline
    \textbf{LLM Survey} &\textbf{Month} &\textbf{Year} & \textbf{\# Refs.} & \textbf{Architecture} & \textbf{Dataset} & \textbf{Pre-training} & \textbf{Fine-tuning} & \textbf{Benchmark} & \textbf{Challenges} & \textbf{MLLMs} & \textbf{Applications} \\
    \hline
    \textbf{Ours} & Jul & 2024 & 427 & \tick & \tick & \tick & \tick & \tick & \tick & \tick & \tick \\
    \hline
    \cite{gallegos2023bias} & Jul & 2024 & 260 & \cross & \tick & \cross & \tick & \tick & \tick & \cross & \cross \\
    \hline
    \cite{shi2024continual} & Jun & 2024 & 392 & \tick & \cross & \tick & \tick & \tick & \tick & \tick & \cross \\
    \hline
    \cite{zhou2024survey} & Jun & 2024 & 299 & \cross & \cross & \tick & \tick & \tick & \tick & \cross & \tick \\
    \hline
    \cite{fang2024large} & Jun & 2024 & 233 & \cross & \tick & \cross & \tick & \tick & \tick & \cross & \tick \\
    \hline
    \cite{das2024security} & Jun & 2024 & 221 & \tick & \cross & \cross & \cross & \cross & \tick & \cross & \tick \\
    \hline
    \cite{jin2024jailbreakzoo} & Jun & 2024 & 154 & \cross & \tick & \cross & \tick & \tick & \tick & \tick & \cross \\
    \hline
    \cite{wu2023survey} & Jun & 2024 & 129 & \cross & \cross & \tick & \tick & \cross & \cross & \cross & \tick \\
    \hline
    \cite{akyash2024evolutionary} & Jun & 2024 & 42 & \cross & \tick & \tick & \tick & \tick & \tick & \cross & \tick \\
    \hline
    \cite{bai2024survey} & May & 2024 & 336 & \cross & \tick & \tick & \tick & \tick & \tick & \tick & \cross \\
    \hline
    \cite{xiao2024comprehensive} & May & 2024 & 269 & \tick & \tick & \tick & \tick & \tick & \tick & \tick & \tick \\
    \hline
    \cite{zhou2024large} & May & 2024 & 207 & \tick & \tick & \tick & \tick & \cross & \tick & \tick & \tick \\
    \hline
    \cite{huang2024comprehensive} & May & 2024 & 176 & \cross & \cross & \cross & \cross & \tick & \tick & \cross & \tick \\
    \hline
    \cite{qu2024tool} & May & 2024 & 171 & \cross & \cross & \cross & \cross & \tick & \tick & \cross & \tick \\
    \hline
    \cite{qin2024large} & May & 2024 & 168 & \cross & \cross & \tick & \tick & \cross & \tick & \cross & \tick \\
    \hline
    \cite{zhang2024large} & May & 2024 & 84 & \cross & \cross & \cross & \cross & \cross & \tick & \cross & \tick \\
    \hline
    \cite{kukreja2024literature} & May & 2024 & 42 & \tick & \tick & \tick & \cross & \cross & \cross & \cross & \tick \\
    \hline
    \cite{qin2024multilingual} & Apr & 2024 & 349 & \cross & \tick & \tick & \tick & \cross & \cross & \cross & \tick \\
    \hline
    \cite{dai2024unifying} & Apr & 2024 & 208 & \cross & \tick & \cross & \tick & \tick & \tick & \cross & \cross \\
    \hline
    \cite{yin2023survey} & Apr & 2024 & 206 & \tick & \tick & \tick & \tick & \tick & \tick & \tick & \tick \\
    \hline
    \cite{zhang2024survey} & Apr & 2024 & 174 & \cross & \cross & \cross & \tick & \tick & \tick & \cross & \tick \\
    \hline
    \cite{hu2024survey} & Apr & 2024 & 170 & \cross & \cross & \cross & \tick & \tick & \tick & \tick & \tick \\
    \hline
    \cite{bai2024hallucination} & Apr & 2024 & 140 & \cross & \tick & \tick & \tick & \tick & \tick & \tick & \cross \\
    \hline
    \cite{huang2024survey} & Apr & 2024 & 108 & \cross & \tick & \tick & \tick & \cross & \tick & \cross & \tick \\
    \hline    
    \cite{li2024pre} & Apr & 2024 & 54 & \tick & \cross & \cross & \tick & \cross & \cross & \cross & \tick \\
    \hline
    \cite{liu2023trustworthy} & Mar & 2024 & 471 & \cross & \tick & \tick & \cross & \tick & \tick & \cross & \tick \\
    \hline
    \cite{yao2024survey} & Mar & 2024 & 353 & \cross & \cross & \cross & \tick & \cross & \tick & \cross & \tick \\
    \hline
    \cite{chang2024survey} & Mar & 2024 & 269 & \cross & \tick & \cross & \cross & \tick & \tick & \cross & \tick \\
    \hline
    \cite{gao2023retrieval} & Mar & 2024 & 182 & \cross & \tick & \tick & \tick & \tick & \tick & \cross & \tick \\
    \hline
    \cite{zhang2023instruction} & Mar & 2024 & 165 & \tick & \tick & \cross & \tick & \tick & \cross & \tick & \cross \\
    \hline    
    \cite{hong2024advances} & Mar & 2024 & 148 & \cross & \tick & \cross & \tick & \tick & \tick & \cross & \cross \\
    \hline
    \cite{yan2024protecting} & Mar & 2024 & 140 & \cross & \cross & \tick & \tick & \cross & \tick & \cross & \cross \\
    \hline
    \cite{cao2024survey} & Mar & 2024 & 128 & \tick & \cross & \cross & \tick & \tick & \tick & \cross & \tick \\
    \hline
    \cite{liu2024large} & Mar & 2024 & 127 & \cross & \tick & \cross & \cross & \tick & \tick & \tick & \tick \\
    \hline
    \cite{esmradi2023comprehensive} & Mar & 2024 & 121 & \tick & \cross & \cross & \cross & \cross & \tick & \cross & \tick \\
    \hline
    \cite{chowdhury2024breaking} & Mar & 2024 & 66 & \cross & \tick & \cross & \cross & \tick & \tick & \cross & \cross \\
    \hline
    \cite{sun2023short} & Mar & 2024 & 25 & \cross & \tick & \cross & \cross & \cross & \tick & \cross & \tick \\
    \hline
    \cite{zhao2024explainability} & Feb & 2024 & 218 & \cross & \cross & \cross & \tick & \tick & \tick & \cross & \tick \\
    \hline
    \cite{zhu2023large} & Jan & 2024 & 231 & \tick & \tick & \tick & \tick & \tick & \tick & \cross & \tick \\
    \hline
    \cite{li2024cross} & Jan & 2024 & 46 & \cross & \tick & \cross & \cross & \tick & \cross & \cross & \cross \\
    \hline
    \cite{xi2023rise} & Dec & 2023 & 675 & \cross & \tick & \tick & \tick & \cross & \tick & \cross & \tick \\
    \hline
    \cite{huang2023survey} & Nov & 2023 & 268 & \cross & \tick & \tick & \tick & \tick & \tick & \tick & \cross \\
    \hline
    \cite{shayegani2023survey} & Oct & 2023 & 285 & \tick & \tick & \tick & \tick & \cross & \cross & \tick & \tick \\
    \hline
    \cite{zhang2023survey} & Oct & 2023 & 172 & \tick & \cross & \cross & \tick & \cross & \tick & \tick & \tick \\
    \hline
    \cite{min2023recent} & Sep & 2023 & 362 & \tick & \tick & \cross & \tick & \cross & \cross & \cross & \tick \\
    \hline
    \cite{zhang2023siren} & Sep & 2023 & 215 & \cross & \tick & \tick & \tick & \tick & \tick & \tick & \tick \\
    \hline
    \cite{zhu2023survey} & Sep & 2023 & 132 & \cross & \tick & \cross & \tick & \tick & \tick & \cross & \cross \\
    \hline
    \cite{10234662} & Aug & 2023 & 109 & \tick & \cross & \tick & \tick & \cross & \tick & \tick & \cross \\
    \hline
    \cite{wang2023pre} & Jul & 2023 & 378 & \tick & \tick & \tick & \tick & \tick & \cross & \tick & \tick \\
    \hline
    \cite{huang2022towards} & May & 2023 & 131 & \cross & \tick & \cross & \tick & \tick & \tick & \cross & \cross \\
    \hline
    \cite{kasneci2023chatgpt} & Apr & 2023 & 65 & \cross & \cross & \cross & \cross & \cross & \tick & \cross & \tick \\
    \hline
    \cite{zhao2023survey} & Mar & 2023 & 946 & \tick & \tick & \tick & \tick & \tick & \tick & \tick & \tick \\
    \hline
    \cite{mialon2023augmented} & Feb & 2023 & 191 & \cross & \cross & \tick & \tick & \tick & \tick & \cross & \cross \\
    \hline
    \cite{kalyan2022ammu} & Feb & 2022 & 216 & \tick & \cross & \cross & \tick & \tick & \tick & \cross & \tick \\
    \hline
    \cite{kalyan2021ammus} & Aug & 2021 & 304 & \tick & \tick & \tick & \tick & \tick & \tick & \cross & \cross \\
    \hline
    \cite{zaib2020short} & Feb & 2020 & 20 & \tick & \cross & \tick & \tick & \tick & \tick & \cross & \cross \\
    \hline
    \end{tabular}
    \label{table_survey_papers}
\end{table*}

\section{Previous Domain-based LLM Surveys}

In this section, we conduct a comprehensive analysis of the existing surveys on large language models (LLMs). We provide a comparative evaluation of these survey papers based on the topics they address. The surveys are chronologically organized in Table \ref{table_survey_papers}, allowing readers to track the evolution of research focus over time. By examining the content covered in these surveys, as summarized in the table, readers can gain a nuanced understanding of the progress made in the development of advanced LLMs. The categories are comprised of:

\begin{itemize}
    \item \textbf{Architecture:} Details about the structural design of the LLMs discussed, including model types and configurations, including Decoder only, Encoder only, and Decoder-Encoder Models.
    \item \textbf{Dataset:} Information about the datasets used for training and evaluating the LLMs.
    \item \textbf{Pre-training:} Methods and techniques used for training the foundation LLMs.
    \item \textbf{Fine-tuning:} Strategies for adapting pre-trained LLMs to specific tasks or domains to improve domain-specific performance.
    \item \textbf{Benchmark:} Evaluation metrics and benchmark datasets used to assess LLM/ MLLM performance.
    \item \textbf{Challenges:} Identification of challenges and techniques to optimize development and deployment of LLMs.
    \item \textbf{MLLMs:} Discussion on Multilingual Language Models and their specific considerations.
    \item \textbf{Applications:} Real-world applications and use cases of state-of-the-art LLMs.    
\end{itemize}

Our survey provides a brief overview of all these categories but delves deeply into the Architecture, Benchmark, and Challenges aspects. This in-depth focus aims to offer a detailed understanding of the structural innovations, evaluation methodologies, and challenges in the field of large language models. Some notable surveys in the domain of LLM/MLLMs are highlighted:
\begin{itemize}
    \item \textbf{The survey by Xiao et al. \cite{xiao2024comprehensive}} provides an extensive overview of LLMs and MLLMs within the medical domain. However, our survey addresses the general methodologies for fine-tuning and LLM architectures, applicable to a wide range of use cases beyond the medical field. By addressing these broader aspects, our survey provides a more holistic view of the advancements, challenges, and future directions in the field of LLMs, making it a valuable resource for a wider audience.
    
    \item \textbf{The survey by Yin et al. \cite{yin2023survey}} offers an overview of the progress in multimodal large language models (MLLMs). It covers the basic formulation, related concepts, research topics, technical points, challenges, and future directions of MLLMs. While the MLLM survey provides valuable insights, our survey fills the gaps by offering detailed technical analysis, broader domain coverage, comprehensive comparative evaluations, and in-depth discussions on challenges. Our survey is a more versatile and informative resource for a wider audience in regards to MLLMs.
    
    \item \textbf{The notable survey from Zhao et al. \cite{zhao2023survey}} from early 2023 offers a comprehensive review of the development and applications of LLMs. However, this survey does not cover the latest developments and models in the fast-evolving field of LLMs. Since the pace of advancement in LLMs is rapid, many newer models and techniques that have emerged in the latter part of 2023 and early 2024 are not included. Our survey fills this gap by including the most recent advancements and models, and providing an in-depth benchmarking analysis for researchers and practitioners.
\end{itemize}

Table \ref{tab:survey_table} presents a review of the available survey papers on large language models (LLMs) as of the time of this writing. Each survey offers unique insights into the application of LLMs across various domains. To facilitate a clearer understanding, these surveys are categorized into three types:

\begin{itemize}
\item General surveys providing an overview of the evolution of LLMs.
\item Application-oriented surveys focusing on specific fields.
\item Technical surveys detailing the algorithms and techniques used in modern LLMs.
\end{itemize}

This comprehensive categorization aids in understanding how LLMs adapt to specialized vocabularies and regulatory frameworks across different settings.

Overall, our survey on LLMs serve as a valuable resource for researchers and practitioners, offering a consolidated view of the state-of-the-art in LLMs.

\begin{table} [ht]
    \caption{Collection of previous LLM survey paper categorized into the major topic of their discussion}\label{table1}
    \centering
    \footnotesize
    \begin{tabular}{|m{1.2cm}<{\centering}|m{3.1cm}|m{3cm}<{\raggedright\arraybackslash}|}
        \hline
        \textbf{Category} & \textbf{Field} & \textbf{Literature} \\
        \hline
        \multirow{3}{*}{Overview} & General LLMs & \cite{zhao2023survey, min2023recent, qin2024large, kukreja2024literature, kalyan2021ammus} \\
        \cline{2-3}
         & Multilingual & \cite{qin2024multilingual} \\
         \cline{2-3}
        & Multimodality & \cite{yin2023survey, bai2024survey} \\
        \hline
        \multirow{3}{*}{Technology} & Evaluation & \cite{chang2024survey} \\
        \cline{2-3}
         & Agent Systems & \cite{xi2023rise, zhang2024survey, hu2024survey} \\
        \cline{2-3}
         & Explainability \& Reasoning & \cite{zhao2024explainability, huang2022towards, hong2024advances} \\
         \cline{2-3}
         & Instruction Tuning & \cite{zhang2023instruction} \\
          \cline{2-3}
         & Model Compression & \cite{zhu2023survey} \\
       \cline{2-3}
         & Hallucination & \cite{zhang2023siren, bai2024hallucination, huang2023survey} \\
        \cline{2-3}
         & Fast Inference & \cite{zhou2024survey} \\
         \cline{2-3}
         & Augmentation & \cite{gao2023retrieval, mialon2023augmented, 10234662, huang2024survey} \\
          \cline{2-3}
         & Alignment & \cite{liu2023trustworthy} \\
       \cline{2-3}
         & Tool Usage & \cite{qu2024tool} \\
      \cline{2-3}
         & Continual Learning & \cite{shi2024continual} \\
       \cline{2-3}
         & Controllable Generation & \cite{zhang2023survey} \\
       \cline{2-3}
         & Security \& Privacy & \cite{yan2024protecting, esmradi2023comprehensive, gallegos2023bias, dai2024unifying, chowdhury2024breaking, das2024security, li2024cross, shayegani2023survey, jin2024jailbreakzoo, yao2024survey} \\
        \hline
        \multirow{3}{*}{Application} & Casual Inference & \cite{liu2024large} \\
        \cline{2-3}
         & Information Retrieval & \cite{zhu2023large} \\
        \cline{2-3}
         & Education & \cite{kasneci2023chatgpt} \\
         \cline{2-3}
         & Reinforcement Learning & \cite{cao2024survey} \\
          \cline{2-3}
         & Legal & \cite{sun2023short} \\
           \cline{2-3}
         & Recommendation System & \cite{wu2023survey} \\
        \cline{2-3}
         & Telecommunication & \cite{zhou2024large} \\
         \cline{2-3}
         & Conversational AI & \cite{zaib2020short} \\
          \cline{2-3}
         & Text Generation & \cite{li2024pre} \\
       \cline{2-3}
         & Medical \& Biological  & \cite{huang2024comprehensive, kalyan2022ammu, xiao2024comprehensive, wang2023pre} \\
        \cline{2-3}
         & Transportation  & \cite{zhang2024large} \\
         \cline{2-3}
         & Data Science  & \cite{fang2024large} \\
          \cline{2-3}
         & Security & \cite{akyash2024evolutionary} \\
        \hline
    \end{tabular}
    \label{tab:survey_table}
\end{table}

\section{Comparative Analysis of LLMs}
\label{sec:comparative_analysis}
This section provides a comparative analysis of prominent Language Models using various benchmarks, which assess the models' capabilities in language understanding, reasoning, and multimodal tasks. These benchmarks are designed to evaluate different aspects of language comprehension and cognitive abilities.

\subsection{Major Benchmarks}
Standardized benchmarks are essential for evaluating LLMs' performance across tasks, offering a means to compare models and identify their strengths and weaknesses. This section highlights some of the most widely used benchmarks.

\label{subsec:major_benchmarks}

\subsubsection{MMLU (Massive Multitask Language Understanding)}
\label{sssec:mmlu}
The Massive Multitask Language Understanding (MMLU) benchmark is a collection of 57 tasks that cover a wide range of subjects from human-level concepts to high school examinations. This benchmark evaluates the comprehensive understanding and generalization power of language models across diverse topics.

\textbf{Key Features:}
\begin{itemize}
    \item \textbf{Coverage:} Broad domain coverage from humanities to STEM.
    \item \textbf{Task Types:} Multiple choice questions with four options, requiring not only language understanding but also domain-specific knowledge.
    \item \textbf{Evaluation Metric:} Accuracy of predicting the correct answer among the given choices.
\end{itemize}

\begin{figure}[ht]
\centering
\includegraphics[width=\linewidth]{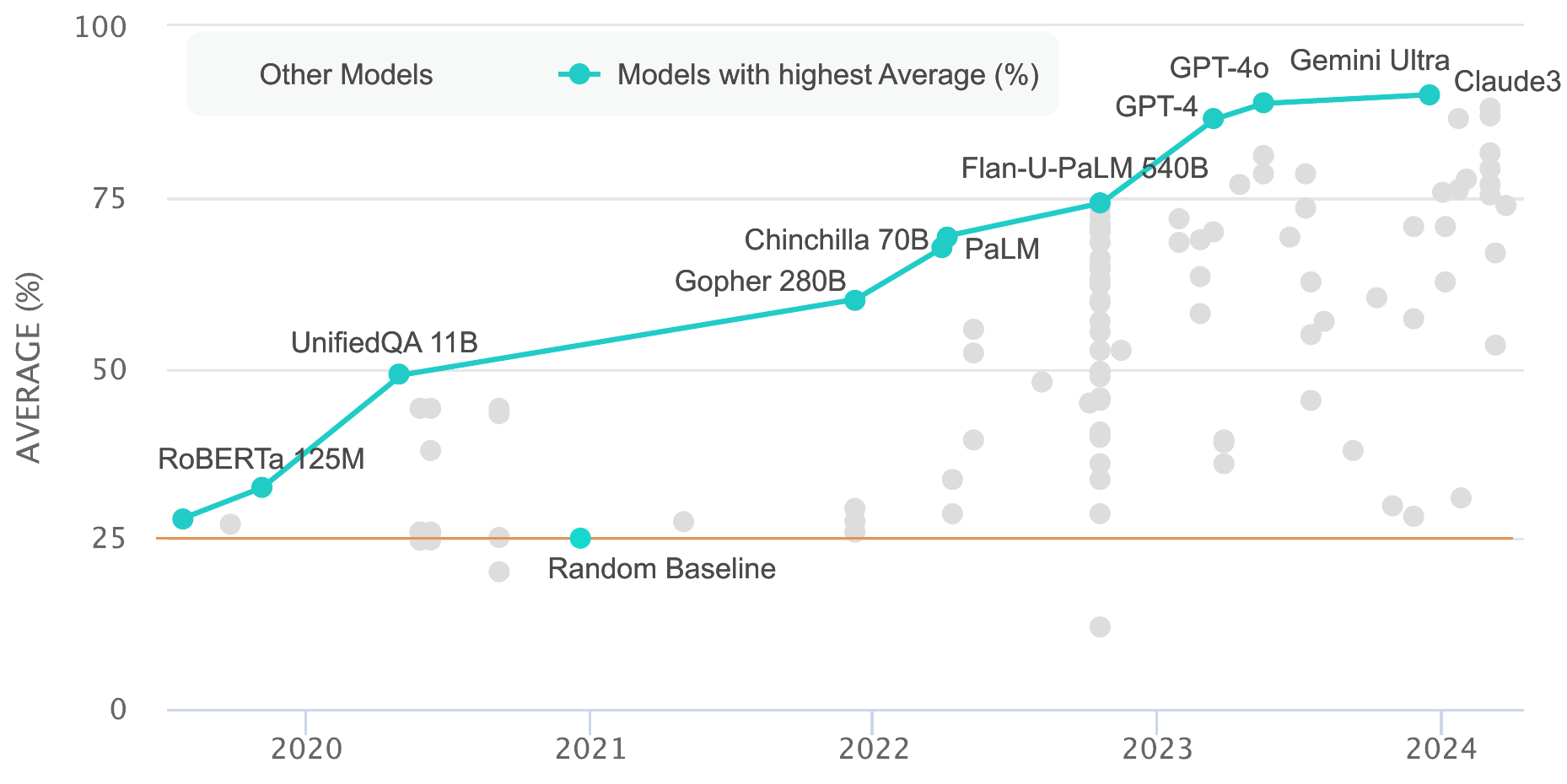}
\caption{ Performance trajectory of various LLMs on the MMLU benchmark illustrating  average accuracy percentages. The 'Random Baseline' represents a lower bound of performance, while the highlighted teal line traces the models with the highest average scores each year, culminating with the introduction of models like GPT-4o and Gemini Ultra that set new benchmarks for language understanding.}
\label{fig:mmlu}
\end{figure}

\subsubsection{SuperGLUE}
\label{sssec:superglue}
SuperGLUE is designed as an advanced benchmark to evaluate and promote improvements in the critical reasoning and prediction-making abilities of AI models beyond the GLUE benchmark. It includes a set of more challenging tasks that require deeper natural language understanding and broader reasoning capabilities.

\begin{figure}[ht]
\centering
\includegraphics[width=\linewidth]{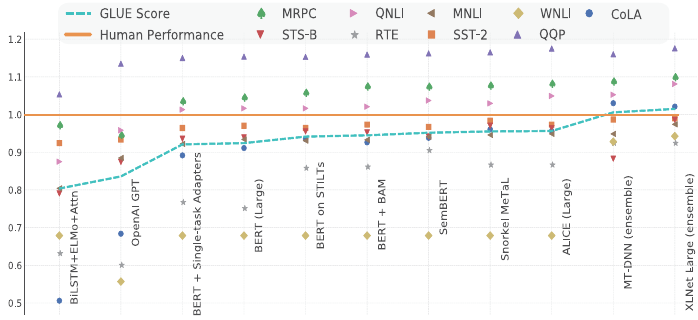}
\caption{ Comparative performance visualization of models on the GLUE benchmark, adjusted to a unified scale with human performance normalized to a score of 1.0. The summary score represents an aggregate of nine individual tasks, with an averaged score for tasks containing multiple metrics. The breakdown across the tasks demonstrates the relative strengths and weaknesses of each submitted system in various areas of language understanding.}
\label{fig:GLUE}
\end{figure}

\textbf{Key Features:}
\begin{itemize}
    \item \textbf{Complexity:} Tasks are specifically chosen to be more difficult and diverse than those in GLUE.
    \item \textbf{Task Variety:} Includes question answering, entailment, coreference resolution, and word sense disambiguation, among others.
    \item \textbf{Data Sources:} Comprises datasets that have been either newly created or significantly expanded upon for heightened difficulty and variability.
    \item \textbf{Evaluation Metric:} Composite score that is calculated based on performance across all constituent tasks, promoting models that achieve balanced capabilities across a broader array of challenges.
\end{itemize}

\subsubsection{HellaSwag}
\label{sssec:hellaswag}
HellaSwag is a benchmark designed to test a model's common sense and ability to complete scenarios using everyday knowledge. The tasks involve predicting the ending of descriptions of everyday activities.

\textbf{Key Features:}
\begin{itemize}
    \item \textbf{Contexts:} Comes from diverse sources such as Wikipedia and instructional videos.
    \item \textbf{Task Type:} Choose the most plausible continuation among four provided options.
    \item \textbf{Evaluation Metric:} Accuracy of predictions.
\end{itemize}

\begin{figure}[ht]
\centering
\includegraphics[width=\linewidth]{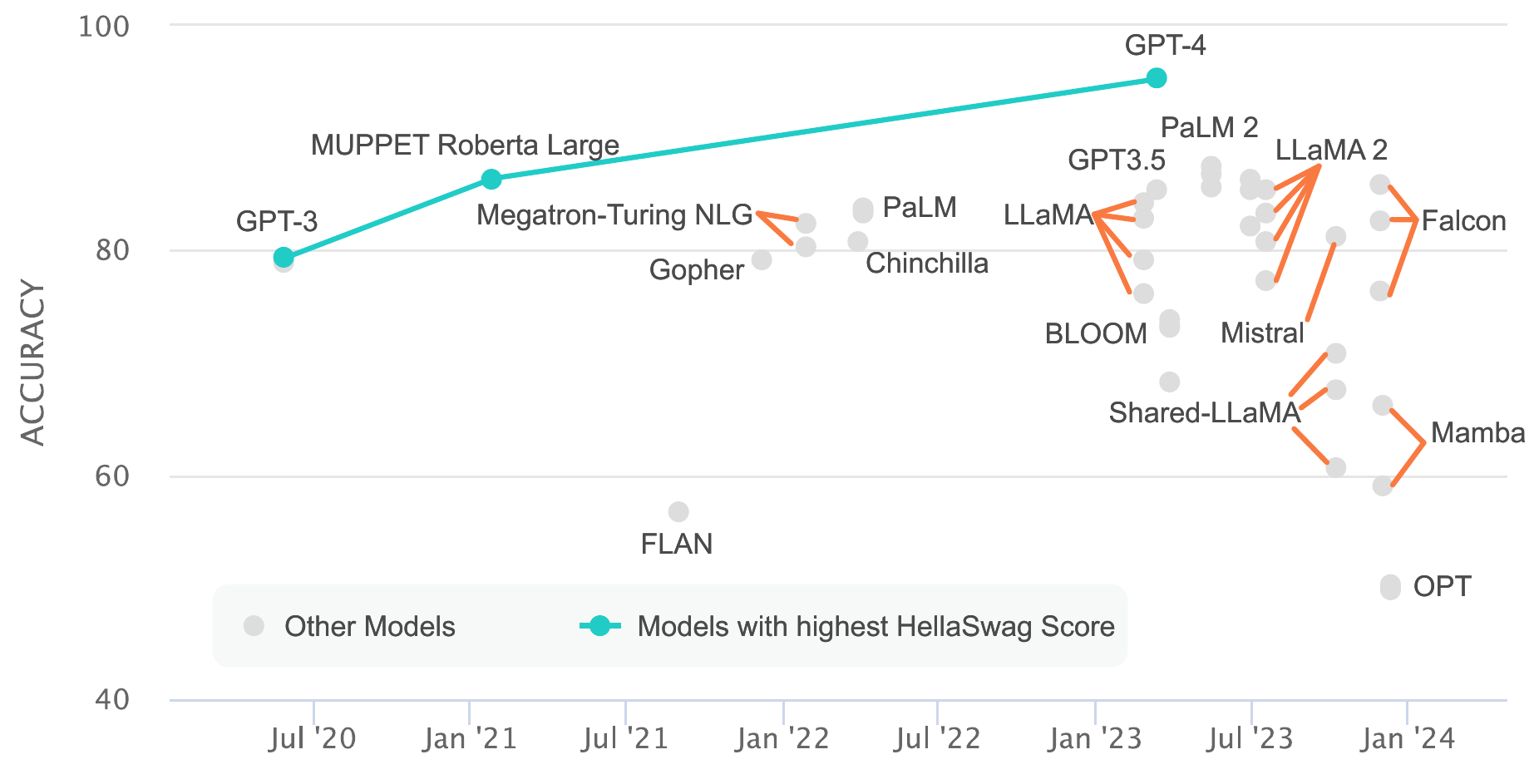}
\caption{Progression of HellaSwag benchmark accuracy scores from tracks the performance of various LLMs achieving the highest HellaSwag scores. Notable milestones include the introduction of models like GPT-4 and PaLM 2, which significantly surpass previous models in accuracy, indicating substantial advancements in AI's commonsense reasoning and contextual understanding abilities.}
\label{fig:Hella_Swag}
\end{figure}
\vspace{-10pt}

\subsubsection{ARC (AI2 Reasoning Challenge)} 
ARC presents models with grade-school-level multiple-choice science questions, testing their ability to understand text and apply reasoning skills.


\textbf{Key Features:}
\begin{itemize}
    \item \textbf{Difficulty Levels:} Contains both Easy and Challenge sets to adapt to different model capabilities.
    \item \textbf{Evaluation Metric:} Accuracy on correctly answered questions.
\end{itemize}

\begin{figure}[ht]
\centering
\includegraphics[width=\linewidth]{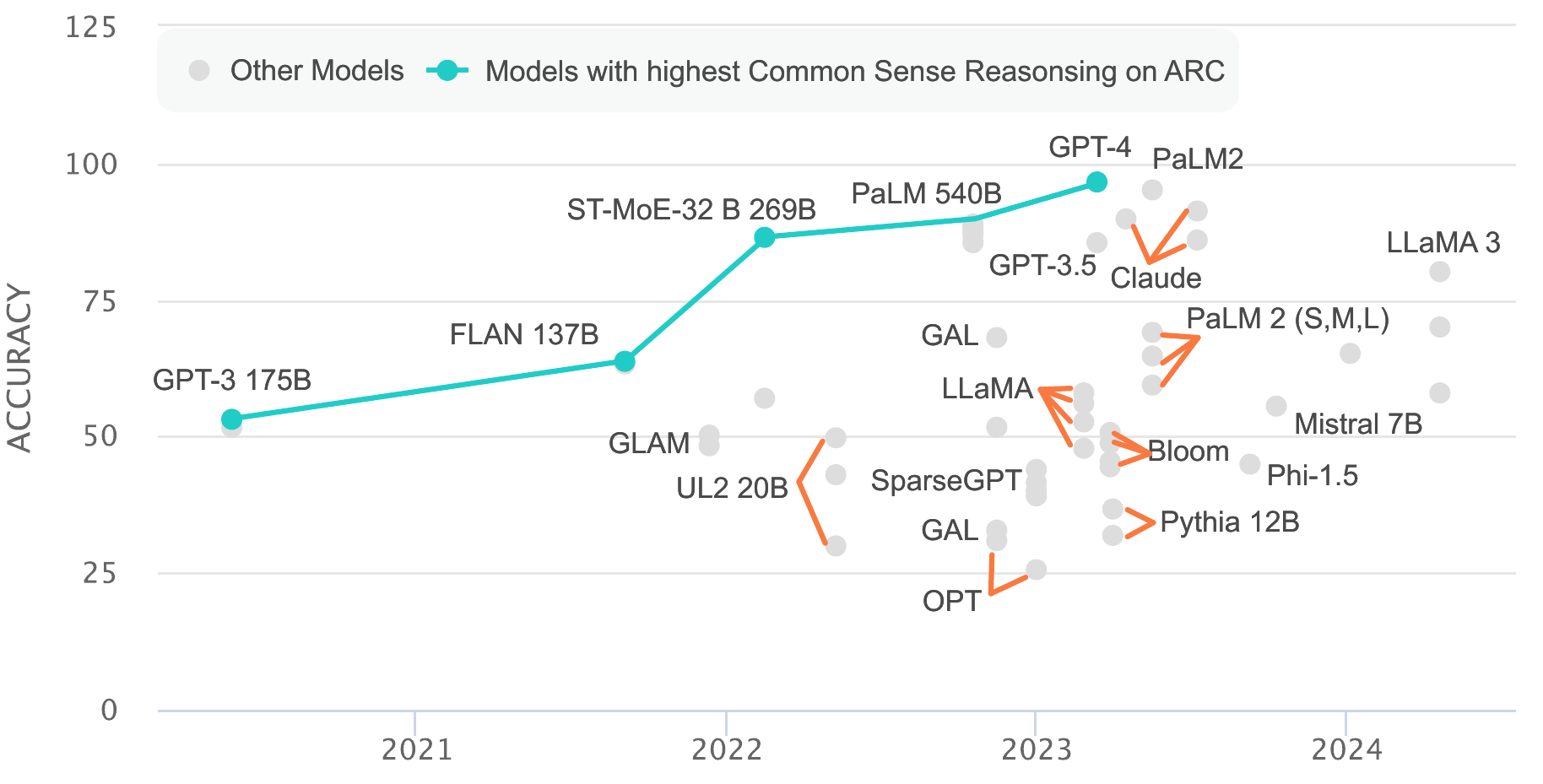}
\caption{  Advancement in accuracy of various LLMs on the AI2 Reasoning Challenge (ARC) aimed at appraising common sense reasoning. Newer models like GPT-4 and PaLM demonstrate significantly improved common sense reasoning capabilities. Notably, variations in model training approaches, such as zero-shot and few-shot learning, are reflected in differentiated performance levels.}
\label{fig:ARC}
\end{figure}
\subsubsection{WinoGrande}
\label{sssec:winogrande}
WinoGrande is a large-scale dataset of winograd schemas that are designed to test common sense reasoning within AI models.

\textbf{Key Features:}
\begin{itemize}
    \item \textbf{Scale:} One of the largest datasets for commonsense reasoning.
    \item \textbf{Task Type:} Sentence completion that requires resolving ambiguous pronouns.
    \item \textbf{Evaluation Metric:} Accuracy of choosing the correct entity.
\end{itemize}

\begin{figure}[ht]
\centering
\includegraphics[width=\linewidth]{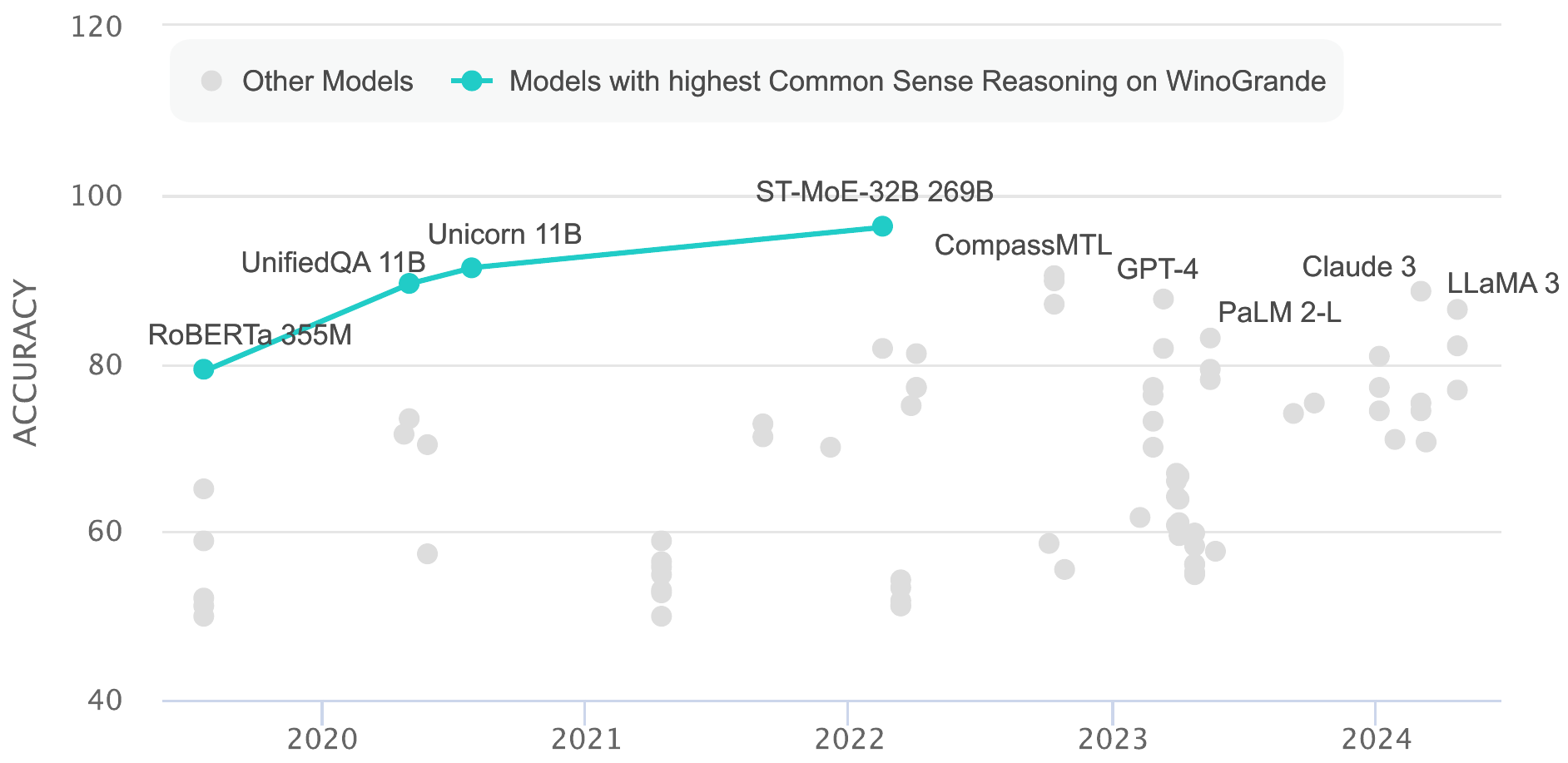}
\caption{ Showcasing the ascent in accuracy of diverse LLMs on the WinoGrande benchmark. Remarkable progress is seen with the likes of GPT-4 and the various iterations of the PaLM model, reflecting significant advancements in the field's pursuit of nuanced language understanding and common sense inference capabilities.}
\label{fig:WinoGrade}
\end{figure}

\subsection{Benchmarks for Multimodal LLMs}
\label{subsec:multimodal_benchmarks}
As the field of AI progresses, benchmarks for evaluating multimodal capabilities of LLMs have become increasingly relevant. Some notable multimodal benchmarks include:

\subsubsection{NLVR2 (Natural Language for Visual Reasoning for Real)}
\label{subsec:nlvr2}


The NLVR2 benchmark is a challenging dataset designed for evaluating AI models' ability in visual reasoning with natural language. It requires models to determine whether a given natural language statement accurately describes a pair of images. Unlike standard object recognition or image captioning tasks, NLVR2 demands a deeper understanding of both the visual content and the semantics of the language, making it a more complex challenge.

\begin{figure}[ht]
\centering
\includegraphics[width=\linewidth]{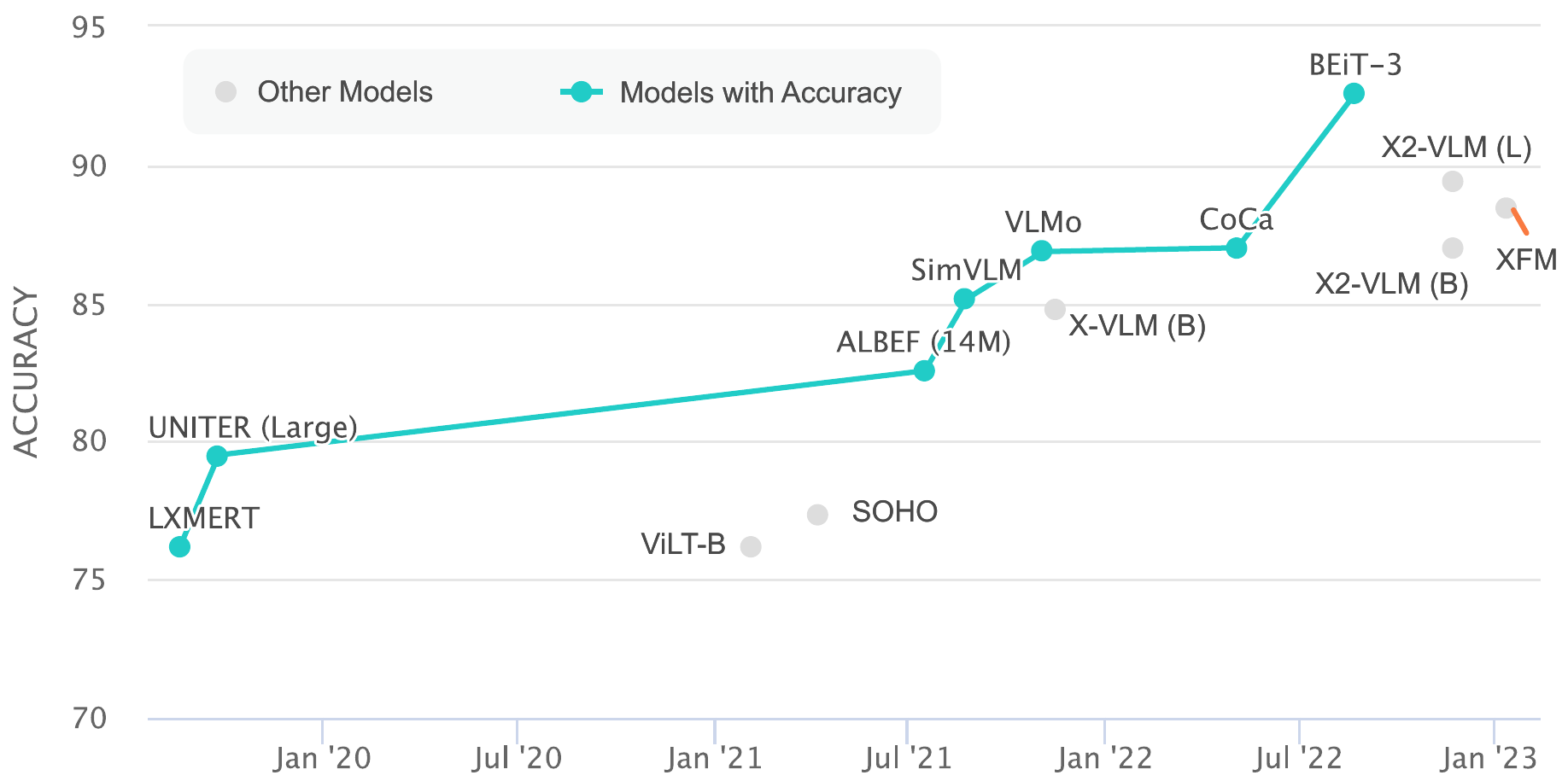}
\caption{ This graph depicts the accuracy trend of models on the NLVR2 benchmark. The multi-modal LLMs showcase rapid improvements in visual and linguistic reasoning capabilities. Notable performers such as BEiT-3 and X2-VLM (L) represent the cutting edge of multi-modal LLMs, indicating their superior proficiency in interpreting complex visual-language tasks.}
\label{fig:NLVR2}
\end{figure}

\textbf{Key Features:}
\begin{itemize}
    \item \textbf{Data Composition:} Consists of pairs of images accompanied by a textual description. The model's task is to verify the truthfulness of the description given the image pair.
    \item \textbf{Reasoning Requirement:} The models must interpret the images in the context of spatial relations, counting, and comparison, making this benchmark particularly suited for evaluating multimodal comprehension.
    \item \textbf{Performance Metrics:} The main metric is accuracy, indicating the model's ability to correctly validate the statement against the visuals.
    \item \textbf{Challenges:} Involves disambiguating ambiguous language, understanding complex statements, and a deep integration of visual and textual information.
\end{itemize}

Models successful on NLVR2 must not only integrate visual and textual information but also accurately capture the subtleties of language that relate to the visual world.

\subsubsection{Visual Question Answering (VQA) Benchmark}
\label{subsec:vqa}

The Visual Question Answering (VQA) benchmark stands as a measure of an AI system's ability to answer questions pertaining to given images. This multimodal benchmark combines natural language processing with image recognition to test a model's comprehensive understanding of visual content as it relates to conceptual and factual queries.

\begin{figure}[ht]
\centering
\includegraphics[width=\linewidth]{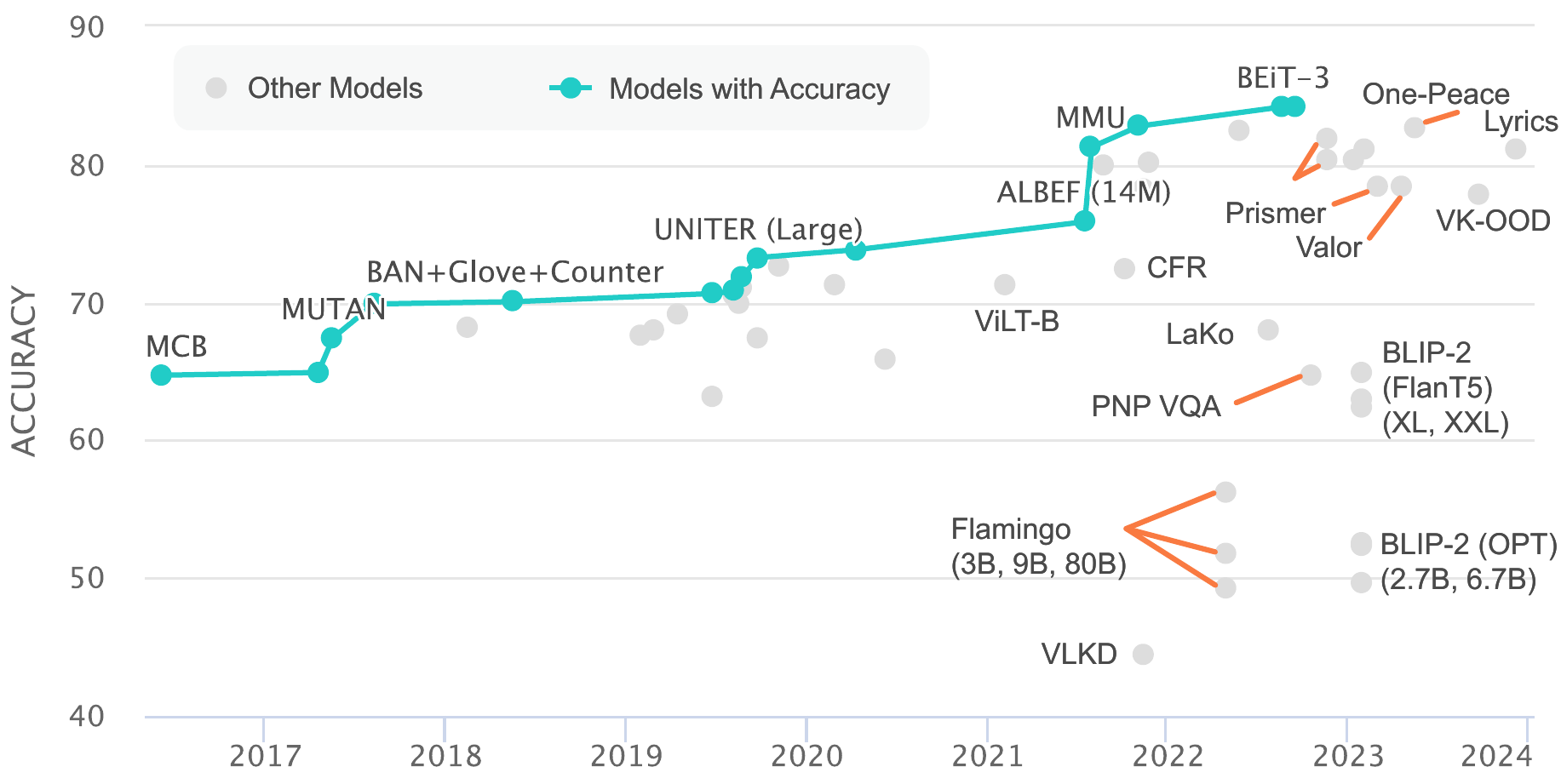}
\caption{ Accuracy performance graph for various AI models on the VQA, illustrating the evolution of AI in comprehending and answering visually grounded questions. The graph underscores the significant strides made in multimodal AI, with models like BEiT-3 showing remarkable precision in visual-textual understanding.}
\label{fig:VQA}
\end{figure}

\textbf{Key Features:}
\begin{itemize}
    \item \textbf{Task Composition:} Involves an open-ended task where a model is presented with an image and a related question in natural language. The model must provide an accurate answer to the question, reflecting a correct understanding of the visual context.
    \item \textbf{Diversity of Questions:} The questions are designed to cover a wide array of types, including object detection, counting, color determination, spatial understanding, and inferential reasoning based on the image content.
    \item \textbf{Answer Formats:} The answers may be in various forms, including single words, numbers, or short phrases.
    \item \textbf{Performance Metrics:} Accuracy is the primary metric, supplemented by consistency and plausibility scores in some variations of the benchmark.
    \item \textbf{Challenges:} The task challenges models to understand and process visual data in conjunction with textual information, requiring a high level of multimodal integration and reasoning.
\end{itemize}

The VQA benchmark is crucial for the development of AI systems that interact with the visual world in a meaningful and contextually aware manner, a necessity for applications ranging from assistive technologies to automated content moderation.

These benchmarks test the integration of visual and textual data, crucial for applications requiring a holistic understanding of multimodal inputs.

\section{Fine-tuning techniques for LLMs}
Fine-tuning methods for LLMs are utilized in a variety of applications including domain specialization, performance improvement, and bias mitigation. Key approaches to fine-tuning LLMs, such as Parameter-Efficient Fine-Tuning (PEFT), are often emphasized due to their diverse applications and reduced computational demands as compared to complete model training and these techniques are detailed in this section. 

\begin{table*}[!htp]
\centering
\footnotesize
\begin{tabular}{| m{2.5cm} | m{2.2cm} | m{11.5cm} |}
\hline
\textbf{PEFT Methods for PLMs} & \textbf{Subcategory} & \textbf{Techniques} \\
\hline
\multirow{5}{2.5cm}{Additive Fine-tuning} & Adapter-based Fine-tuning & Adapter Design: Serial Adapter \cite{houlsby2019parameter}, Parallel Adapter \cite{14-unified-view-transfer-peft}, CIAT \cite{zhu2021counter}, CoDA \cite{lei2023conditional} \newline Multi-task Adaptation: AdapterFusion \cite{4-AdapterFusion}, AdaMix \cite{29-adamix}, PHA \cite{zhao2023prototype}, AdapterSoup \cite{chronopoulou2302adaptersoup}, MerA \cite{he2023mera}, Hyperformer \cite{mahabadi2021parameter} \\
\cline{2-3}
 & Soft Prompt-based Fine-tuning & Soft Prompt Design: Prefix-tuning \cite{11-prefix-tuning}, Prefix-Propagation \cite{Prefix-Propagation}, p-tuning v2 \cite{47-P-tuning-v2}, APT \cite{zhang2023towards}, p-tuning \cite{3-GPT-understands-too}, prompt-tuning \cite{13-prompt-tuning}, Xprompt \cite{xprompt}, IDPG \cite{IDPG}, LPT \cite{LPT}, SPT \cite{SPT}, APrompt \cite{Aprompt} \newline Training Speedup: SPoT \cite{49-spot}, TPT \cite{50-transfer-prompt}, InfoPrompt \cite{wu2023infoprompt}, PTP \cite{ptp}, IPT \cite{51-intrinsic-subspace}, SMoP \cite{SMoP}, DePT \cite{shi2023dept} \\
\cline{2-3}
 & Others & (IA)³ \cite{16-few-shot-peft-in-context-learning}, MoV \cite{zadouri2023pushing}, SSF \cite{ssf}, IPA \cite{IPA} \\
\hline
\multirow{2}{2.5cm}{Selective Fine-tuning} & Unstructural Masking & U-Diff pruning \cite{guo2020parameter}, U-BitFit \cite{lawton2023neural}, PaFi \cite{liao2023parameter}, FishMask \cite{36-fish-masks}, Fish-Dip \cite{das2023unified}, LT-SFT \cite{ansell2021composable}, SAM \cite{6-On-the-effectiveness-of-parameter-efficient-fine-tuning}, Child-tuning \cite{54-raise-child} \\
\cline{2-3}
 & Structural Masking & S-Diff pruning, S-BitFit, FAR \cite{53-FAR}, BitFit \cite{1-BitFit}, Xattn Tuning \cite{gheini2021cross}, SPT \cite{SPT} \\
\hline
\multirow{4}{2.5cm}{Reparameterized Fine-tuning} & Low-rank Decomposition & Intrinsic SAID \cite{intrinsic-SAID}, LoRA \cite{5-LoRA}, Compacter \cite{2-Compacter}, KronA \cite{43-krona}, KAdaptation \cite{44-he2023parameter}, HiWi, VeRA \cite{kopiczko2023vera}, DoRA \cite{liu2024dora} \newline Dynamic Rank: DyLoRA \cite{valipour2022dylora}, AdaLoRA \cite{adalora}, SoRA \cite{SORA}, CapaBoost \cite{haobo2023increasing}, AutoLoRA \cite{zhang2024autolora} \\
\cline{2-3}
 & LoRA Derivatives & LoRA Improvement: Laplace-LoRA \cite{yang2023bayesian}, LoRA Dropout \cite{lin2024lora}, PeriodicLoRA \cite{meng2024periodiclora}, LoRA+ \cite{hayou2024lora+}, LongLoRA \cite{chen2024longloraefficientfinetuninglongcontext}
 \newline Multiple LoRA: LoRAHub \cite{huang2023lorahub}, MOELoRA \cite{liu2023moelora}, MoLoRA, MoLA \cite{feng2024mixture}, MoLE \cite{wu2024mixture}, MixLoRA \cite{li2024mixlora} \\
\hline
Hybrid Fine-tuning & \centering{-} & UniPELT \cite{15-uni-pelt}, S4 \cite{7-PEFT-design-space}, MAM Adapter, NOAH \cite{noah}, AUTOPET \cite{zhong2022autopet}, LLM-Adapters \cite{hu2023llm}, S³PET \cite{hu2022sparse} \\
\hline
\end{tabular}
\caption{Parameter Efficient Fine-tuning methods for Pre-trained Language Models}
\label{tab:PEFT}
\end{table*}

\subsection{Low-Rank Adaptation in LLMs}
Low-Rank Adaptation (LoRA) \cite{hu2021loralowrankadaptationlarge}, and specifically Low-Rank-parametrized Update Matrices, provides an efficient strategy for fine-tuning pre-trained Transformer-based language models. This technique is articulated by the following update rule:
\begin{equation}
    \Delta W = BA
\end{equation}
where $W_0 \in \mathbb{R}^{d \times k}$ is the original weight matrix, and $\Delta W$ is the low-rank update represented by matrices $B \in \mathbb{R}^{d \times r}$ and $A \in \mathbb{R}^{r \times k}$, with the rank $r \ll \min(d, k)$. During adaptation, $W_0$ remains unchanged, and only $B$ and $A$ are trained.

LoRA posits a reduction in the number of trainable parameters, significantly reducing computational overhead. It also generalizes full fine-tuning, theoretically allowing a model to approximate the expressiveness of full-rank weight matrices by selecting an appropriate $r$. For new tasks, one can quickly adapt the base model $W_0$ by adjusting $BA$, thus avoiding additional inference latency.

Furthermore, LoRA introduces a scaling parameter $\alpha$ in the adaptation step:
\begin{equation}
    \Delta W x = \alpha \frac{\Delta W x}{r}
\end{equation}
which helps maintain stability in the learning rate when varying the rank $r$, reducing the necessity for hyperparameter retuning. This parameter-efficient method for adapting LLMs presents an avenue for tailoring models to specialized domains or tasks without forfeiting their original capabilities.

\subsection{Continual Learning}
Continual Learning (CL) with PEFT for LLMs is an approach that focuses on adapting a model to new tasks over time while avoiding catastrophic forgetting of previously learned information. It leverages PEFT methods to introduce minimal, task-specific updates to the model's parameters. Techniques such as AdapterCL use residual adapters to encapsulate new knowledge for each task. These strategies help maintain the model's performance across a sequence of tasks by incorporating mechanisms like entropy-based classifiers for adapter selection, and by employing strategies to ensure knowledge transfer between tasks. The goal is to achieve a balance where the model continually accumulates and refines knowledge without substantial loss of prior learning.

\subsection{Context Window Extension}
Context Window Extension in PEFT refers to the adaptation of LLMs to process input sequences that exceed their initially defined context lengths. Through PEFT, such as LongLoRA \cite{chen2024longloraefficientfinetuninglongcontext}, LLMs can be efficiently fine-tuned to extend their context windows, allowing them to handle longer input sequences without a significant increase in computational requirements. This is particularly useful for tasks where the ability to maintain longer context is crucial for performance. LongLoRA and similar techniques modify attention mechanisms and introduce sparse attention patterns to manage longer sequences, enhancing the model's applicability to real-world scenarios with lengthy textual data.

\subsection{Visual Instruction Tuning}
One notable PEFT technique is visual instruction tuning, where LLMs, traditionally text-based, are adapted to handle visual inputs, enabling them to perform tasks like image captioning and visual question answering. The integration of visual and language processing in LLMs through visual instruction tuning represents a significant leap in multimodal AI capabilities. The process involves using LLMs like GPT-4 to generate language-image instruction-following data, which is then used to fine-tune a model capable of understanding and interacting with both textual and visual inputs. The resulting model, dubbed LLaVA (Large Language and Vision Assistant) \cite{liu2023visual}, showcases impressive multimodal conversational abilities and has set new benchmarks in accuracy for tasks such as Science QA \cite{lu2022learnexplainmultimodalreasoning}. This approach underscores the potential of LLMs in general-purpose visual and language understanding tasks. Alternatively, PEFT approaches such as adapter modules are employed to refine models like VL-BART \cite{sung2022vladapter} for image-text tasks more efficiently.

These methods represent only a fraction of the PEFT techniques used to adapt LLMs for specialized applications, highlighting the field's adaptability and ongoing innovation. Table \ref{tab:PEFT} offers an in-depth categorization of these PEFT methods for LLMs, showcasing their variety and application potential in NLP.

\begin{figure*}[!ht]
    \centering
    \includegraphics[width=\textwidth]{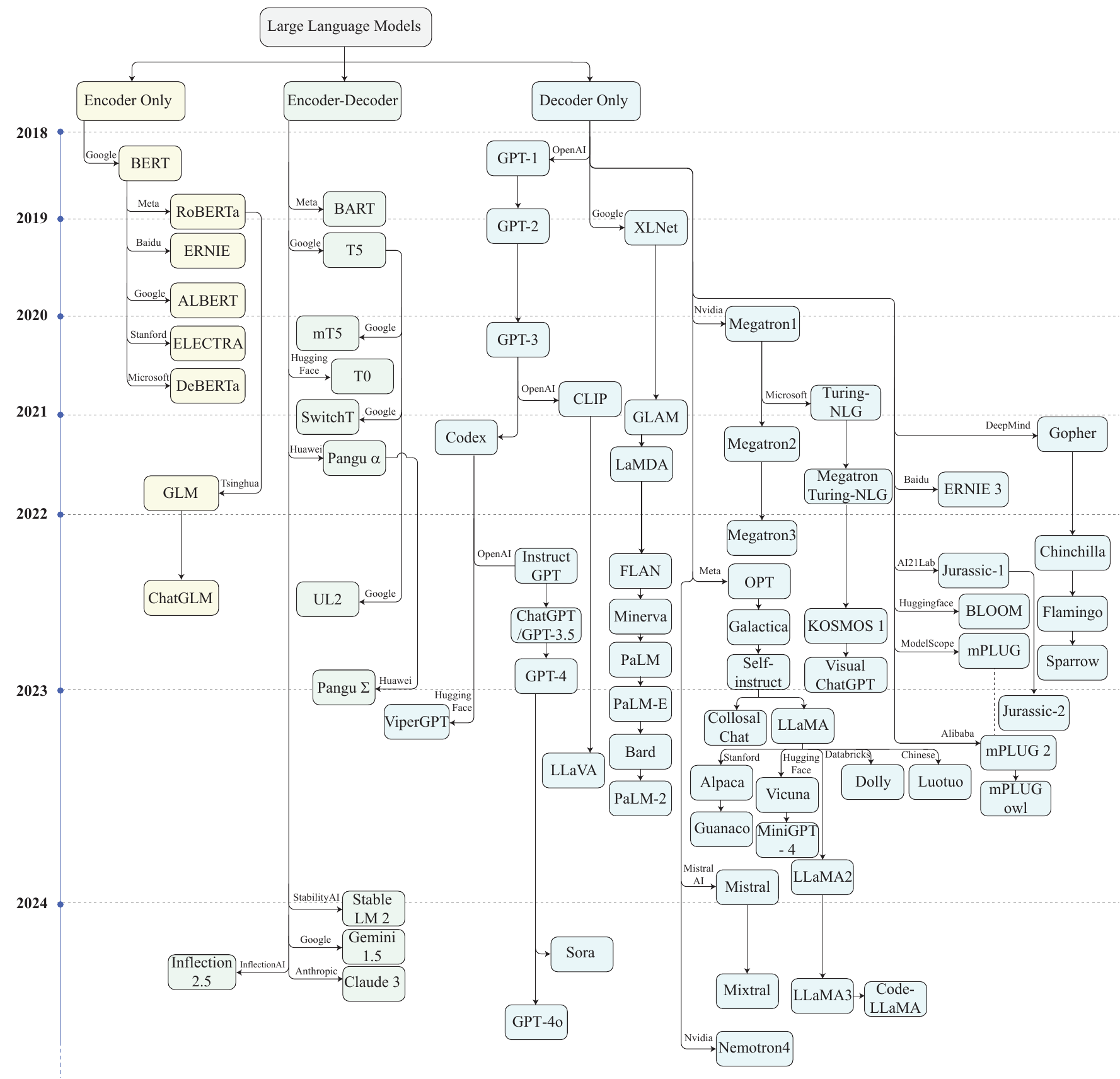} 
    \caption{An evolutionary tree illustrating the progression of mainstream LLMs. Models are categorized based on their architectural construct: encoder-only, decoder-only, and encoder-decoder. Models stemming from the same lineage or released by identical entities are interconnected with solid lines. Independent research contributions are demarcated by purple lines.}
    \label{fig:evotree}
\end{figure*}

\begin{figure*}[!ht]
    \centering
    \includegraphics[width=0.9\textwidth]{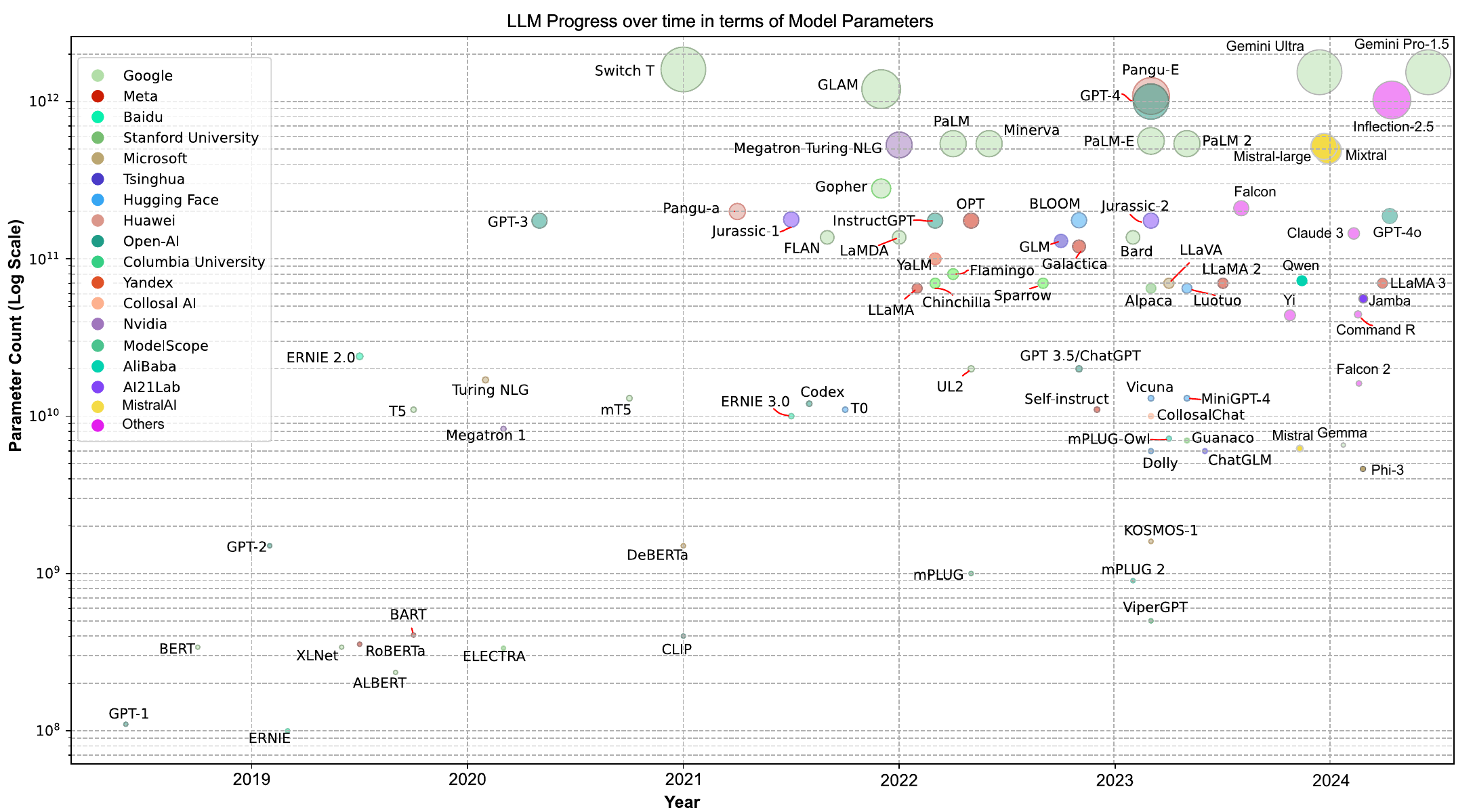} 
    \caption{Developments in LLMs and their parameter count reflecting quantitative increase in model sizes across time. The models are color coded based on their research organization.}
    \label{fig:LLMs}
\end{figure*}

\begin{figure*}[!ht]
    \centering
    \includegraphics[width=\textwidth]{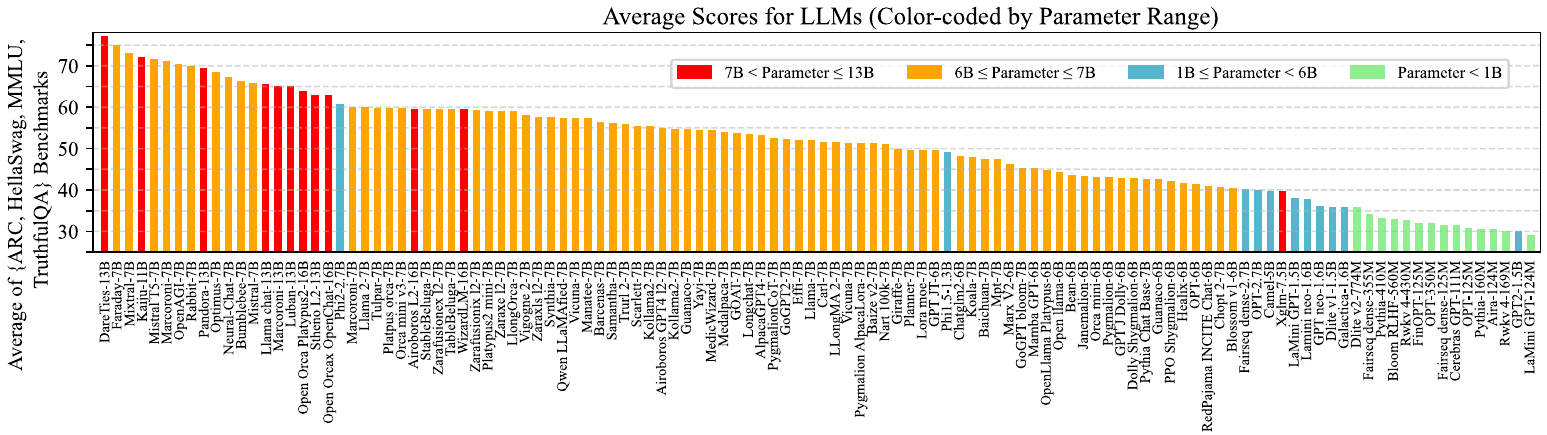} 
    \caption{Open-LLM Leader-board benchmark competing various state-of-the-art LLMs across diverse benchmarks, encompassing TruthfulQA, MMLU, ARC, and HellaSwag for a comprehensive evaluation.}
    \label{fig:LLMScore}
\end{figure*}

\section{Cutting edge LLMs\label{sec:models}}
In the following section, we provide an overview of large language models (LLMs) based on their architecture and the series they belong to, as of the date of this survey. This will offer a comprehensive understanding of the various LLMs and their respective design frameworks. Figure \ref{fig:evotree} illustrates the evolutionary tree showing the progression of LLMs across different architectures. Figure \ref{fig:LLMs} presents a plot of parameter counts for prominent LLMs, highlighting the trend of increasing parameter sizes over the years.
Similarly, Figure \ref{fig:LLMScore} displays the average scores from the Open-LLM Leaderboard for various benchmarks, including MMLU, ARC, HellaSwag, and TruthfulQA.

\subsection{Auto-Encoding Models}
Auto-encoding models, often referred to as `encoder-only models,' utilize solely the encoder component of the Transformer architecture. Their primary function is to map input data from a higher-dimensional space to a lower-dimensional vector space, effectively capturing and integrating contextual information into the data representation. These models commonly employ training strategies like masked language modeling and bi-directional training. The inception of auto-encoding models can be traced back to 2018 with the release of BERT, a pioneering model that harnessed the encoder-only architecture. This innovation significantly augmented the capabilities of natural language understanding models, enabling them to tackle a diverse range of NLU tasks, including reading comprehension, cloze tasks, and question answering.

\subsubsection{BERT}
BERT \cite{lan2019albert} stands out as one of the pioneering pre-trained models employing an auto-encoding architecture. Its training strategy primarily hinges on two tasks. The first, Masked Language Modeling (MLM), involves randomly masking tokens in the input data during the preprocessing phase. The majority of these masked tokens are hidden, prompting the model to predict them during training. However, a fraction of these tokens might be substituted with random ones. Occasionally, these replacements are erroneously embedded, compelling the model to forecast the original tokens using cross-entropy loss. It's worth noting that some of these tokens remain unaltered. This methodology equips BERT with the capability to anticipate contextual information at the token level. 

In addition to MLM, BERT introduced the Next Sentence Prediction (NSP) task to capture information at the sentence level. In the NSP task, the model determines whether an input sentence sequentially follows another within the broader context.

 For effective NSP task training, both positive and negative samples are used to enhance the model's robustness. BERT also incorporates a unique tokenization system, marking the start of a sentence with [CLS], the end with [SEP], and using the [MASK] token to obscure certain tokens during training. This approach allows BERT to generate contextually accurate and coherent language representations.

\begin{figure}
    \centering
    \includegraphics[width=0.8\columnwidth]{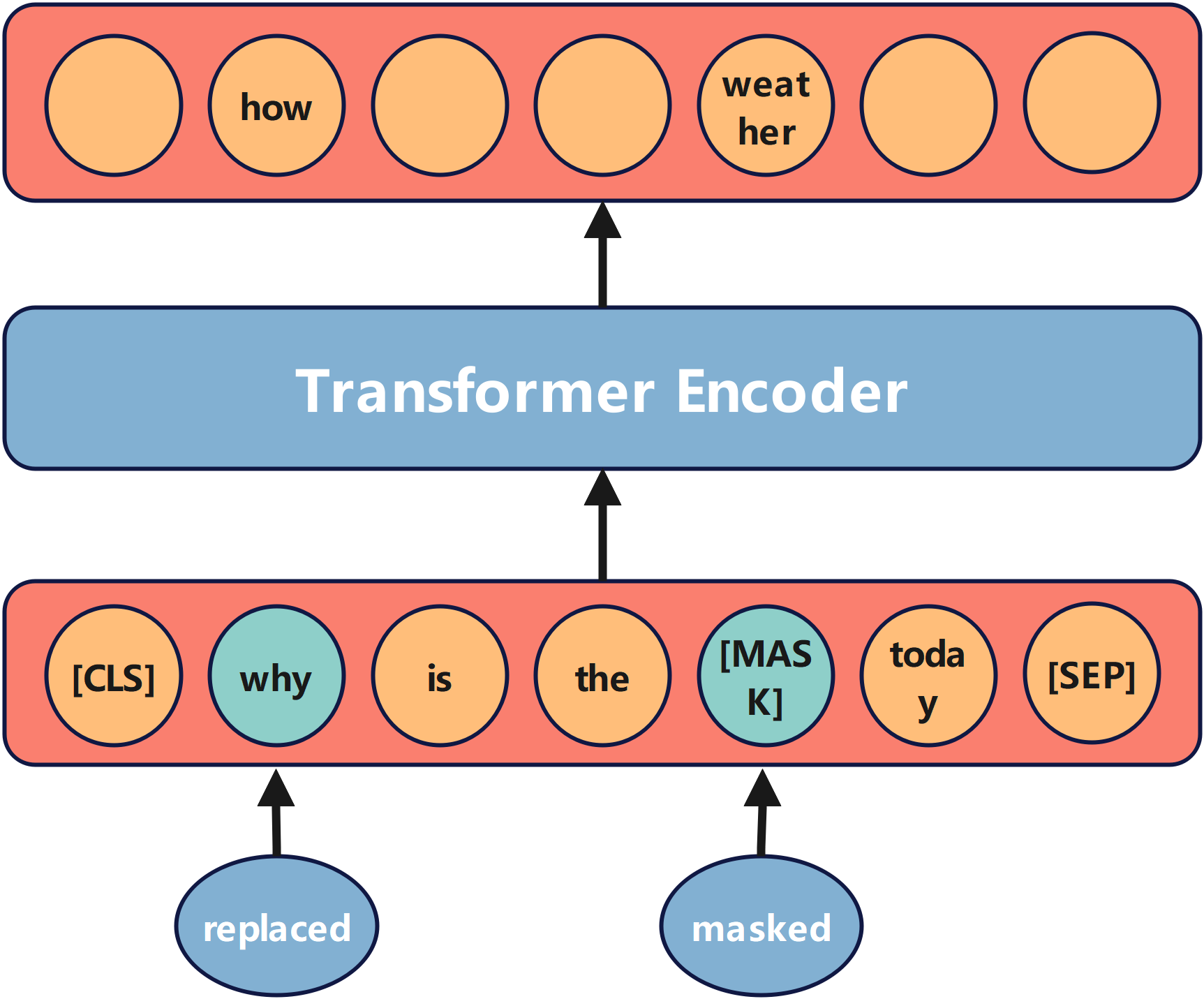} 
    \caption{Masked-Language Modeling used by BERT, in this case, the word "how" and “weather” were masked out for BERT to perform prediction tasks.
    \label{fig:bert1}}
\end{figure}
 
\subsubsection{Varients of BERT}
Several BERT variants have emerged to cater to diverse tasks.  BERT-wwm \cite{cui2021pre} employs a whole-word masking (WWM) approach for the MLM task. Contrary to the original BERT's subword-based tokenization, BERT-wwm applies masking to entire words, mitigating issues associated with subwords. BERT-wwm-ext \cite{berwwmext} extends BERT-wwm by training on more extensive datasets over increased iterations. SpanBERT \cite{joshi2020spanbert} offers an expanded MLM version where masked tokens extend to neighboring ones based on a geometric distribution with predefined randomness. SpanBERT omits the NSP task. Efficiency-enhancing adaptations of BERT have  emerged. DistillBERT \cite{sanh2019distilbert} leverages knowledge distillation to derive a streamlined BERT model with half the original's layers. It adopts RoBERTa's \cite{liu2019roberta} optimization techniques, featuring dynamic masking and enlarged batch sizes, while discarding the NSP task. TinyBERT \cite{jiao2019tinybert}  employs distillation techniques but optimizes BERT's efficiency further. VisualBERT \cite{li2019visualbert} incorporates multimodal support into BERT, pairing it with a convolutional neural network (CNN) to extract features from images. Unlike the original BERT, VisualBERT's token predictions rely on textual context and image-derived information. Lastly, MacBERT \cite{cui2020revisiting} introduces synonyms (sourced from word2vec) as MLM replacements and integrates both WWM and n-gram masking. This aims to mirror the objectives of BEiT \cite{bao2021beit}, BEiT v2 \cite{peng2022beit}, and BEiT v3 \cite{wang2022image}, which fuse a BERT-based encoder with dVAE \cite{vahdat2018dvae}.

\begin{figure}[ht]
    \centering
    \includegraphics[width=1\columnwidth]{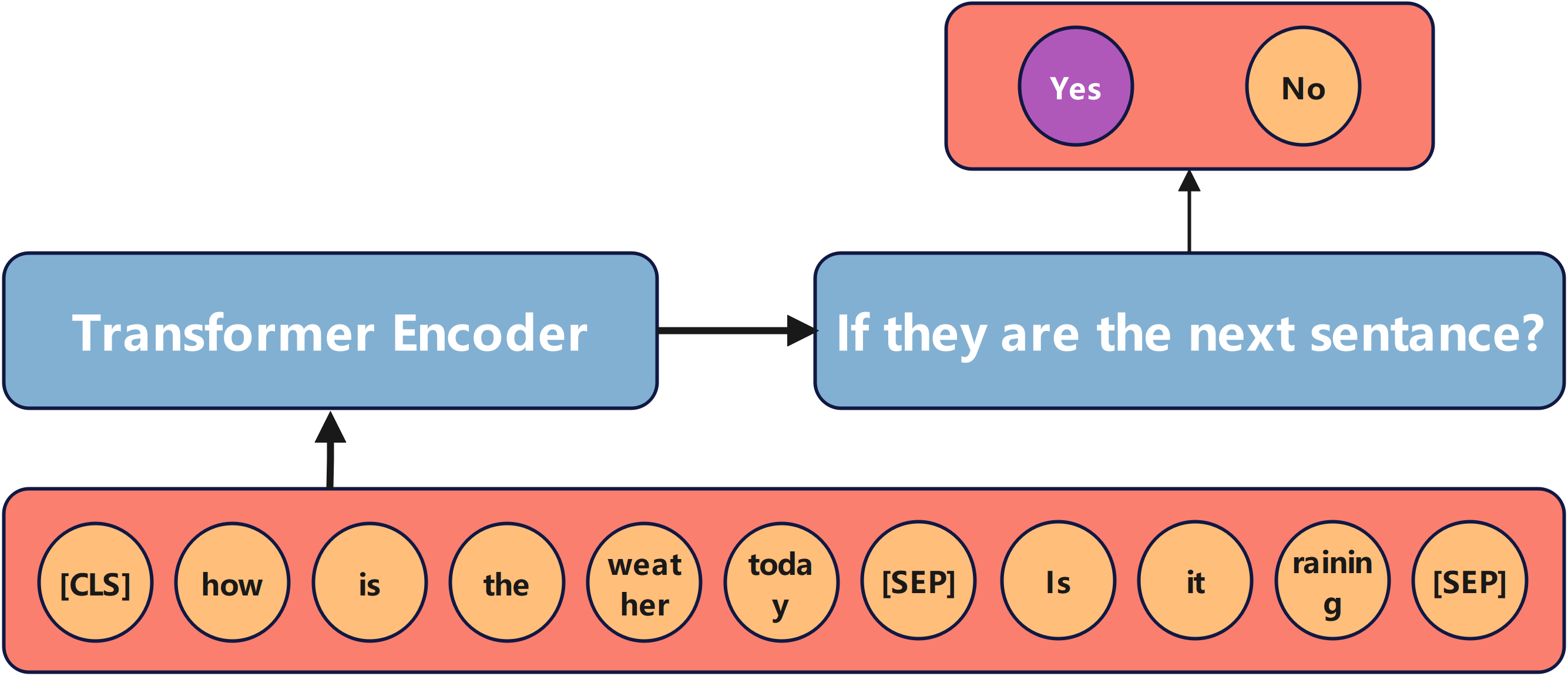} 
    \caption{Next Sentence Prediction used by BERT, the model was asked to justify if the sequence "is it raining" should be the next sentence of "how is the weather today"
    \label{fig:bert2}}
\end{figure}

\subsubsection{RoBERTa}
Several models, while distinct from BERT, owe their genesis to the original BERT framework. One such is RoBERTa \cite{liu2019roberta}, designed to enhance the robustness of BERT's training process, primarily through the introduction of a dynamic mask strategy. Contrary to BERT's MLM that employs static masks in a consistent manner for each input sample, RoBERTa's mask selection is dynamic. For any given input sequence, this sequence undergoes multiple masking processes; in their study, the authors adopted a hyper-parameter value of 10. During each masking iteration, a unique set of tokens is selected to be masked, eschewing the repetitive use of static masks. Additionally, RoBERTa's research evaluated various sentence pair configurations for the NSP tasks. Findings suggested that certain sentence-pair configurations adversely impacted the fine-tuning of downstream tasks. Consequently, to maintain document integrity during training, RoBERTa omits the NSP task. To optimize performance, RoBERTa employs larger batch sizes, an expansive training corpus, and deeper training iterations.

A notable variant of RoBERTa, termed RoBERTa-wwm \cite{xu2021roberta}, aims to fine-tune  RoBERTa for the Chinese corpus. Unique to this adaptation, tokenization and masking occur at the character, rather than word level.

\begin{figure}
    \centering
    \includegraphics[width=1\columnwidth]{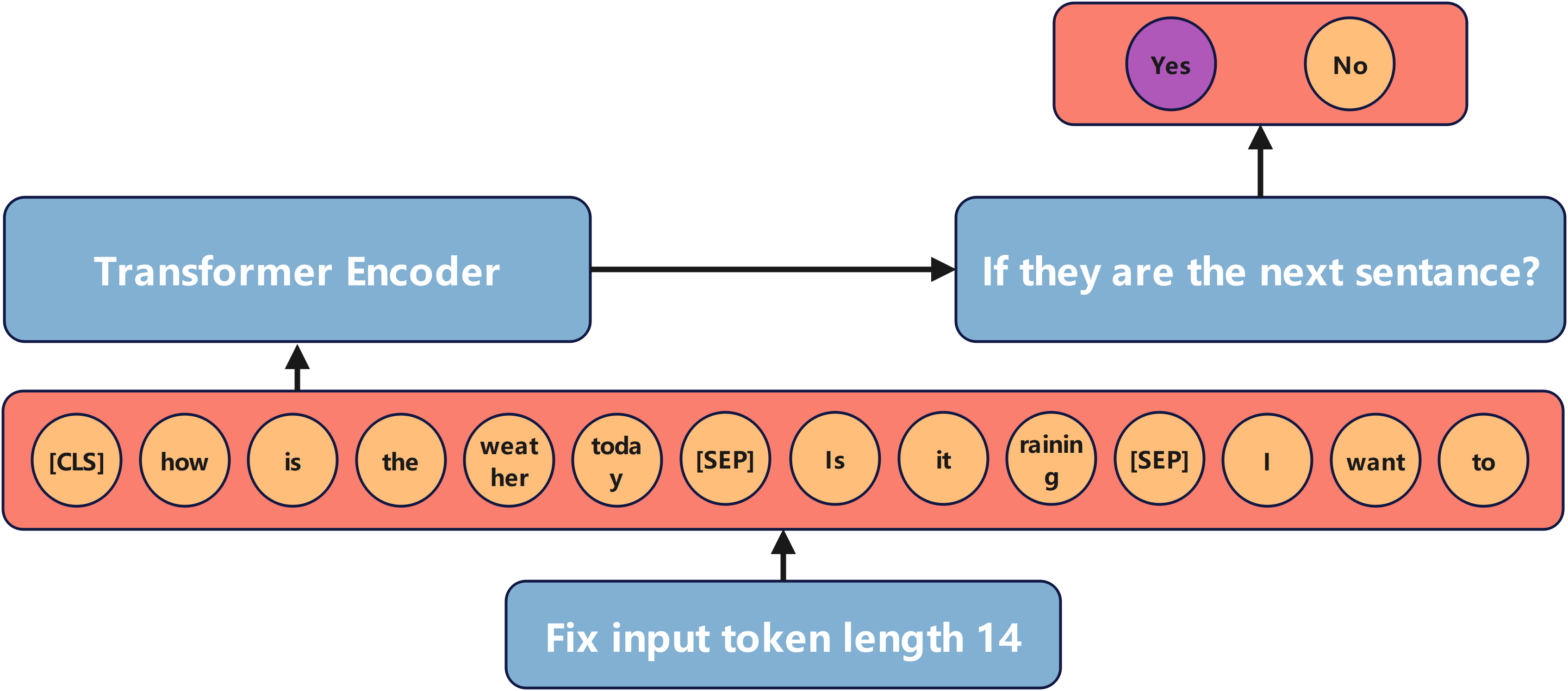} 
    \caption{changed version of NSP in RoBERTA, different from BERT which used fixed two sentences, RoBERTA would have a fixed length of the whole input sequences and give the NSP input until this fixed length was reached
    \label{fig:roberta}}
\end{figure}

\subsubsection{ERNIE}
ERNIE \cite{zhang2019ernie} employs a multi-level masking strategy distinct from BERT to optimize its performance for Chinese languages. This strategy encompasses basic-level masking, which masks out characters or words similar to BERT's MLM approach; phrase-level, where entire phrases are masked rather than individual words; and entity-level, which targets and masks entire entities within the input sequence. Additionally, ERNIE introduces the Dialogue Language Model (DLM) technique. It leverages both genuine dialogues from online forums and artificially generated ones. Within DLM, the model's training loss incorporates its capability to differentiate between authentic and fake dialogues. Like many auto-encoding model variants, ERNIE utilizes an expansive training dataset, which includes content from Chinese Wikipedia and news articles from Baidu. Architecturally, ERNIE is built upon the Transformer-XL framework.
Traditional multi-task learning, which trains tasks sequentially, often grapples with efficiency and "forgetting" challenges. To address this, ERNIE 2.0 \cite{sun2020ernie} introduces continual multi-task learning. This method deviates from the conventional sequential approach by integrating new tasks directly into the existing task set, facilitating combined training. ERNIE 2.0 also proposes three types of aware pre-training tasks. The word-aware task is an enhanced version of ERNIE's multi-level masking that incorporates token-documentation relation tasks. It predicts a token's likelihood of appearing in segments and its case (uppercase or lowercase). The structure-aware task involves the permutation of sentences into sub-sentences, predicting their original order and establishing if sentences are adjacent. Lastly, the semantic-aware task determines relationships between sentences, deducing their cohesive structure and relationships between queries and titles at the information relevance level.
ERNIE 3.0 \cite{sun2021ernie} introduces both universal representation and task-specific representation. The former is utilized for natural language understanding tasks, while the latter aids in natural language generation tasks to extract contextual semantic features. A novel knowledge-aware pre-training task is also added, which uses universal knowledge-text prediction to discern the relationships between various knowledge points within the training set. Keeping pace with modern large language models, ERNIE 3.0 expanded its parameters to 10B, supported by a larger dataset.
ERNIE has a multimodal variant, ERNIE-VilG \cite{yu2021ernie}. It trains on image datasets through three foundational tasks: object prediction, where associated tokens in the text are masked in relation to image data; attribute prediction, masking attributes of randomly selected objects; and relationship prediction, which targets and predicts the relationship between two objects in an image. ERNIE-Vil 2.0 \cite{shan2022ernie} introduces multi-view contrastive learning. This approach trains image-text pairs based on various pairings, differing from ERNIE-Vil's single image-text pairing strategy, to enhance cross-modality representation.

\begin{figure}[H]
    \centering
    \includegraphics[width=1\columnwidth]{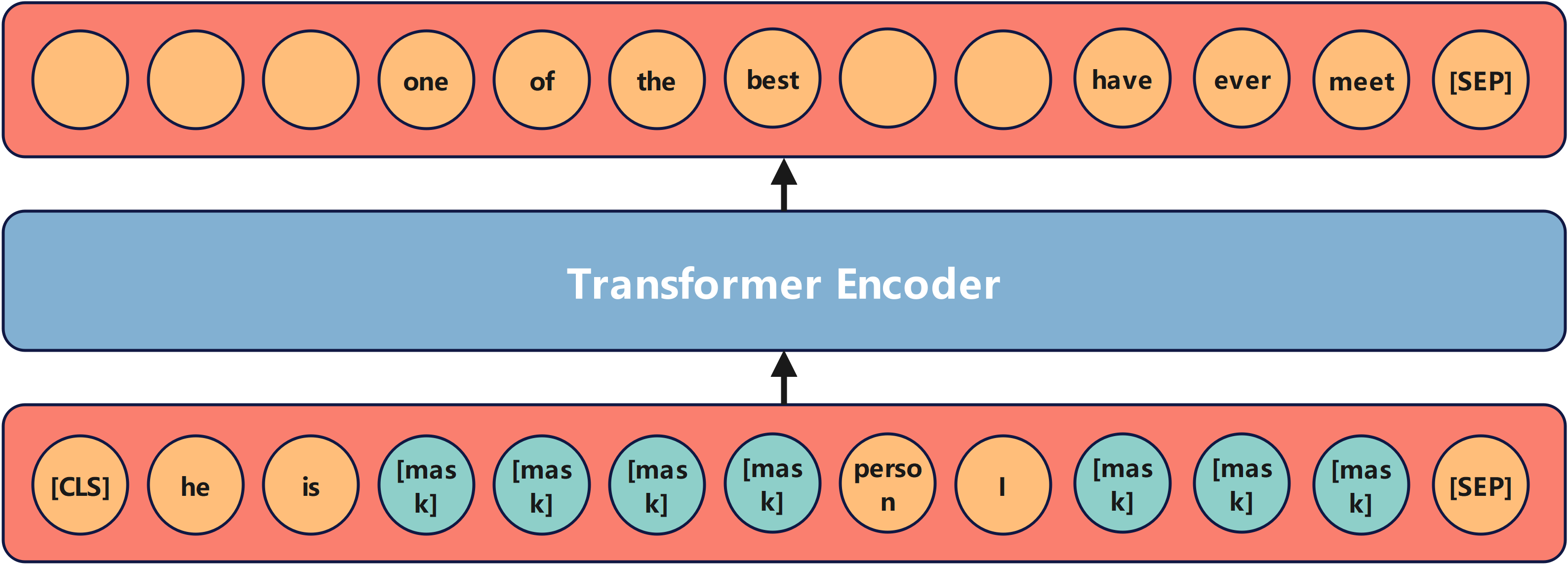} 
    \caption{ERNIE would mask out the whole related word of the chosen masked word to perform predict, this strategy could improve the capability of the model to infer the relationship between tokens
    \label{fig:ernie}}
\end{figure}

\subsubsection{ALBERT}
ALBERT \cite{lan2019albert} was developed to optimize training as model parameters grew. It observed that in earlier models like BERT, XLNet, and RoBERTa, the size of embedding layers was equivalent to that of hidden layers. These embedding layers focused on acquiring context-independent representations, while the hidden layers concentrated on context-dependent representations. Given that the models typically experienced greater enhancements from context-dependent information, it stood to reason that the dimensions of hidden layers should surpass those of embedding layers. 
To address this, ALBERT introduced the concept of factorized embedding parameterization. Instead of directly transitioning from one-hot encoding to vector embedding, this technique employs one-hot embeddings to first map to a relatively smaller embedding layer. Subsequently, this is mapped to considerably larger hidden layers. This separation ensures that the model doesn't heavily rely on the direct mapping of the vector embedding from the one-hot encoding, which can be restrictive given the differing roles of embeddings and hidden layers.
Moreover, ALBERT shared parameters between the feed-forward network and attention layers, leading to further efficiency in training. In a departure from models like RoBERTa which eliminated the NSP task outright, ALBERT evolved the task into Sentence Order Prediction (SOP). This modification expanded upon the NSP concept by transitioning from predicting the subsequent sentence to deducing the order of sentences, addressing the criticism that NSP was rudimentary for the model's potential.

\begin{figure}
    \centering
    \includegraphics[width=1\columnwidth]{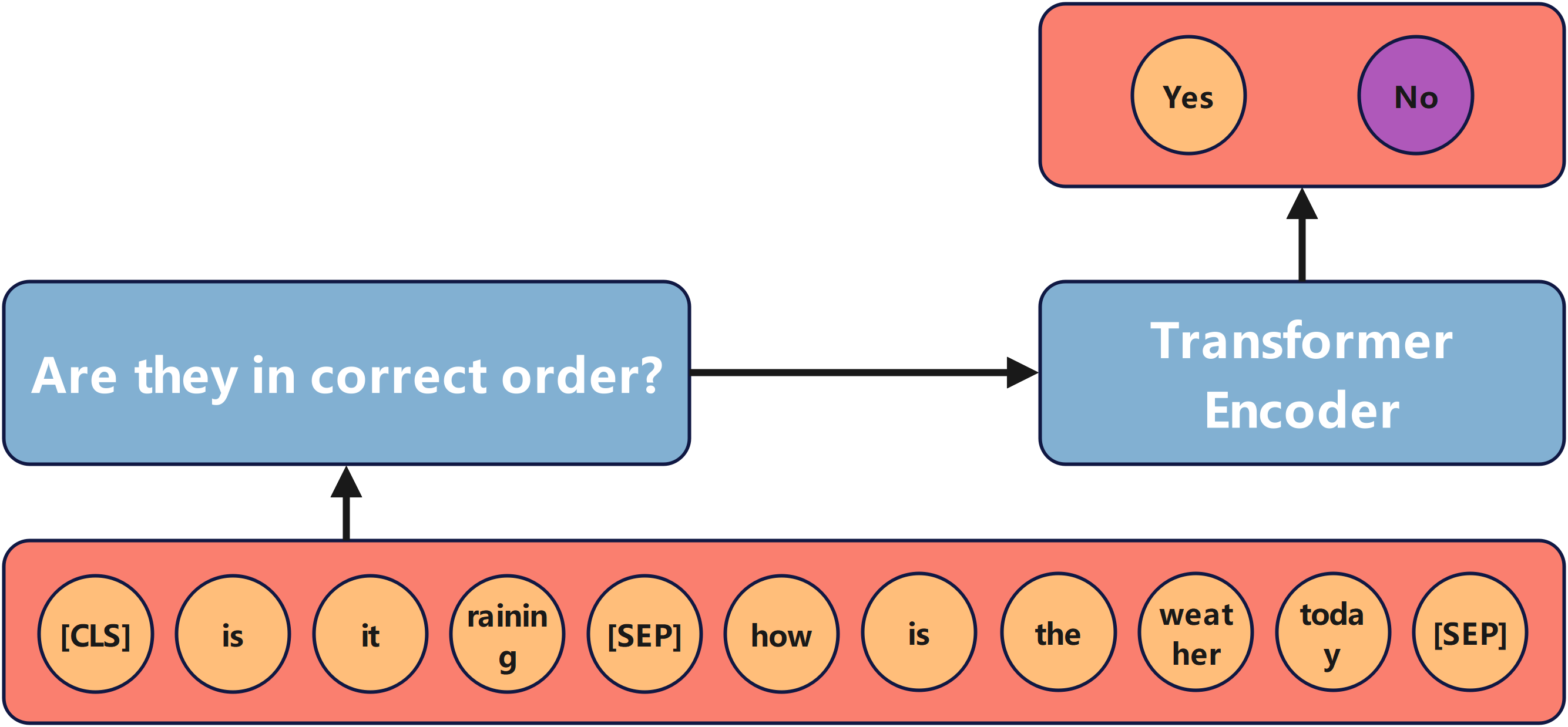} 
    \caption{ALBERT used sentence-order prediction instead of NSP, during the training, the negative sample were rotated and the model should discriminate if the sentance order was correct
    \label{fig:albert}}
\end{figure}

\subsubsection{ELECTRA}
ELECTRA \cite{clark2020electra} was introduced with the intention of effectively training auto-encoding models on smaller-scale corpora. Unlike traditional Masked Language Models (MLM) that only predict masked-out words, in ELECTRA, the objective is shifted towards predicting all words in the input sentence.
Different from the previous models which used different tasks such as masks to enhance the capability of the model during the pre-training process, the key innovation in ELECTRA is to take the feature of generative adversarial network (GAN) \cite{goodfellow2020generative} which introduced a two-part system: a generator and a discriminator. The generator, akin to the traditional MLM, randomly masks certain tokens and then attempts to predict them. The discriminator's role is to examine each token from the generator's output and determine if it is the actual token from the original input or if it's the token produced by the generator.

\begin{figure}[ht]
    \centering
    \includegraphics[width=0.8\columnwidth]{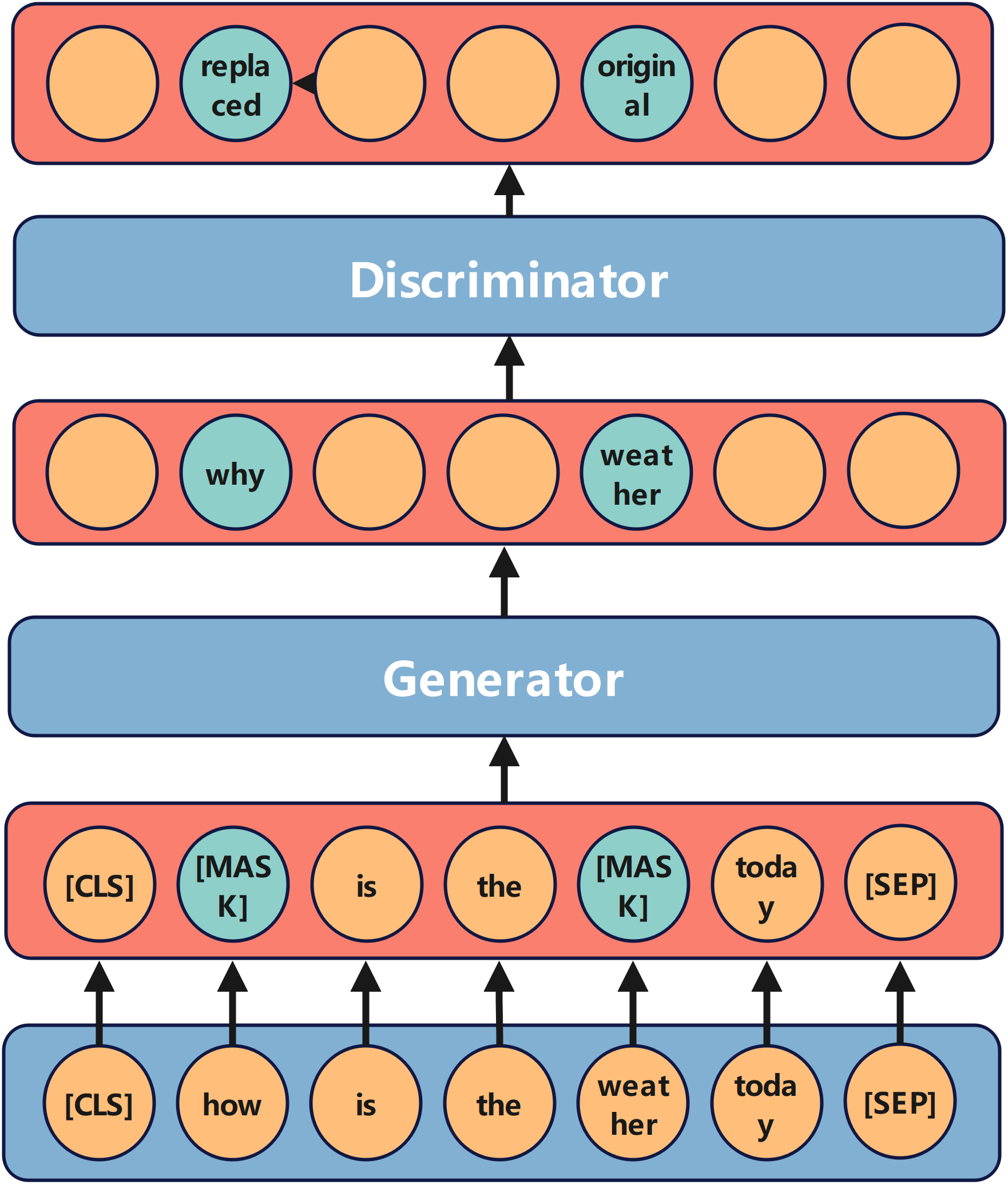} 
    \caption{ELECTRA used replaced token detection instead of MLM. In this case, the word "how" and "weather" were masked and passed the masked sentence into the generator, the generator produced the predicted masked word "why" and "weather" and used the discriminator to discriminate if the generated word matched the original word, in this case, the word "why" was discriminated as "replaced" and the word "weather" was regarded as the original correct word.
    \label{fig:electra}}
\end{figure}

By employing this adversarial training technique, the model can more effectively understand contextual information. An important detail is that the token embeddings' parameters are shared between the generator and the discriminator. This shared parameterization not only reduces the model's overall parameters but also ensures consistent representation between the two parts, enabling them to better work in tandem. The result is a model that can be trained efficiently on smaller datasets, yet achieve competitive, if not superior, performance compared to its counterparts trained on much larger corpora.

\subsubsection{DeBERTa}
In conventional auto-encoding models, the Masked Language Model (MLM) technique is typically not employed during the fine-tuning stage. This omission often leads to discrepancies between the pre-training accuracy and the fine-tuning accuracy. Furthermore, the attention score is somewhat influenced by the token positions. To address the inconsistency in accuracy and the distribution of token positions, DeBERTa \cite{he2020deberta} was introduced. This model incorporates a novel attention mechanism termed "disentangled attention," which factors in the relative positional information of tokens when computing the attention score. In addition, DeBERTa employs an enhanced mask decoder that replaces the conventional softmax function with an Extended Mask Decoder (EMD) and adds several transformer layers prior to the original softmax function. Within the Enhanced Mask Decoder, 10\% of the relative positional embedding information is substituted with absolute positional embedding, bolstering the attention across diverse positional information.

\subsubsection{Transformer-XL}
Despite of the wide usage of BERT and its derived models, handling long sequences in the standard transformer model presents challenges. Specifically, tokens are segmented into fixed-length sub-sequences. Post-encoding, the model lacks a mechanism to discern relationships between these segments. This limitation hinders the flow of contextual information, as the model cannot ascertain how the fragments relate within the entire sequence, leading to context fragmentation. Transformer-XL \cite{dai2019transformer} 
is another family of auto-encoding model that addresses this by introducing segment-level recurrence coupled with state reuse. This method integrates a memory layer to retain information from preceding sequences, facilitating the establishment of relationships between sub-sequences during encoding. Such a mechanism not only enables the model to understand semantic connections between segments but also encodes relationships in longer sequences. Moreover, Transformer-XL employs relative positional encoding, replacing absolute positional encoding to mitigate confusion between the current sub-sequence and the entire sequence.
XLNet \cite{yang2019xlnet} adopts the Transformer-XL architecture to rectify the MLM drawbacks found in BERT. The authors postulated that the introduction of masks might disrupt contextual information construction. To counteract this, XLNet introduces a suite of techniques. One key method is the permutation language model, which factorizes input sequences in varying orders. This replaces BERT's bi-directional training, enabling the model to learn contextual information bidirectionally by alternating factorization permutation orders while sharing parameters. To facilitate factorization permutation, XLNet introduces a two-stream self-attention mechanism that employs two hidden states instead of one, alleviating issues induced by permutation. Additionally, in lieu of complete predictions, XLNet employs partial prediction, akin to BERT's sub-wording, to predict only a token segment, enhancing the model's convergence speed.

\subsection{Auto-Regressive Models}
Auto-regressive models are often referred to as "decoder-only" models. In contrast, auto-encoding models compute attention bi-directionally, enhancing the model's capacity for natural language understanding by perceiving the entire context. While self-attention computes attention scores globally, assessing the correlation between each token in a sequence, auto-regressive models utilize uni-directional attention. This means that the current generation is dependent solely on previously generated sequences; tokens following the current one are masked out. Compared to auto-encoding models, this architecture excels in generation tasks. The uni-directional attention offers superior performance in handling long sequences, which is one of the reasons this architecture is widely adopted in contemporary large language models.

\subsubsection{GPT}
GPT \cite{radford2018improving} was one of the early auto-regressive models released in 2018, contemporaneously with BERT. It was a pioneer in introducing auto-regression techniques. Before GPT's inception, language model training largely relied on large corpora, often manually or automatically annotated, which were costly to assemble. Many of these earlier models were domain-specific, limiting their zero-shot capabilities. GPT uniquely adopted unsupervised learning, training generatively on unlabeled datasets and subsequently fine-tuning for specific tasks. It incorporated strategies such as natural language inference to assess relationships between sentences, question answering and commonsense reasoning for semantic comprehension, semantic similarity evaluations, and classification tasks. Despite utilizing a relatively smaller dataset, the BookCorpus \cite{bookcorpus}, it employed a 12-layer transformer encoder.

GPT-2 \cite{radford2019language}, a successor to GPT, embraced multi-task learning. Building on GPT's foundation, it postulated that with sufficiently large data and model size, supervised tasks could be implicitly learned, as indicated in decaNLP \cite{mccann2018natural}. Consequently, GPT-2 leveraged WebText, a dataset vastly larger than GPT's BookCorpus, comprising over 8 million websites from Reddit, exceeding 40GB. The architecture was also expanded to 48 layers, enlarging the parameter count to nearly 13 times that of GPT.

GPT-3 \cite{brown2020language} continued GPT-2's ethos of expanding dataset size and model depth. It adopted an in-context learning training strategy, aiding in faster convergence through the model-agnostic meta-learning method \cite{finn2017model}. The distinct fine-tuning phase, present post-pretraining in GPT and GPT-2, was eschewed in GPT-3 due to the reasons mentioned and the enormity of its dataset. GPT-3 sourced data from five diverse repositories, including Common Crawl \cite{commoncrawl}, WebText 2, two iterations of BookCorpus, and Wikipedia, aggregating over 45TB. The model architecture doubled GPT-2's, employing 96 layers of transformer decoder, culminating in a staggering 175B parameters, more than 100 times that of GPT-2.

InstructGPT \cite{ouyang2022training} melded reinforcement learning with human feedback into GPT-3's fine-tuning process. Using the SFT dataset, it established a reward model reflecting human feedback on the model's output through proximal policy optimization, enhancing the model's human-like behavior. Drawing from InstructGPT's technology, OpenAI unveiled ChatGPT \cite{chatgpt}, a fine-tuned iteration of GPT-3, colloquially termed GPT-3.5.

GPT-4's \cite{2303.08774} technical report, while not fully revealing its strategies, highlights some salient features. GPT-4 can handle both image and text inputs, producing textual outputs, classifying it as a genuine MLLM. This flexibility broadens GPT-4's task repertoire, especially for multimodal requirements. While ChatGPT, based on GPT-3.5, supports up to 4096 tokens (roughly 3000 English words), GPT-4 manages a remarkable 32767 tokens, or about 25000 words, enabling the processing of more extended text sequences with heightened accuracy. GPT-4 emphasizes security, addressing challenges like Adversarial Usage, Unwanted Content, and Privacy Concerns through strategies like RLHF (reinforcement learning with human feedback), real-world use case simulations, and an adversarial testing program. Additionally, GPT-4 allows users to use precise prompts, granting more control over the model's behavior.

Several models have been derived from or inspired by the GPT series. GPT-Neo \cite{gptneo} seeks to offer an open-sourced version of GPT-3 for local deployment. GPT-GNN \cite{hu2020gpt} integrates pre-training on graph neural networks to understand node interrelations. GPT-J \cite{gpt-j} is GPT-inspired, grounded on the Mesh-transformer-JAX \cite{Huebner_2023}. GPT-NeoX \cite{black2022gpt} succeeds GPT-Neo, integrating parallel computing from GPT-J using the Deepspeed \cite{deepspeed} framework. DialoGPT \cite{zhang2019dialogpt}, an extension of GPT-2, aims to refine text generation using maximum mutual information to elevate hypothesis ranking.

\begin{figure*}[H]
    \centering
    \includegraphics[width=2\columnwidth]{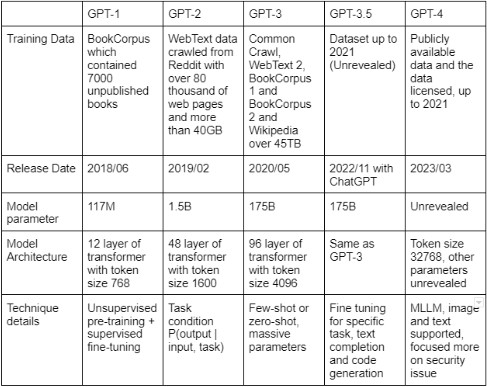} 
    \caption{The comparison between GPT series models
    \label{fig:gptevo}}
\end{figure*}

\subsubsection{Pathways and PaLM}
The "Pathways" architecture, as introduced by Google \cite{Dean_2021}, represents a significant shift in AI design. Instead of the traditional method of constructing distinct models for individual tasks, Pathways proposes a single AI system capable of generalizing over thousands, if not millions, of tasks. This versatility is attributed to the "mixture-of-experts" (MoE) concept. By constructing a single model that can be trained on extensive datasets encompassing both text and code, different tasks or inputs are managed using a gate function, directed by the respective experts. A notable feature of models built using the Pathways architecture is their ability to be fine-tuned for specific tasks through few-shot learning. This approach empowers the model to master a new task using only a handful of examples, ensuring efficiency, versatility, and precision in its applications.
PaLM \cite{palm} is the first language model trained by Pathways architecture which is a model contains up to 540B number of parameters due to the sparsity architecture of Pathways. PaLM used a variant version of traditional transformer decoder which made the modification below:
\begin{enumerate}
    \item used SwiGLU \cite{shazeer2020glu} activation function
    \item used multi-query attention, sharing all parameters between all heads of multi-head attention to boost the efficiency of decoding
    \item put the feed forward network in parallel with attention layers
    \item used RoPE \cite{su2021roformer} positional embedding
    \item Remove all bias in the neural network
    \item sharing the embedding among the input and output
\end{enumerate}

The "Pathways" architecture's utilization of few-shot learning, as detailed in PaLM, yielded state-of-the-art outcomes closely paralleling human performance across four tasks. Moreover, the introduction of the BIG-bench, encompassing 150 tasks, further highlighted its prowess, especially in comparison to models like GPT-3.

The Embodied Pathways Language Model, or PaLM-E \cite{driess2023palm}, is an evolution of the preceding PaLM. It distinguishes itself as the premier Large Language Model (LLM) developed using Google's Pathways architecture. This architecture was crafted to facilitate sparse activations across multimodal and multi-task scenarios, implying that for a distinct task, only a section of the model activates, curtailing computational intricacy. PaLM-E, much like Microsoft's KOSMOS-1, is adept at deciphering both natural language and imagery, and boasts potential applications in robotics for accomplishing embodied tasks. Training for this model is conducted on the expansive GEM dataset, encompassing both textual and visual data sources like ImageNet \cite{5206848}, COCO \cite{lin2014microsoft}, and Visual Genome \cite{krishna2017visual}. PaLM-E integrates two primary components: the language pathway, rooted in the Transformer-XL framework, and the visual pathway, modeled after the Vision Transformer (ViT) framework. A fusion module bridges these pathways, enabling seamless integration of both modalities. Its efficacy is demonstrated through exceptional performances across a gamut of tasks such as image captioning, visual question answering, and embodied navigation. Architecturally, PaLM-E synergizes two distinct models: the PaLM, which serves as a textual decoder, and the ViT 22B \cite{dehghani2023scaling}, a dedicated visual transformer. The PaLM delivers linguistic processing proficiency, while the ViT 22B specializes in image processing. The default model amalgamates a 540B PaLM model and a 22B ViT model, cumulating to a massive 562B parameters. In terms of training data, PaLM-E benefits from a staggering 780B tokens, derived from diverse sources like social media, web content, Wikipedia, and GitHub. Meanwhile, the ViT 22B module trains on the JFT dataset \cite{mehta2022large}, encompassing roughly 4B semi-automatically annotated images. The model's input representation strategy mirrors KOSMOS-1 \cite{huang2023language}, where visual or robotic-related inputs receive specific tags, and distinct encoding techniques interpret the content within these tags. Directly, textual data is channeled into the language model. Additionally, the Minerva model \cite{lewkowycz2022solving}, fine-tuned from a vast 118G dataset teeming with scientific publications and mathematical expressions, is predicated on PaLM and targets challenges in scientific and mathematical domains. Lastly, PaLi \cite{chen2022pali}, another PaLM derivative, tackles text-vision quandaries using an image-and-text to text-only paradigm, incorporating models like mT5-XXL \cite{xue2020mt5}, ViT-G \cite{zhai2022scaling}, and ViT-e \cite{dosovitskiy2020transformers}.

\begin{figure}
    \centering
    \includegraphics[width=0.7\columnwidth]{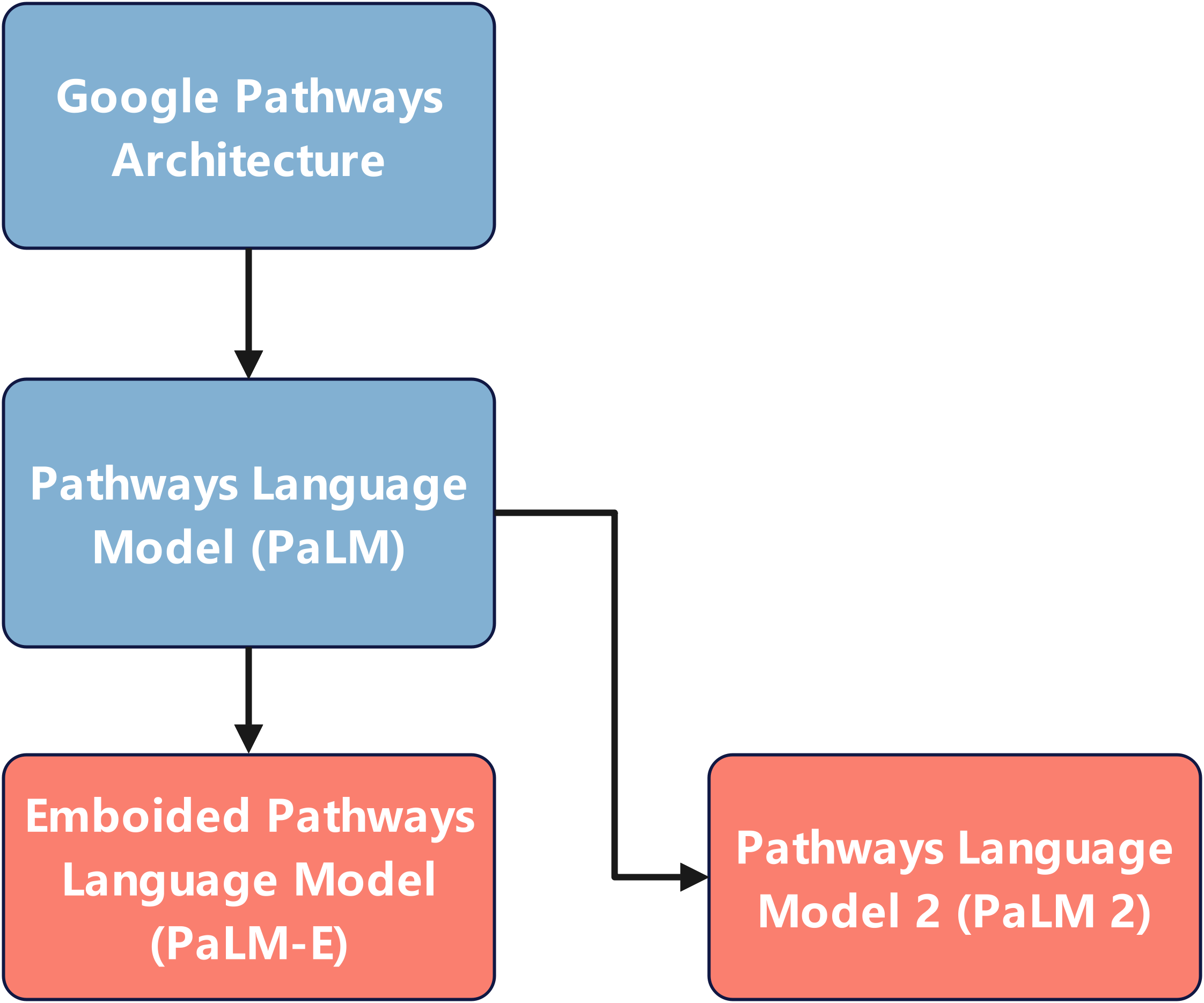} 
    \caption{The relationship between models under PaLM family by Google
    \label{fig:palme-branch}}
\end{figure}

\begin{figure}
    \centering
    \includegraphics[width=1\columnwidth]{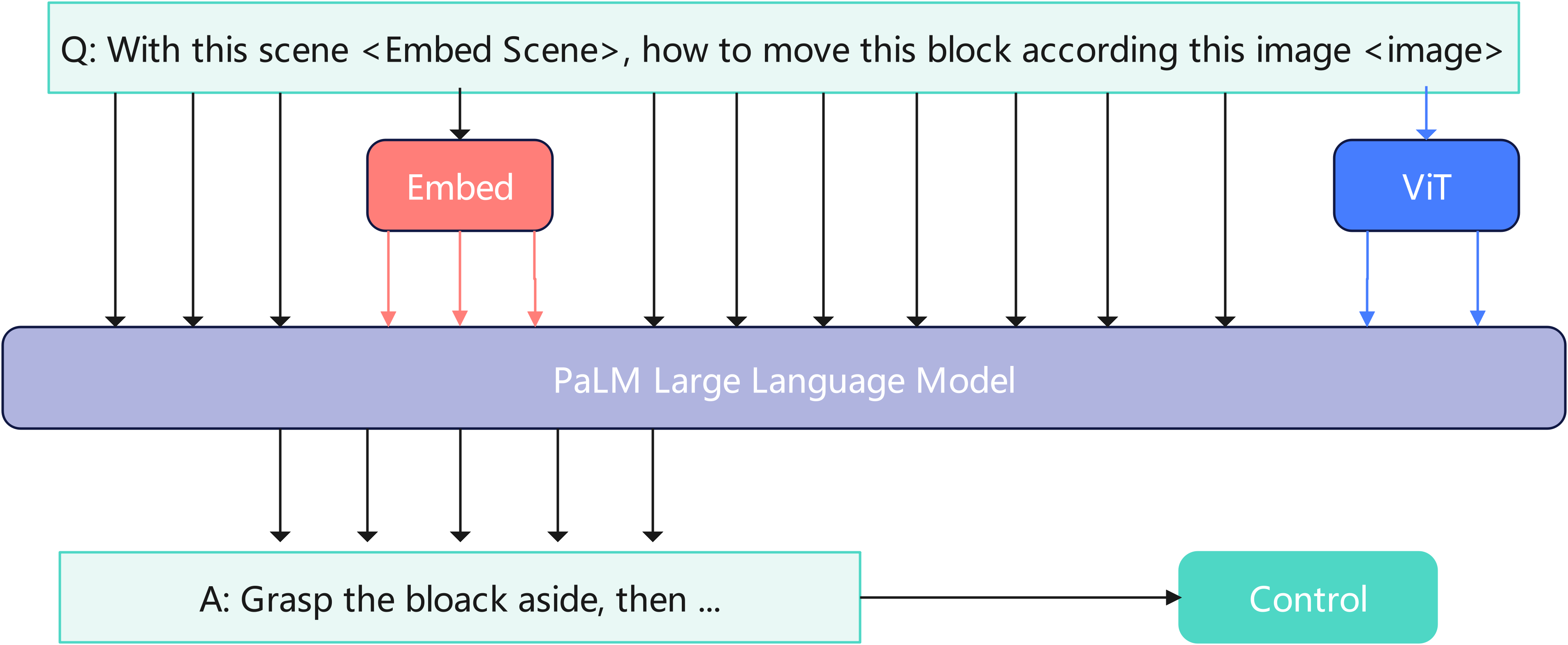} 
    \caption{Architecture of PaLM-E, first the visual information would go pass the ViT, and the scene would be embedded into vector format, then all the token embedding would go through a PaLM model to generate the response text, that response text would then be transferred to the control signal
    \label{fig:palme}}
\end{figure}

The PaLM-2 model \cite{anil2023palm} augments the capabilities of its predecessor, the PaLM, with enhanced performance in mathematics, coding, inference, and multi-language tasks. Leveraging JAX and TPU technology, the primary objective of PaLM-2 is to amplify the efficacy of the PaLM model while simultaneously utilizing fewer parameters. In contrast to PaLM-E, PaLM-2 does not offer multi-modal support. The PaLM-2 architecture encompasses four variants differentiated by parameter size: Gecko, Otter, Bison, and Unicorn. These versions ensure versatility, catering to varied deployment needs across diverse platforms. Notably, the Gecko model, being the most compact, is optimized for deployment on mobile devices. In the realm of medical question-answering (QA), the Med-PaLM 2 \cite{singhal2023towards} was introduced. This specialized version, derived from PaLM-2, employs fine-tuning techniques centered on the ensemble refinement approach applied to medical datasets. Impressively, it registered an accuracy rate of 86.5\% on the MedQA benchmark \cite{jin2021disease}, marking a significant improvement from its antecedent, the Med-PaLM \cite{singhal2022large}.

\begin{figure*}
    \centering
    \includegraphics[width=1.7\columnwidth]{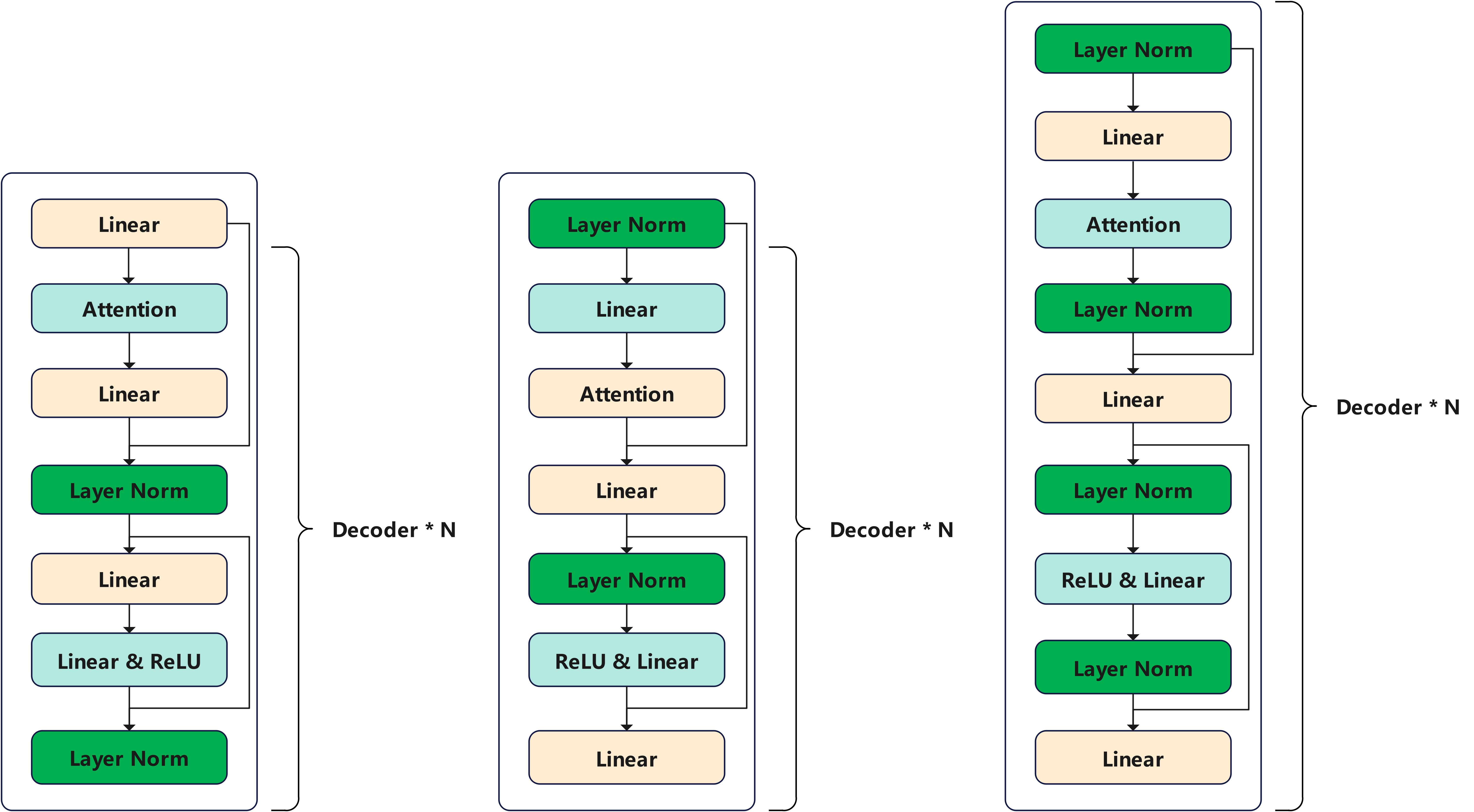} 
    \caption{Structure of the foundation transformer used in KOSMOS-1
    \label{fig:lntrans}}
\end{figure*}
\subsubsection{Microsoft KOSMOS-1}

The KOSMOS-1 model \cite{huang2023language} is underpinned by the magneto transformer \cite{wang2022foundation}, a derivative of the original transformer architecture. The primary distinction between the two lies in the magneto transformer's integration of a layer normalization between its linear layer and the activation function, a modification that enhances training stability and scalability. Specifically designed with 24 decoder layers, each having 2048 dimensions, 8192 hidden size, and 32 attention heads, KOSMOS-1 boasts approximately 1.6 billion parameters.

Rather than directly processing images, KOSMOS-1 expedites training by harnessing the CLIP ViT-L/14 model to capture image features with 1,024 dimensions. Moreover, it employs the XPOS technique, leveraging length-extrapolation to reconcile disparities in length between training tokens and predicted tokens. These innovations equip KOSMOS-1 with the versatility to excel in a range of tasks, such as image captioning and visual question answering, underscoring its prowess in multimodal learning.

KOSMOS-1 is enriched by three primary datasets. The text corpus includes The Pile, a vast English text dataset tailored for LLMs, and other resources like Common Crawl, CC-Stories, and RealNews. Image-caption pairings are sourced from datasets such as English LAION-2B, LAION-400M, COYO-700M, and Conceptual Captions, all of which were acquired through web crawling. Additionally, there's an interleaved image-text compilation featuring combined image and text fragments. This data is culled from an initial collection of 2 billion web pages, which was subsequently condensed to 71 million pages in the finalized dataset.

For data preprocessing, text sequences were designated with the <s> tag, while images received the <image> label. Notably, while KOSMOS-1 also extends support to audio sequences, the associated paper lacks comprehensive details on this aspect, hinting at its potential developmental stage.

Conclusively, KOSMOS-1 signifies a monumental stride in the evolution of multimodal large language models, poised to have transformative impacts on both natural language processing and computer vision research arenas.

\begin{figure*}[H]
    \centering
    \includegraphics[width=2\columnwidth]{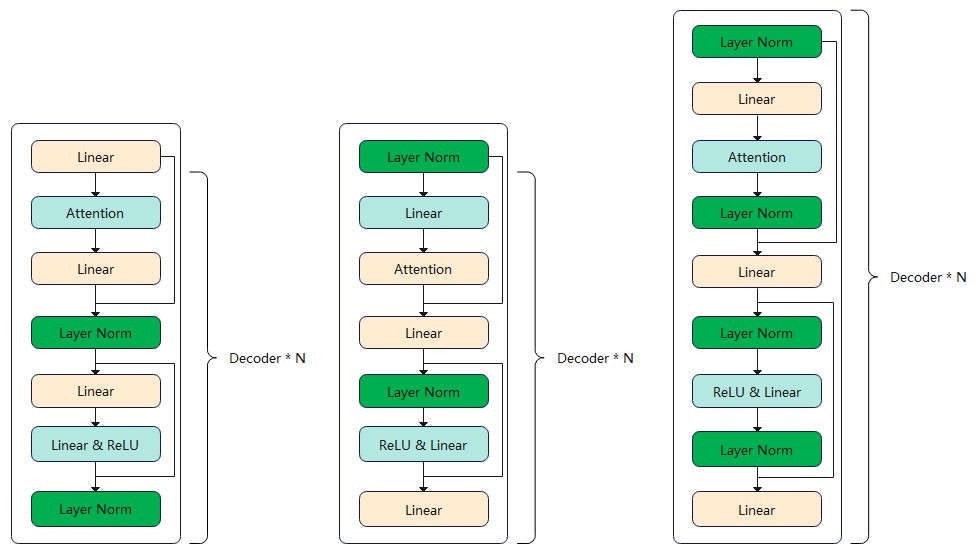} 
    \caption{Left: Post Layer Norm, the LN layers were applied after the attention and activation function; Middle: Pre Layer Norm, the LN layers were applied before the attention and activation function; Right: Sub Layer Norm proposed by Microsoft, which was the transformer architecture used in KOSMOS-1, in addition to the Pre Layer Norm, an extra layer norm was applied inside the decoder
    \label{fig:mstrans}}
\end{figure*}

\subsubsection{Megatron}
Nvidia Megatron \cite{megatron} is a framework proposed by Nvidia to solve the parallel computing of the LLM training, it mainly used two strategies to boost the model training and relief the shortage of VRAM during the training process. Inter-layer parallel, also known as tensor parallel, aimed to divide a single model into several layers and deploy each layer into separate devices, while intra-layer parallel, also known as tensor parallel, divided a single model into multiple layers to deploy on different devices. The final output would be concatenated from these model segments after processing directly. Data parallel, divides the whole dataset into different devices while each device still holds a full copy of the model to boost the training process of the model.
Megatron LM \cite{shoeybi2019megatron} is a LLM trained with Megatron framework with 8.3B parameters and trained distributively on multi-GPU with both tensor parallel and data parallel strategies. While its successor, Megatron LM 2 \cite{narayanan2021efficient}, introduced a strategy called PTD-P which combined all of the tensor parallel, pipeline parallel and data parallel to train the model which it claimed, is able to train a model over 1T size with IO 502 PTFLOG/S. Megatron LM 3 \cite{korthikanti2023reducing} introduced sequence parallel which divide the non-tensor part of the model along the sequence dimensions and also introduced a reduce-scatter operation which reduce the VRAM required for the activation function.
Turing NLG \cite{turingnlg} was an early LLM released by Microsoft which aimed to solve the similar parallel computing problem as Megatron models, which used DeepSpeed \cite{deepspeed} and ZeRO \cite{rajbhandari2020zero} to boost the training based on distributed computing which is a breakthrough of both hardware and software with reduced training time around $2/3$. Turing NLG contains 78 layers of transformer decoder and 170B parameter which is the largest model when it was released following by Megatron LM.
ChatGPT
Nvidia's Megatron \cite{megatron} is a framework specifically designed to address the challenges associated with the parallel computing of large language model (LLM) training. The architecture primarily incorporates two strategies to expedite model training and alleviate the constraints of VRAM during the training phase. Firstly, the intra-layer parallelism, also known as tensor parallelism, partitions the layers of a single model into fragments that are then distributed across different devices. Once these individual segments process the data, their outputs are concatenated to produce the final result. In contrast, inter-layer parallelism involves segmenting a model into distinct layers, with each layer being allocated to a separate device. Another strategy, data parallelism, involves distributing the dataset across multiple devices, but each device retains a full copy of the model, thereby accelerating the model's training process.

Megatron LM \cite{shoeybi2019megatron}, a large language model nurtured using the Megatron framework, boasts 8.3B parameters. The model is trained distributively across multiple GPUs, leveraging both tensor parallelism and data parallelism. Its successor, Megatron LM 2 \cite{narayanan2021efficient}, integrated a method termed PTD-P, which synergistically combines tensor parallelism, pipeline parallelism, and data parallelism. This integrated approach empowers the framework to train models exceeding 1T in size, achieving an impressive IO of 502 PTFLOG/S. Subsequently, Megatron LM 3 \cite{korthikanti2023reducing} incorporated sequence parallelism, which segments the non-tensor components of the model along the sequence dimensions. This edition also introduced a "reduce-scatter" operation, significantly diminishing the VRAM demands of the activation function.

On a parallel trajectory, Microsoft's Turing NLG \cite{turingnlg} emerged as an early contender in the LLM space, targeting parallel computing challenges akin to those addressed by the Megatron models. Turing NLG integrates the capabilities of DeepSpeed \cite{deepspeed} and ZeRO \cite{rajbhandari2020zero}, pioneering advancements in both hardware and software realms. This integration has culminated in a substantial reduction in training time, approximately by two-thirds. Architecturally, Turing NLG comprises 78 transformer decoder layers and 170B parameters. At the time of its release, it held the title of the largest model in its category, subsequently succeeded by Megatron LM.

Megatron-Tuiring NLG \cite{smith2022using} is the successor of Turing NLG in combination with the Megatron architecture which contains over 530B of the parameters over the Turing NLG. It was trained based on 280 Nvidia A100 GPUs and contains 105 layers of transformer deocder. The training data was from a variety of sources shown below:

\begin{table}
    \centering
    \footnotesize
    \caption{Training dataset of Megatron-Turing NLG}
    \begin{tabular}{|m{2.5cm}|m{1.5cm}|m{1.5cm}|m{1.2cm}|}
    \hline
    \textbf{Dataset} & \textbf{Tokens (B)} & \textbf{Weights} (\%) & \textbf{Epochs} \\ \hline
    Books3 & 25.7 & 14.3 & 1.5 \\ \hline
    OpenWebText2 & 14.8 & 19.3 & 3.6 \\ \hline
    Stack Exchange & 11.6 & 5.7 & 1.4 \\ \hline
    PubMed Abstracts & 4.4 & 2.9 & 1.8 \\ \hline
    Wikipedia & 4.2 & 4.8 & 3.2 \\ \hline
    Gutenberg (PG-19) & 2.7 & 0.9 & 0.9 \\ \hline
    BookCorpus2 & 1.5 & 1.0 & 1.8 \\ \hline
    NIH ExPorter & 0.3 & 0.2 & 1.8 \\ \hline
    Pile-CC & 49.8 & 9.4 & 0.5 \\ \hline
    ArXiv & 20.8 & 1.4 & 0.2 \\ \hline
    GitHub & 24.3 & 1.6 & 0.2 \\ \hline
    CC-2020-50 & 68.7 & 13.0 & 0.5 \\ \hline
    CC-2021-04 & 82.6 & 15.7 & 0.5 \\ \hline
    RealNews & 21.9 & 9.0 & 1.1 \\ \hline
    CC-Stories & 5.3 & 0.9 & 0.5 \\ \hline
    \end{tabular}
\end{table}

There are other models that leverage the power of parallel computing from Megatron architecture, BioMegatron \cite{shin2020biomegatron} used Megatron to train a LLM based with biomedical domain specific and Megatron-CNTRL \cite{xu2020megatron} which proposed a framework of LLM that has controllability output by keyword prediction, knowledge retrieval, contextual knowledge rank and conditional text generation.

\subsubsection{LLaMA}
The LLaMA model was introduced by Meta in 2022 and made open-source. Its primary aim was to enhance model capabilities while maintaining a smaller size suitable for local deployment \cite{touvron2023llama}. LLaMA employed three distinct strategies. Firstly, it adopted the RMS Pre-Norm \cite{zhang2019root} in the transformer decoder, replacing the conventional layer normalization. This modification eliminated the re-centering operation, retaining only the re-scaling operation, thereby facilitating smoother gradient descent. Additionally, LLaMA utilized methods similar to PaLM, incorporating both SwiGLU \cite{shazeer2020glu} and RoPE positional embedding. The LLaMA model suite consists of four variants: 7B, 13B, 33B, and 65B parameters. Despite having significantly fewer parameters than GPT-3, Meta suggested that LLaMA could be locally deployed. However, the performance of LLaMA, with its reduced parameter count, did not significantly surpass that of GPT-3.

Following LLaMA, a series of derivative models emerged. Alpaca \cite{alpacastanform} sought to fine-tune the original LLaMA model using over 52,000 fine-tuning data samples extracted from the text-davinci-003 model \cite{openaimodels} developed by OpenAI for GPT-3. The resulting Alpaca model achieved performance comparable to text-davinci-003 but at a reduced cost. Subsequently, Guanaco \cite{guanaco} was introduced as Alpaca's successor, integrating block-wise k-bit quantization. This feature marked the LLaMA series' first foray into quantization methods, compressing the model by converting the original FP32 data format to a more compact int8. Additionally, Guanaco employed low-rank adapters (LoRA) to keep model parameters constant, only adjusting the optimizer with a smaller dataset batch. This approach further reduced the fine-tuning costs. Alpaca-LoRA \cite{alpacalora} combined Alpaca with the LoRA technology. Vicuna \cite{vicuna}, with its 13B parameters, represented a fine-tuned LLaMA model derived from dialogue datasets from ShareGPT \cite{sharegpt}, achieving approximately 90\% of ChatGPT's performance.

Distinct from earlier iterations, which primarily fine-tuned the original LLaMA, Dolly \cite{dolly} employed GPT-J-6B \cite{gpt-j} to emphasize the importance of fine-tuning over baseline models. Dolly v2 \cite{dollyv2} transitioned to using the databricks-dolly-15k dataset and replaced GPT-J-6B with Pythia \cite{biderman2023pythia}, making the model more accessible for business applications. Koala \cite{koala}, applied to FastChat \cite{fastchat}, adopted approaches similar to Vicuna, drawing from dialogue data for its fine-tuning. LLaMA 2 \cite{touvron2023llama}, a direct successor of LLaMA, presented three versions with 7B, 13B, and 70B parameters. It increased the original LLaMA training dataset by approximately 40\%. LLaMA 2 introduced the grouped-query attention (GQA) strategy, which grouped attention calculations between K and V values into sets of 8, thereby optimizing attention score computations. A notable distinction between LLaMA 2 and its predecessor was the doubling of the context length, coupled with enhanced data cleaning methods. Baize \cite{xu2023baize} offered a novel pipeline leveraging ChatGPT to autonomously generate new fine-tuning data.

Several LLaMA derivatives aimed to achieve multi-lingual capabilities. Chinese-Vicuna and Luotuo (also termed Chinese-alpaca-lora \cite{luotuo}) fine-tuned the LLaMA model using the LoRA approach to support Chinese, among other languages. Chinese Llama Alpaca 2 \cite{cllamaalpaca2} expanded upon this, incorporating datasets from various languages, including German and French.

Others sought to integrate multimodal support into LLaMA. LLaMA adapter \cite{zhang2023llama} introduced a pipeline adapting LLaMA for visual instruction-following, yielding a smaller and quicker fine-tuning model. Its successor, LLaMA adapter V2 \cite{gao2023llama}, preserved instructions from its predecessor while refining the image-text projection to enhance visual-text alignment. MiniGPT-4 \cite{zhu2023minigpt}, grounded in the Vicuna model, implemented a two-phase pre-training approach. After freezing both the language model and visual encoder, it constructed a new image-text dataset for fine-tuning MiniGPT-4, ensuring superior visual-text alignment. Both Chinese-LLaVA \cite{chinesellava} and LLaSM \cite{shu2023llasm} were derived from LLaMA 2 but incorporated multi-modal support. Visual-LLaMA \cite{visualllama} utilized a technique consistent with KOSMOS-1 and PaLM-E, merging visual and text tokens for training. Video-LLaMA \cite{zhang2023video} integrated both audio and visual information, using a two-layer Q-former \cite{zhang2023vision} for video embedding. This model underwent training on the Webvid-2M dataset and the image captioning dataset from LLaVA \cite{liu2023visual}, a visual-language model built on 150K multimodal data samples generated by GPT-4. Post pre-training, fine-tuning instructions from MiniGPT-4, LLaVA, and VideoChat \cite{li2023videochat} were applied. The audio data in Video-LLaMA was processed similarly, but with the inclusion of the ImageBind-Huge encoder \cite{girdhar2023imagebind} for audio information embedding.

\subsubsection{Gopher and DeepMind}
Gopher, introduced by DeepMind, is a large language model (LLM) with a training pipeline that encompasses six models, varying in parameter count from 44M to 280B \cite{rae2021scaling}. This model employs the RMS Pre-norm strategy \cite{zhang2019root} and incorporates relative positional embeddings using a 32,000 token SentencePiece tokenizer \cite{kudo2018sentencepiece}. For training and evaluation, the Gopher utilizes the JAX \cite{jax} and Haiku \cite{haiku} frameworks. Parallel computing is facilitated by JAX pump, and the model is trained using the MassiveText dataset.

Chinchilla, also a product of DeepMind and the successor to Gopher, posits that as the size of the model increases, the number of tokens trained should proportionally increase \cite{hoffmann2022training}. It suggests that the optimal Gopher model should be four times smaller when trained on a dataset that is four times larger. Various fine-tuning strategies based on Gopher were explored to ascertain the optimal ratio between model size and training data. These included altering the number of tokens per batch, maintaining consistent FLOPs, and training with a parameterized loss function. The latter was identified as the optimal approach.

DeepMind further developed a visual model named Flamingo, with 80B parameters, tailored for few-shot learning \cite{alayrac2022flamingo}. It employs the Perceiver resampler, training with a pre-established visual model known as NFNet \cite{brock2021high}. During the pre-training phase of the language model, NFNet remains static. Subsequently, the Perceiver resampler is integrated with the frozen language model and the visual model, yielding visual representations. The language model is then fine-tuned, leveraging these visual representations, which strengthens the nexus between the language and visual models, resulting in state-of-the-art performance.

\subsubsection{Other auto-regressive models}
Jurassic-1, introduced by AI21 Lab in collaboration with the AI21 Studio, was developed with the objective of offering an open conversational API \cite{ai21studio}. At the time of its release, the Jumbo edition of Jurassic-1, with its 178B parameters, was heralded as the most intricate and expansive model available \cite{lieber2021jurassic}. This model was trained on a corpus encompassing 250K labeled datasets. AI21 asserted that the dataset underpinning the Jurassic-1 model was quintuple the size of concurrent datasets. Succeeding Jurassic-1, Jurassic-X integrated the Modular Reasoning, Knowledge, and Language (MRKL) system—a composite of mixed expert data extraction from multiple databases. The outputs from these extractions are then processed by the language model, achieving a balance between universality and sparsity in large language models \cite{karpas2022mrkl}. In 2023, AI21 Lab unveiled the Jurassic-2 model \cite{j2model} with enhancements spanning multilingual capabilities, accuracy, and latency.

Anthropic launched the Claude model series, comprising two iterations: Claude \cite{claude} and Claude 2 \cite{claude2}. The foundational ethos of Claude echoes the principles laid out in the InstructGPT paper—namely, the creation of AI language models that are helpful, honest, and harmless \cite{ouyang2022training}. With this framework, Claude was trained on a dataset meticulously curated to align with these objectives, blending datasets that were both assistance-focused and harm-avoidant. This amalgamated dataset bolstered the performance model's propensity for helpfulness while eschewing potentially detrimental instructions. Claude 2, the subsequent model, showcased amplified performance metrics, supporting extended sequences, enhanced coding proficiency, and heightened security measures. Anthropic posited that Claude 2 boasts a security standard twice as robust as its predecessor, Claude 1.3.

Falcon, a pioneering model accessible to the public, stands as a worthy counterpart to several proprietary models \cite{mei2022falcon}. It comprises three variants: Falcon-7B, Falcon-40B, and Falcon-180B. Predominantly trained on the RefinedWeb dataset—a refined iteration of the CommonCrawl dataset \cite{penedo2023refinedweb, commoncrawl}—Falcon harnesses multi-query attention, mirroring the PaLM model. This approach substantially curtails the memory overhead of the K, V values during training, slashing memory usage by factors ranging from 10 to 100.

OpenAI, beyond the GPT series, has released several domain-specific models. DALL-E \cite{ramesh2021zero} is a model designed for image generation. It incorporates three distinct models: the dVAE \cite{vahdat2018dvae}, which encodes and compresses the input image; a BPE Encoder combined with a transformer for auto-regressive training; and CLIP \cite{radford2021learning} to measure the similarity between text-image pairs. DALL-E 2 \cite{ramesh2022hierarchical} employs CLIP to align the feature spaces of both text and image data. Subsequently, it introduces a module named "prior" that uses caption data to embed the data from the CLIP model, then decodes this data into an image. Notably, DALL-E 2 has approximately 3.5B parameters, substantially fewer than DALL-E's 12B parameters. DALL-E 3 \cite{dalle3}, released in 2023, is reportedly built on ChatGPT, though detailed technical information had not been disclosed at the time of this survey. Whisper \cite{radford2023robust} is designed for audio-to-text conversion and boasts 1550M parameters. The training of Whisper utilized over 680K audio samples from 98 languages, supporting multi-tasking in a supervised learning context. Each audio sample was segmented into 30-second chunks and processed into log-Mel-frequency cepstrum. Codex \cite{chen2021evaluating} targets coding challenges, comprising 12B parameters with \cite{li2023starcoder} which is also a model that is specific on coding tasks. It is trained using a dataset amalgamated from GPT data and open-source code from more than 5400 GitHub repositories, which, post-cleaning, amounted to 159G. Its efficacy was validated using the HumanEval dataset, which includes 164 manually generated programming problems to ensure the output's accuracy.

In addition to the PaLM series model developed using the Pathways architecture, Google has unveiled several domain-specific models. Meena targets end-to-end question answering and employs the evolved transformer architecture \cite{so2019evolved}. Its performance is assessed through human evaluation and perplexity metrics. LaMDA \cite{thoppilan2022lamda} employs knowledge-based queries, building on Meena to endow the model with the ability to answer industry-related knowledge questions. ALIGN \cite{cohen1997align} is designed to tackle representation learning for visual-textual data. It processes an extensive dataset containing over 1B noisy image captions, incorporating EfficientNet \cite{koonce2021efficientnet} for image-text matching, retrieval, and visual classification tasks. The objective of GaLM \cite{du2022glam} is to harness the sparsity of LLM for efficient few-shot learning. The model is trained on a colossal dataset featuring 1600B tokens, predominantly sourced from websites. GaLM utilizes the MoE architecture for training and houses approximately 1.2T parameters across 64 experts. Intriguingly, during inference, only 97B of these parameters are activated, enabling rapid response times. Lastly, Gemini \cite{team2023gemini} is a model that recently released by Google as their next-generation of large language model, which contains Gemini Ultra, Gemini Pro and Gemini Nano, Genmini.

In June 2023, Microsoft proposed Phi-1, a model with just 1.3 billion parameters. Its goal is to achieve high accuracy performance with a small model, highlighting the significance of high-quality data during the pre-training phase. They stated that they used exercises using 1B tokens and GPT-3.5 along with 6B tokens from online and artificially generated textbooks, all of which were of "textbook quality". Following the release of Phi-1, Microsoft also released Phi-1.5 \cite{phi15}, which was the same size as Phi-1, and Phi-2 \cite{phi2}, which was their most recent model and a foundation model with 2.7B of parameters to exhibit the potential capability on small language models (SLMs).

mPLUG is a series of models developed by Alibaba, tailored for multimodal support in Large Language Models (LLMs). The conceptual foundation of mPLUG \cite{li2022mplug} draws inspiration from the modularity of the human brain. In this architecture, each modality is treated as a separate module, allowing for specialized task handling. Conversely, mPLUG-2 \cite{xu2023mplug} adopts a unified model approach, managing all tasks within a single model, yet through different modules. mPLUG-Owl is a conversational LLM that has been open-sourced. It employs a Vision Transformer (ViT) to learn from visual-language captioning data. Subsequently, the model is fine-tuned using LoRA to align both uni-modal and multi-modal data, while preserving the foundational visual-text modules trained in the initial phase. There are other models like MM-REACT \cite{yang2023mm} and HuggingGPT \cite{shen2023hugginggpt} that serve to facilitate collaboration across various modalities. Additionally, models such as Youku-mPLUG \cite{xu2023youku}, mPLUG-DOCOWL \cite{ye2023mplug}, and MultiVENT \cite{sanders2023multivent} are derivatives of the original mPLUG design.

Numerous institutions, companies, and research affiliations have released auto-regressive models to push the boundaries of machine learning. AlexaTM \cite{soltan2022alexatm}, a multi-lingual model housing 20B parameters, has achieved state-of-the-art performance, especially for low-resource languages. PLATO \cite{bao2019plato} leverages discrete latent variables to encapsulate invisible background knowledge. Its successor, PLATO-2 \cite{bao2020plato}, refines the original by expanding both the training data and the number of parameters, incorporating a curriculum learning approach. The WuDao series boasts models like WuDao 2.0 \cite{wudao20}, one of the most extensive LLMs with 1.75T parameters, and WenLan \cite{huo2021wenlan}, a 5.3B parameter model tailored for Chinese-English visual captioning, rooted in a 650M image-captioning dataset and utilizing the Deep Structured Semantic Model (DSSM) \cite{huang2013learning} technology. Cogview \cite{ding2021cogview}, a 4T parameter Chinese multimodal LLM, and Lawformer \cite{xiao2021lawformer}, an early LLM dedicated to the legal domain, are noteworthy contributions. OPT \cite{zhang2022opt}, developed by Meta, is an endeavor towards open-sourcing LLMs, with models ranging from 120M to 175B parameters. Its enhanced version, OPT-IML \cite{iyer2022opt}, comprises two models with 30B and 175B parameters, respectively, and is fine-tuned using datasets spanning over 2000 languages. YaLM \cite{yalm100b}, a 100B LLM by Yandex, is trained on 1.7T text data sourced from websites, books, and other mediums. BLOOM \cite{scao2022bloom} is a comprehensive model with 176B parameters, trained on data from 46 natural languages and 13 programming languages, representing the collaborative efforts of over 1000 scholars. Lastly, Galactica \cite{taylor2022galactica}, developed in partnership between Meta and Papers with Code, mirrors the ambitions of GLaM in addressing scientific challenges. Mistral \cite{jiang2023mistral} claimed it uses mix-of-expert outperformed GPT-3.5 with smaller model size with the usage of mix-of-expert.

\subsection{Sequence to Sequence Models}

Another prominent architecture in the domain of large language models combines the features of both Auto-encoding and Auto-regressive models, known as the sequence-to-sequence model. Typically, it employs the complete Transformer framework, encompassing both encoder and decoder structures. This hybrid model amalgamates the strengths of its predecessors, inheriting the natural language understanding capabilities from the encoder and the generation competencies from the decoder. To integrate the functionalities of both encoder and decoder, cross-attention is implemented between their respective layers, facilitating the interplay between different sequences. Owing to their distinctive advantages over models that solely use an encoder or decoder, sequence-to-sequence models are predominantly chosen for conditional generation tasks like summarization and machine translation. Compared to the previous two architectures, which multiple models may share one single origin, sequence to sequence models are more independent, which means it has fewer models derived from previous predecessors but normally, they formed their own family by them own regarding different tasks hence cause less consistency compared with the auto-encoding models and auto-regressive models.

\subsubsection{BART}
BART \cite{lewis2019bart} represents a significant development in the Sequence-to-Sequence model lineage and stands as one of the earliest models harnessing the full Transformer architecture. It masterfully combines BERT's \cite{devlin2018bert} bi-directional encoder characteristics with the auto-regressive decoder features from GPT. In training BART for sequence generation tasks, the Google research team employed a reconstruction loss, defined as the cross-entropy between the conditionally generated output based on the input and the actual ground truth. They also enhanced the masking technique inherited from BERT. Several masking strategies were adopted:

\begin{table}
    \centering
    \footnotesize
    \caption{Training Strategies used in BART}
    \begin{tabular}{|p{0.85in}|p{2.25in}|}
    \hline
        Token Masking & Replace tokens with masks randomly \\ \hline
        Token Deletion & Delete tokens randomly \\ \hline
        Text infilling & Replace span with difference length with masks \\ \hline
        Sentence Permutation & Permute the order of several sentences \\ \hline
        Document Rotation & Rotate the order of the sequences inside a document with a randomly chosen token \\ \hline
    \end{tabular}
\end{table}

BART's training strategies can vary depending on the targeted downstream tasks. For sequence classification tasks, both the encoder and decoder process the sequence, and the final state of the output is utilized. Similarly, for token classification tasks, the decoder's final state serves as the definitive representation for each token, though these tokens are considered independent. In machine translation tasks, the encoder's embedding is supplanted with an additional encoder termed the "randomly initialized encoder," which can leverage a disjoint vocabulary.

In a step forward, mBART \cite{chipman2022mbart} harnesses corpora from multiple languages, bestowing BART with multi-linguistic capabilities. This version was fine-tuned from the foundational BART model.

\subsubsection{Based on T5}
T5 \cite{raffel2020exploring} introduced a universal framework in the realm of pre-trained models. One of its major contributions was presenting a clear guide for future researchers concerning parameter and architecture selection. The model underwent training on 750G of tokens, encompassing 11B parameters. In T5's approach, every NLP task was conceptualized as a text-to-text task, leading to the adoption of the Sequence-to-Sequence architecture. The model utilized three distinct masking techniques: fully-visible masks where attention scores are computed from all input and output tokens; causal masks, similar to GPT's, where scores are derived only from the current token and its antecedents, predicting the subsequent token based on previously generated content; and causal masks with prefix, where the input sequence employs a fully-visible mask, but for the output's predicted tokens, only causal masks are used.

The study also evaluated three architectures: the encoder-decoder, the encoder-only (as seen in GPT-2), and the Prefix LM. In the latter, while the encoder can perceive bi-directional information, the decoder is limited to previously generated tokens. Furthermore, the C4 dataset, crafted from the Common Crawl database \cite{commoncrawl}, emerged from this work. The dataset involved a refinement of about 750GB of web data, retaining lines that ended with standard punctuation, eliminating offensive words, omitting lines containing Javascript, discarding pages that resembled code, excluding lorem ipsum passages, and ensuring sentences that appeared more than three times were uniquely preserved.

\begin{enumerate}
    \item Only Kept lines end with normal punctuation 
    \item Deleted all bad words
    \item Removed lines with Javascript
    \item Deleted all pages like code
    \item Removed lorem ipsum
    \item Only kept once for same sentences appears more than three times
\end{enumerate}

\begin{figure}
    \centering
    \includegraphics[width=1\columnwidth]{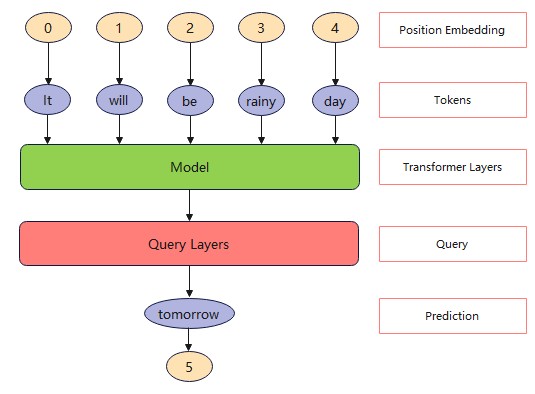} 
    \caption{The architecture of Pangu-$\Sigma$, the query layer proposed in that model was used after the embedding passed through the transformer layers, then the prediction of the next token wouold be produced from the query layer with position embedding
    \label{fig:pangusigma}}
\end{figure}

Building on T5's foundation, mT5 \cite{xue2020mt5} aimed at machine translation tasks within a multilingual context. It borrowed strategies from the original T5 model but incorporated more vocabulary to ensure a broader linguistic coverage. This was realized by drawing samples from languages with limited training data, using an approximation technique as suggested by \cite{zhang2022robust}. T0 \cite{wei2021finetuned}, evolving from T5, explored the enhancement possibilities of LLM focusing on prompt engineering for multitasking and robustness. Trained atop T5-LM, T0 incorporated training prompts from 177 datasets, amassing 2073 prompts in total. The performance indicated marked improvements in comparison to GPT-3. In a departure from FLAN \cite{zhang2022robust}, which also built upon T5, T0 retained the encoder-decoder architecture, whereas FLAN was solely anchored on T5's decoder.

\subsubsection{Pangu}

Pangu is a series of models comprising $Pangu-\alpha$ \cite{zeng2021pangu}, $Pangu-Coder$ \cite{christopoulou2022pangu}, and $Pangu-\Sigma$ \cite{ren2023pangu}. Unlike its precursor, $Pangu-\alpha$, the $Pangu-Coder$ was specifically designed as a decoder-only model tailored for generation tasks in both Chinese corpus and code generation. In contrast, $Pangu-\Sigma$ adopts an Encoder-Decoder architecture. Introduced in March 2023, the primary goal of $Pangu-\Sigma$ was to develop a large language model focusing on the Chinese corpus, while also being adept at multilingual tasks. Huawei announced that this model was the first China-centric LLM boasting a colossal 1T parameters size, approximately 1.085T, and underwent training with 2.17T data spanning over 300B tokens. The data sources for its training included 200G from WuDaoCopora 2.0 \cite{YUAN202165}, 100G from CLUECorpus 2020 \cite{xu2020cluecorpus2020}, 800G from the Pile \cite{gao2020pile}, and 750G from C4 \cite{raffel2020exploring}. Given its capability to execute code-related tasks, the $Pangu-\Sigma$ also incorporated 157G of Python code and 161G of Java code.

Two innovative technologies bolstered the model's efficiency. The first was a novel sparsity technique termed "Random Routed Experts" (RRE). Implemented in two phases, the initial step involved grouping the experts based on their parameters by the same task. Then, for every prompt, RRE deviated from the conventional MoE approach. Instead of allocating tokens to a specific expert with the most compatible matchup governed by a gate function, RRE would haphazardly assign an expert for a token. This gate-free procedure enabled researchers to derive sub-models from $Pangu-\Sigma$, facilitating their application to other downstream activities like translation and chatting. The second technological breakthrough was the "Expert Computation and Storage Separation". This mechanism strategically distributed model training across clusters, resulting in a remarkable 69,905 tokens-per-second I/O efficiency and a substantial reduction in communication overhead between servers and devices.

\begin{figure}
    \centering
    \includegraphics[width=1\columnwidth]{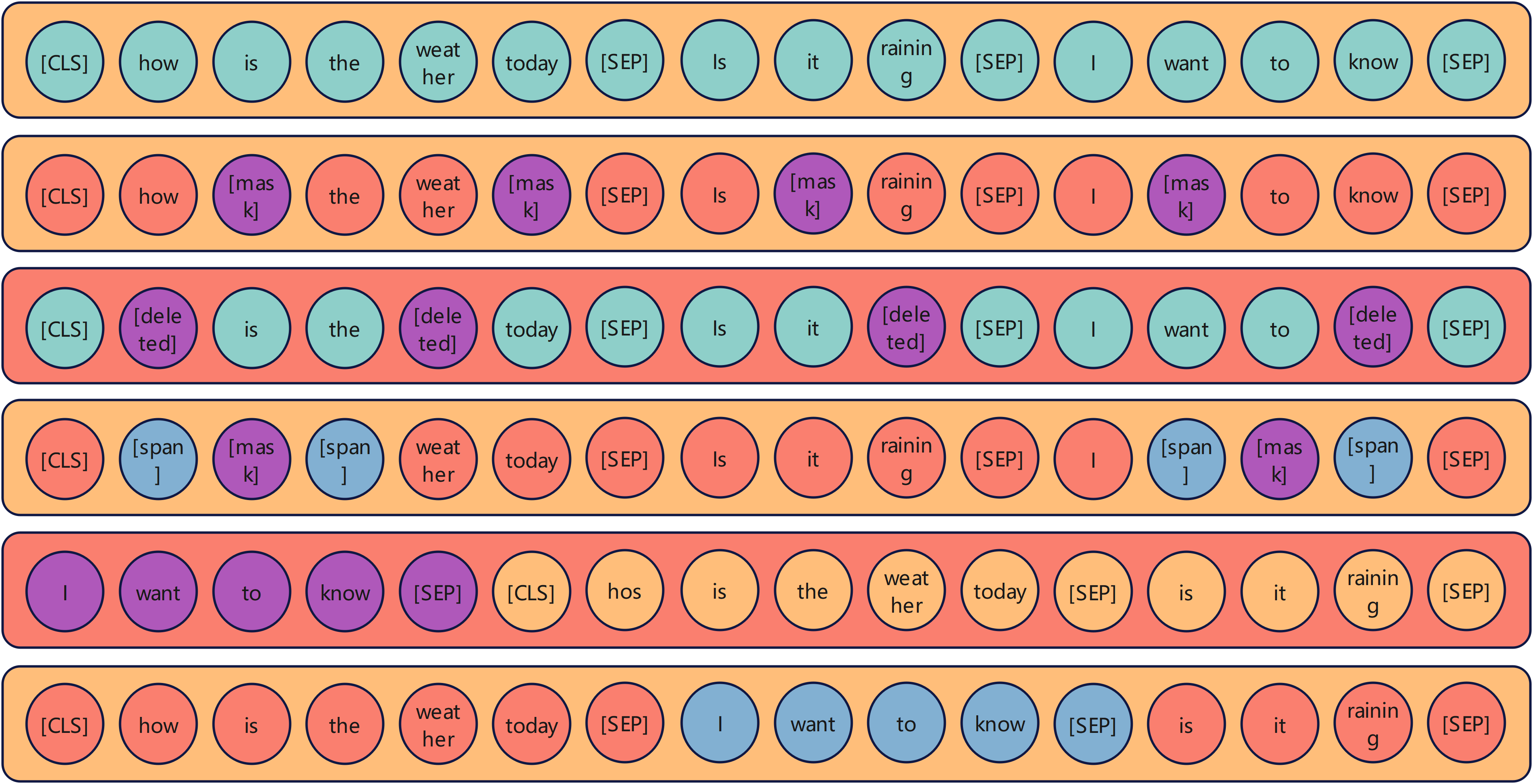} 
    \caption{Noise used by BART. 1. Original sequences, different sequence denoted by different colour 2. Token masking, in this case, the word "is, today, it, want" were masked 3. Token deletion, the word "how, weather, raining, know" were deleted 4. Token infilling, token "is, to" were masked and it was required to infill the token "how, the" and "want, know" spanned from the masked tokens 4. Document rotation, in this case the last sentence "I want to know" was rotated to the front of the document 5. Sentence Permutation: the order of the sentence in the document was changed, in this case, the last sentence "I want to know" was changed to the second sentence to find the contextual information between sentences
    \label{fig:bart}}
\end{figure}

\subsubsection{Switch Transformer}
Switch Transformer \cite{fedus2022switch} employs the sparsity inherent in LLMs to accelerate both training and inference. The term "switch" in its name refers to the shifting between experts. Thus, its primary technology is the Mixture of Experts (MoE). In the architecture of this model, there are several experts, each having distinct parameters that are regulated by a gating function. For any given input or prompt, only certain sections of the model are activated based on the input parameters, thereby reducing the computational complexity associated with each task. As an illustrative example from the paper, a particular task might comprise a sequence of tokens, with each token designated to different experts.

Another significant advancement introduced in the Switch Transformer is the concept of "Simplified Sparse Routing". Traditionally, a single token would be routed to multiple 'k' experts, and the optimal response would be selected as the output to ensure both relevance and accuracy. Google streamlined this process. They posited that even if k equals 1, meaning each token is processed by just a single expert, the model can maintain its accuracy, all the while boosting efficiency.

Lastly, the model introduces "Efficient Sparse Routing". Google provided a formula to compute the optimal number of experts needed. This calculation takes into account the number of tokens in each batch and the number of experts active at a given stage. The "capacity factor" introduced in this mechanism serves as a buffer, permitting a surplus of experts to counter potential token overflows during the routing process.

\subsubsection{GLM}

GLM \cite{du2021glm} introduced an innovative approach known as auto-regressive blank infilling to refine the masking and infilling techniques, setting it apart from models like BERT and T5. Instead of simply masking out tokens, GLM strategically removes continuous words from the input and then attempts to reconstruct them. A distinctive feature of GLM is that the prediction of masked regions can be permuted. Moreover, its positional embedding is designed in a 2-dimensional format. Impressively, with fewer parameters, GLM managed to outperform BERT on the SuperGLUE benchmark \cite{superglue}. By leveraging Pattern Exploiting Training \cite{schick2020exploiting}, GLM transformed various natural language understanding tasks into cloze-style problems, which could potentially have multiple correct answers.

Building upon the capabilities of GLM, ChatGLM \cite{githubChatGLM6BREADMEenmdMain} was developed to provide conversational support.

Further expanding its applications, a multimodal version of GLM was introduced as VisualGLM \cite{githubGitHubTHUDMVisualGLM6B}. This model supports visual conversations in both English and Chinese languages. Beyond the foundational architecture of GLM-6B, VisualGLM incorporates image data training based on the BLIP2-Qformer framework \cite{li2023blip}.

\section{Pre-training Methods for LLMs}

Pre-training is a critical phase in the development of Large Language Models (LLMs). This phase involves training the models on vast amounts of textual data to learn language patterns, structures, and contextual nuances. The effectiveness of pre-training significantly impacts the performance of LLMs on downstream tasks such as text generation, machine translation, summarization, and more. This section delves into various state-of-the-art pre-training methods for LLMs, categorized into different strategies such as training data reduction, neural architecture search, progressive learning, mixed precision training.

\subsection{Training Data Reduction}

Training data reduction techniques aim to minimize redundancy and improve the efficiency of the training process by selecting or augmenting the most relevant data.

\begin{itemize}
    \item \textbf{COPA} \cite{jiang2024copaefficientvisionlanguagepretraining}: Combining pre-training and adaptation strategies to enhance generalization.        
    \item \textbf{MixMAE} \cite{liu2023mixmaemixedmaskedautoencoder}: Data augmentation and masking strategies to create diverse and challenging training examples.
    \item \textbf{Deduplicate Text Datasets} \cite{lee-etal-2022-deduplicating}: This method involves removing duplicate entries from the training data to reduce redundancy and improve training efficiency.
    \item \textbf{TRIPS} \cite{jiang-etal-2022-trips}: Task-aware pre-training data selection to ensure that the training data is relevant to the specific tasks the model will perform.    
    \item \textbf{PatchDropout} \cite{liu2022patchdropouteconomizingvisiontransformers}: Randomly dropping patches of input data to reduce computational requirements.
    \item \textbf{TPS} \cite{wei2023jointtokenpruningsqueezing}: Token Pruning Strategy for efficient training by selectively pruning less important tokens.
\end{itemize}

\subsection{Neural Architecture Search}

Neural Architecture Search (NAS) involves automatically finding the best neural network architecture for a given task. These methods optimize the model design to achieve better performance.

\begin{itemize}
    \item \textbf{PreNAS} \cite{wang2023prenaspreferredoneshotlearning}: Informed architecture search based on pre-training results.    
    \item \textbf{PASHA} \cite{bohdal2023pashaefficienthponas}: Progressive architecture search that evolves hybrid architectures over multiple stages.
    \item \textbf{ZICO} \cite{li2023zicozeroshotnasinverse}: Zero-shot architecture search to identify optimal model structures without extensive training.
    \item \textbf{ElasticViT} \cite{tang2023elasticvitconflictawaresupernettraining}: Adaptive vision transformer architecture that adjusts computation based on input complexity.
    \item \textbf{RankNAS} \cite{hu-etal-2021-ranknas}: Ranking neural architectures based on performance metrics .
    
\end{itemize}

\subsection{Progressive Learning}

Progressive learning strategies involve training models in stages, gradually increasing the complexity and scale to improve performance and stability.

\begin{itemize}
    \item \textbf{LiGO} \cite{wang2023learninggrowpretrainedmodels}: Layerwise growth optimization to efficiently scale models.
    \item \textbf{Staged Training} \cite{pmlr-v162-shen22f}: Gradual increase in training complexity through multiple stages.
    \item \textbf{Knowledge Inheritance} \cite{qin-etal-2022-knowledge}: Transferring knowledge progressively across model versions.
    \item \textbf{CompoundGrow} \cite{gu-etal-2021-transformer}:  A strategy for progressively increasing model size during training.
    \item \textbf{stackingBERT}\cite{pmlr-v97-gong19a}: Stack-based training approach for incremental learning in BERT models.
        
\end{itemize}

\subsection{Mixed Precision Training}

Mixed precision training techniques aim to balance training speed and model precision by using different numerical precisions for different parts of the model.

\begin{itemize}
    \item \textbf{Mesa} \cite{pan2022mesamemorysavingtrainingframework}: Scheduling multi-epoch training with mixed precision adaptations.
    \item \textbf{GACT} \cite{liu2022gactactivationcompressedtraining}: Gradient accumulation with compression techniques to train large models efficiently.
    \item \textbf{blpa} \cite{chakrabarti2019backpropapproximateactivationsmemoryefficient}: Block-level precision adaptation for efficient training.
    \item \textbf{Mixture} \cite{micikevicius2018mixedprecisiontraining}: Employing mixed precision training to enhance speed while maintaining accuracy.
    
\end{itemize}

The pre-training phase is vital for developing robust and efficient LLMs. The methods described represent recent advancements in pre-training, improving model performance, efficiency, and applicability. These strategies highlight the dynamic nature of NLP research. As LLMs evolve, exploring and implementing novel pre-training methods will be crucial for further advancements.

\section{Challenges of LLMs\label{sec:discussion}}

Large language models are the result of the development of neural networks and the technologies that followed such as like deep learning. For the modern cutting edge models, challenges still exist in the following phases:
\begin{itemize}
    \item \textbf{Data Drawbacks}: Large language models requires a massive computational resource for the pre-training and fine-tuning.
    \item \textbf{Model Compression}: Large language models normally contains over billions of parameters, which cause memory intensive during both the training and deployment phase.
    \item \textbf{Distributed Computation}: Due to the increasing model size, some state-of-the-art large language models are trained on high performance clusters rather than local devices, which presents a challenge for distribution computation in the LLM field.
    \item \textbf{Multimodal Support}: Large language models can not only handle natural language processing tasks, but also dealing with data from different format, that casues the multimodality support challenge of LLMs.
\end{itemize}

Cutting edge methods like Chain of Thought, Reinforcement Learning with Human Feedback, Transformer, and Mix of Expert have been proposed to train a model on a massive scale. A cursory review of the technologies that underpin the LLM technique will be provided in this section.

\subsection{Data Drawbacks}


The use of massive datasets is one of the main characteristics of LLMs. These datasets are essential to the pre-training, fine-tuning, and evaluation processes of these models, as demonstrated by the experiment in \cite{hoffmann2022training}, which demonstrates the superior performance of a smaller model trained with more labelled data than a larger model trained with less labelled data. However, as LLMs scale up, data challenges persist. The main points of the LLM problems pertaining to the data will be summarized in this section.

\subsubsection{Quality of data}

When it comes to relevance, richness, and redundancy, the quality of the data used to train LLMs is just as crucial as the model itself. Poor data quality can provide inaccurate and unreliable knowledge to the model as it learns from the datasets. The following factors could be the cause of the taxonomy of the data quality in this survey: (1. \textit{Inaccurate Data}: The model will pick up problematic knowledge from these data, resulting to inaccurate or deceptive information in the model. (2. \textit{Outdated Data}, this is particularly problematic in domains like technology and public events that are changing quickly. As an illustration, the most recent version of GPT-4, GPT-4 Turbo \cite{openai2023}, used knowledge up to April 2023, but the original GPT-4 \cite{2303.08774} used knowledge dated back to September 2021. Various technologies, including machine learning \cite{bourtoule2021machine} and model editing \cite{mitchell2021fast, mitchell2022memory, reid2022learning}, could be employed to mitigate this drawback. (3. \textit{Redundancy of the data} describes the existence of redundant or overly similar information in the training dataset; if a dataset has multiple copies of the same content, this will lead to an overrepresentation of the model on the viewpoints, which will increase the model's biased understanding of particular topics. Certain measurements, like data deduplication \cite{he2010data}, or the common solution from a range of state-of-the-art models, like Llama \cite{touvron2023llama} and GPT \cite{2303.08774}, involve using data from many sources.

\subsubsection{Bias of the data}

The training data frequently involves biases in human languages or other forms of data input. These biases can span a wide range of topics, including gender, color, culture, religion, profession, and philosophy, along with preconceptions. As a result, concerns over justice and ethnicity may arise.The bias of language models was characterized by \cite{liang2021towards} from many perspectives with respect to the social impact. A few studies have tried to lessen the effects of language model bias. For example, \cite{bordia2019identifying} uses local edit to lessen gender bias, and \cite{liu2021mitigating} uses reinforced calibration to lessen political bias. \cite{nadeem2020stereoset} offered a novel way to measure the bias stereotype on these pretrianed models.


\begin{figure}
    \centering
    \includegraphics[width=1\columnwidth]{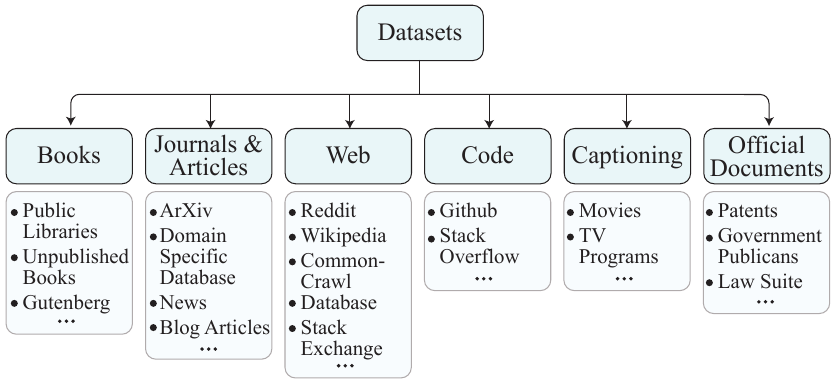}
    \caption{The common source of the dataset used for pre-training
    \label{fig:source}}
\end{figure}

\subsubsection{Scale of Data}
LLMs require massive amount of data to improve its accuracy and understanding of the prompts, which caused the challenge on the scale of the data regarding data collection, precessing and storage. The figure 15 shows the common source of the the datasets.

Following table shows the size of come popular datasets used by modern LLMs, single dataset such as The Pile and C4 already have hundreds of gigabytes and web-based database such as Common Crawl has already reached terabytes level.

\begin{table}[ht]
\centering
\footnotesize
\begin{tabular}{|l|l|l|}
\hline
\textbf{Dataset Name} & \textbf{Category} & \textbf{Data Size} \\ \hline
Common Crawl & Web & Terabytes level \\ \hline
RefinedWeb & Web & 5 trillion tokens \\ \hline
The Pile & Diverse & 800 GB \\ \hline
C4 & Web & 750 GB \\ \hline
Starcoder Data & Programming & 783 GB \\ \hline
BookCorpus & Books & 985 million words \\ \hline
ROOTS & Multilingual & 1.6 TB \\ \hline
Wikipedia & Encyclopedia & 19.88 GB \\ \hline
Red Pajama & Diverse & 1.2 trillion tokens \\ \hline
\end{tabular}
\caption{Open-Sourced Datasets for Training Large Language Models}
\label{tab:open_sourced_datasets}
\end{table}

Current cutting edge models normally applied datasets from different sources such as LLama \cite{touvron2023llama} used up to 4.5T of the datasets and \cite{ren2023pangu} applied 1.1T of the training data, which is also a huge challenge for the device used for the pre-training tasks.

\subsection{Model Compression}

In response to the space challenges posed by LLMs, this section provides a brief overview of three state-of-the-art model compression technologies. "Space challenges" in LLMs refer to the device memory limitations during pre-training, fine-tuning, and deployment due to the large size of model parameters. As LLMs grow larger to increase accuracy, more parameters are added, which intensifies computational demands. Model compression offers a solution by optimizing the internal structure of models to improve efficiency without significantly sacrificing performance. Three methods comprise the state-of-the-art model compression technique: (1) Pruning; (2) Quantization; and (3) Knowledge Distillation. These methods effectively address the efficiency and scalability challenges of LLMs by reducing model size and computational requirements.

\subsubsection{Pruning}
Pruning in large language models refers to the process of reducing model size by eliminating redundant and less critical structures. It can be categorized into two types: unstructured pruning and structured pruning. This optimization technique aims to shrink LLMs without significantly compromising their performance by selectively removing parameters considered less important for the model's task. Pruning methods eliminate redundant or non-informative parameters and can be performed in a structured or unstructured manner, each with its own strategies and impacts on model performance and efficiency. Typically, pruning involves a criterion to determine which weights to remove, based on factors like weight magnitude, training gradients, or other measures of importance. After pruning, the model usually undergoes fine-tuning to recover any lost performance due to parameter removal.

\textbf{Structured pruning} refers the pruning on the entire sets of the model structure such as channels, layers and weights. Structured pruning is advantageous for its compatibility with hardware optimization as it leads to a more regular and streamlined model structure with the trade-off of the substantial impact on the model's accuracy due to the removal of the entire structure. 

\textbf{Unstructured pruning} only removes certain individual weights or nodes from a neural network with the least importance inside the model connection, which leads to a more fine-grained pruning results with significant reduction on the model size. Different from the structured pruning which has the regular structure, unstructured pruning is hard to be implemented on the hardware due to its irregularity.

Table 7 shows transformer based pruning technology that can be applied on large langauge models

\begin{table}
    \centering
    \footnotesize
    \renewcommand{\arraystretch}{1.5}
    \caption{Pruning techniques used in Large language models categorized in structured, unstructured, and their combination.}
    \begin{tabular}{|m{0.6in}|m{2.4in}|}
    \hline
        Structured & Sanh \textit{et al.} \cite{sanh2020movement}, Cheong \textit{et al.} \cite{cheong2019transformers}, Gordon \textit{et al.} \cite{gordon2020compressing}, Wang \textit{et al.} \cite{wang2019structured}, Cui \textit{et al.} \cite{cui2021joint}, Yang \textit{et al.} \cite{yang2022textpruner}, Frantar \textit{et al.} \cite{frantar2023massive}, Zhang \textit{et al.} \cite{zhang2023pruning}, Chen \textit{et al.} \cite{chen2023lorashear}, Sun \textit{et al.} \cite{sun2023simple} \\ \hline
        Unstructured & Xia \textit{et al.}  \cite{xia2022structured}, Zhu \textit{et al.} \cite{zhu2021vision}, Voita \textit{et al.} \cite{voita2019analyzing}, Fan \textit{et al.} \cite{fan2019reducing}, Lagunas \textit{et al.} \cite{lagunas2021block}, Michel \textit{et al.}  \cite{michel2019sixteen}, Campos \textit{et al.}  \cite{campos2022sparse}, Santacroce \textit{et al.} \cite{santacroce2023matters}, Ma \textit{et al.} \cite{ma2023llm}, Guo \textit{et al.} \cite{guo2023compresso},  Xia \textit{et al.} \cite{xia2023sheared} \\ \hline
        Structured \& Unstructured  &  Mishra \textit{et al.} \cite{mishra2021accelerating}, Fang \textit{et al.} \cite{fang2022algorithm}, Holmes \textit{et al.}  \cite{holmes2021nxmtransformer}, Fang \textit{et al.} \cite{fang2022efficient}, Xu \textit{et al.} \cite{xu2022dense} \\ \hline
    \end{tabular}
\end{table}

\subsubsection{Quantization}

Quantization aims to decrease the memory footprint and computational demands of neural network models by converting high-precision model parameters, typically in extended data formats like 32-bit, into more compact representations, such as 8-bit formats, without significantly compromising performance. This technique is vital in model compression technologies and can be categorized into two main types: Post-Training Quantization (PTQ) \cite{zhang2023post} and Quantization Aware Training (QAT) \cite{jacob2018quantization}.

In the context of LLMs, quantization reduces computational resource requirements by transforming model parameters from high to low precision, typically converting 32-bit floating-point weights and activations to 8-bit integer format. This approach benefits both the storage footprint and computation speed.

Quantization employs a graded approach, ranging from 3-bit quantization for the most compact model size to 8-bit for nearly full precision. Each increase in bit-size generally improves the model's accuracy but also increases its size and computational demands. The most aggressive 3-bit quantization combines different techniques for various parts of the model, while higher bit-sizes use more refined methods, allocating more bits to parts sensitive to precision loss. At the high end, 8-bit quantization closely approaches the model's original floating-point precision, yielding high accuracy at the expense of size and speed. This spectrum of quantization strategies allows flexible deployment of LLMs like LLaMA 2 across different use cases, balancing resource constraints and accuracy needs. Table 6 summarizes these quantization methods, detailing the techniques, bit sizes, and model sizes.

\begin{table*}[ht]
\centering
\small
\begin{tabular}{|m{1cm}|l|l|m{13cm}|}
\hline
\textbf{Quant Method} & \textbf{Bits} & \textbf{Size} & \textbf{Description} \\
\hline
q3\_K & 3 & 2.95 GB & New k-quant method for all tensors, moderate size and RAM requirements. \\
\hline
q3\_K\_M & 3 & 3.28 GB & A variation of k-quant applying different bits for attention and feed-forward tensors. \\
\hline
q3\_K\_L & 3 & 3.60 GB & K-quant with higher bit allocation for select attention and feed-forward tensors. \\
\hline
q4\_0 & 4 & 3.79 GB & Original quant method with uniform 4-bit allocation across all tensors. \\
\hline
q5\_0 & 5 & 4.63 GB & Original 5-bit quant method for higher accuracy at the cost of increased resource usage. \\
\hline
q6\_K & 6 & 5.53 GB & New k-quant method using 6-bit quantization for all tensors, a balance between precision and size. \\
\hline
q8\_0 & 8 & 7.16 GB & 8-bit quantization offering high accuracy, suitable for scenarios where resource constraints are minimal. \\
\hline
\end{tabular}
\caption{Summary of different quantization methods for the LLaMA 2 model with a focus on k-quant strategies and resource implications.}
\label{table:quant_methods}
\end{table*}

\paragraph{Post-Training Quantization}
(PTQ) is a static quantization method applied after the model training process. It directly alters the original data format of the model without the need for additional data or modifications, aside from a few supplemental steps. In deep neural networks, the input typically adheres to a distinct pattern, facilitating statistical analysis. In PTQ, quantization algorithms convert the data format to a lower precision, guided by training data characteristics like the minimum and maximum weights and the distribution of activations. PTQ can be further subdivided into two methods: saturation and no saturation. The saturation approach employs KL divergence to identify a threshold \(T\), which then rescales the data range. In contrast, the no saturation method determines the maximum value of the model weights and then maps this value to a more confined data format range.
\paragraph{Quantization Aware Training}

\begin{table}
    \centering
    \footnotesize
    \renewcommand{\arraystretch}{1.5}
    \caption{Quantization algorithms using QAT and PTQ}
    \begin{tabular}{|m{0.3in}|m{2.7in}|}
    \hline
        QAT & LLM-QAT\cite{liu2023llm}, PEQA\cite{kim2023memory}, QLORA\cite{dettmers2023qlora} \\ \hline
        PTQ & GPTQ\cite{frantar2022gptq}, OPTQ\cite{frantar2022optq}, RPTQ\cite{yuan2023rptq}, FPTQ\cite{li2023fptq}, ZeroQuant\cite{yao2022zeroquant}, ZeroQuant-v2\cite{yao2023zeroquant}, ZeroQuant-FP\cite{wu2023zeroquant}, SmoothQuant\cite{xiao2023smoothquant}, OmniQuant\cite{shao2023omniquant}, OWQ\cite{lee2023owq}, AWQ\cite{lin2023awq}, LLM.int8()\cite{dettmers2022llm}, W4A4\cite{wu2023understanding}, ResQ\cite{abatiresq}, SqueezeLLM\cite{kim2023squeezellm}, QUIP\cite{chee2023quip}, SignRound\cite{cheng2023optimize}, Norm Tweaking\cite{li2023norm}, OLiVe\cite{guo2023olive}, QuantEase\cite{behdin2023quantease}, Outlier Suppresssion\cite{wei2022outlier},
        Outlier Suppresssion+\cite{wei2023outlier}, LUT-GEMM\cite{park2022lut}  \\ \hline
    \end{tabular}
\end{table}

In PTQ, quantization algorithms rely on statistical data from the model to determine the mapping, leading to a more significant discrepancy between the original and compressed models. On the other hand, Quantization Aware Training (QAT) adopts an online approach. Unlike the static methods in PTQ, QAT learns the scale and threshold during the training process by simulating the quantization effects. During QAT, a scaling ratio is established to map intermediate values. By allowing quantization to be back-propagated, this method results in a reduced quantization loss, making the quantized weights more akin to the original model's weights.

\subsubsection{Knowledge Distillation}

Knowledge distillation in large language models (LLMs) is a technique aimed at streamlining their vast knowledge into more compact and efficient forms. This process involves training a smaller, student model to emulate the behavior of a larger, teacher model, effectively transferring the sophisticated decision-making abilities of LLMs to smaller models. This makes the smaller models suitable for environments with limited computational resources, maintaining core functionalities while significantly reducing computational overhead.

Knowledge Distillation (KD) compresses the knowledge of a larger, more complex teacher model into a smaller, more efficient student model. This allows the student model to perform at a level close to the teacher but with a fraction of the computational requirements. Two main approaches, White-Box and Black-Box KD, are used in this process.

\paragraph{White-Box Knowledge Distillation}
This method involves using not only the outputs of the teacher model but also its internal representations and states to guide the training of the student model. This richer transfer of knowledge provides the student with insights into the intermediate processing of the teacher.

\paragraph{Black-Box Knowledge Distillation}
In contrast, Black-Box KD uses only the final outputs of the teacher model. The student learns to mimic the teacher's output distribution without access to its internal workings, making this method more flexible as it doesn’t require the student's architecture to match the teacher's internal structure.

The key challenge in both types of Knowledge Distillation (KD) is transferring as much relevant information as possible from the teacher to the student model. This often involves training the student to reproduce the teacher's output probabilities, which carry more information than just the final predicted class. The effectiveness of KD can be measured by how well the student model performs compared to the teacher on a set of tasks, ideally achieving similar performance while being more efficient to run.

Advancements in this domain have introduced innovative strategies to enhance the student model's learning process, such as selecting the most informative elements from the teacher model's knowledge and utilizing intermediate representations for a richer training experience. These techniques ensure that the distilled model replicates the critical aspects of the teacher model's performance. The impact of knowledge distillation extends beyond model efficiency, enabling the deployment of advanced language processing tools in diverse applications and promoting sustainable AI practices by reducing computational and energy demands.

\subsection{Distribution Computation}

Owing to the immense scale of large language models, which can reach up to trillions of parameters, traditional deep learning methods using single-device training or deployment are insufficient to handle the vast datasets and expansive parameter sizes associated with these models. As a result, distributed computation has emerged as a pivotal solution. Presently, three primary distributed computation methods are employed to address these challenges: data parallelism enhances the speed of model training, while tensor parallelism and pipeline parallelism enable the training of models that exceed the device's memory capacity.
\subsubsection{Tensor parallel}
The fundamental concept of tensor parallelism involves dividing the entire tensor of a model into distinct segments. Each device retains one segment of the tensor, and the final results can be procured by concatenating these tensor segments based on the dimensions from which they were partitioned. As an intra-layer parallelism method, its primary advantage is the ability to obtain results through a singular concatenation operation. However, a drawback of tensor parallelism is the necessity for an additional step to ensure the accuracy of the concatenation.
\subsubsection{Pipeline parallel}
Pipeline parallelism involves segmenting the entire model into multiple sections based on its layers, a method also referred to as inter-layer parallelism. In this approach, each device manages several layers of the model, and data is sequentially processed through the devices in accordance with the model's structure. Unlike tensor parallelism, pipeline parallelism doesn't require additional operations, as each device contains a complete segment of the model. However, a limitation is that devices must await data output from the preceding device, leading to idle periods. Consequently, pipeline parallelism can result in the underutilization of computational resources.
\subsubsection{Data parallel}
Unlike the previously mentioned approaches, which focus on accommodating models larger than what a single device can handle, the objective of data parallelism is to expedite the training process by harnessing computational power from multiple devices. In data parallelism, every device retains a copy of the model and is assigned a distinct data batch. This setup facilitates parallelized training, thereby enhancing the training speed. However, because each device has to store a complete replica of the model, a notable drawback of data parallelism is the inefficient use of device memory.

\subsection{Multimodal Support}

A significant challenge for contemporary Large Language Models (LLMs) is supporting multimodality, especially since the advent of the Vision Transformer (ViT) \cite{dosovitskiy2020transformers}, which showcased the potential of transformers for visual tasks. Unlike conventional LLMs, training models with multimodal support is more intricate due to the need for aligning representations across different modalities. This introduces distinct training tasks for these multimodal LLMs. This section is structured based on these tasks, which are categorized into pre-training tasks and downstream tasks.
\subsubsection{Image-text matching}
Image-Text Matching (ITM) is a method that aligns data from different modalities from a coarse-grained perspective. The primary objective of ITM is to predict the relationship between two segments, typically an image-captioning pair. This enables the model to learn the representation of text and its corresponding images. ITM has been extensively employed in state-of-the-art models. Examples include VILBERT \cite{lu2019vilbert}, B2T2 \cite{alberti2019fusion} — which employs bounding boxes to fuse image patches with textual information for enhanced visual-text integration — as well as LXMERT \cite{tan2019lxmert}, XLXMERT\cite{cho2020x}, VisualBERT \cite{li2019visualbert}, UNITER \cite{chen2020uniter}, Unicoder-VL \cite{li2020unicoder}, Pixel-BERT\cite{huang2020pixel}, ERNIE-VIL \cite{yu2021ernie}, ERNIE-VIL 2.0\cite{shan2022ernie}, and UNIMO \cite{li2020unimo, li2022unimo}. Furthermore, the BLIP series of models [paper, paper, paper] also incorporated ITM as one of their training tasks.
\subsubsection{Cross-modal contrastive learning}
In the Image-Text Matching (ITM) task, the model typically determines whether a visual-text pair's information aligns. Meanwhile, Cross-Modal Contrastive Learning (CMCL) seeks to enhance the association between image and text pairs based on their similarity. More specifically, CMCL operates like a clustering task, aiming to differentiate unrelated visual-text pairs and cluster closely related ones. The study [Leveraging Visual Knowledge in Language Tasks] explored the efficacy of the CMCL task. Several models employed this strategy, including UNIMO \cite{li2020unimo}, UNIMO2 \cite{li2020unimo}, WudaoMM \cite{yuan2022wudaomm}, Taisu \cite{liu2022taisu}, CLIP \cite{radford2021learning}, CLIP 2\cite{zeng2023clip2}, and ALIGN \cite{cohen1997align}.
\subsubsection{Cross-modal masked language matching}
Cross-modal masked language modeling (MLM) draws parallels to BERT by masking a portion of the input data, prompting the model to predict it during training. This straightforward approach is particularly effective for semantically dependent data. It's one of the earliest strategies introduced by ViT and has become popular in LLMs with multimodal support due to its adaptability. Numerous multimodal large language models, such as VisualBERT \cite{li2019visualbert}, ViLT \cite{kim2021vilt}, and InterBERT \cite{lin2020interbert}, incorporate MLM in their training processes.
\subsubsection{Masked region modeling}
Unlike Masked Language Modeling (MLM) which masks textual information, Masked Region Modeling (MRM) is designed for visual input masking tasks and can be categorized into two main approaches: classification and regression.

Masked Region Classification (MRC) has its origins in the Masked Region Prediction task (MRP). However, in MRC, masking is applied to an entire region of interest rather than lower-level tokens like patches or pixels. The objective is to predict the higher-level semantics of multimodal input by determining the classification of the masked region, using visible regions as context. Much like the ubiquity of MLM in text-based models, many multimodal large language models adopt MRC during training, notable examples being VL-BERT and Unicoder-VL. While the conventional MRC task utilizes cross-entropy, some models, such as UNIMO, have incorporated a variant known as MRC-kl, which employs KL-divergence as introduced by the UNITER framework.

Masked Region Feature Regression (MRSR), another offspring of MRP, employs regression techniques to reconstruct the feature map, rather than classifying masked regions. Models like ImageBERT \cite{qi2020imagebert} and UNITER have incorporated MRSR into their training paradigms.
\subsubsection{Other pre-training tasks}
Except the tasks mentioned above, there is other tasks used by either pre-training tasks and downstream tasks on LLMs, which is not as commonly used compared with the previous mentioned models but were used by some specific models
\begin{enumerate}
    \item WRA: Word Region Alignment
    \item Seq2Seq: Sequence to Sequence generation
    \item VQA: Visual Question Answering
    \item MRFR: Masked Region Feature Regression
    \item SGP: Scene Graph Prediction
    \item VLC: Vision-Language Contrastive Learning
    \item MTL: Multi Task Learning
    \item WPA: Word-Patch Alignment
    \item MVM (MFC): Masked Vision Modeling
    \item VLM: Vision-Language Matching
    \item ITC: Image-Text Contrastive
    \item ITG: Image-Text Generation
    \item PrefixLM: Prefix Language Modeling 
    \item MOC: Masked Object Classification
\end{enumerate}

\subsection{Prompt Engineering}

\begin{figure}
    \centering
    \includegraphics[width=1.0\columnwidth]{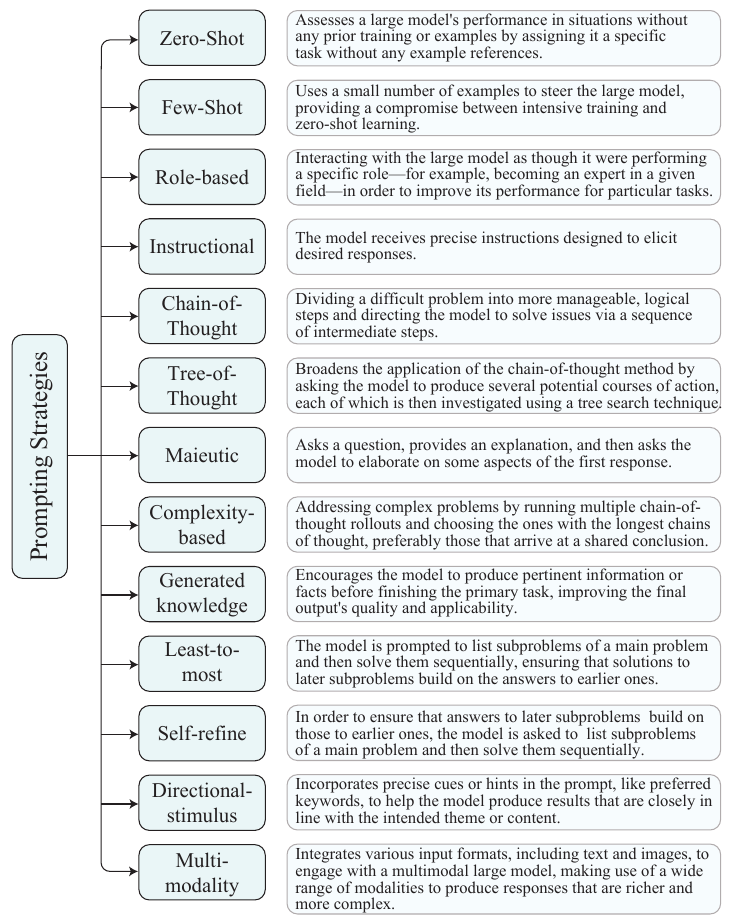}
    \caption{The most commonly utilized Prompt Engineering strategies. 
    \label{fig:prompt}}
\end{figure}

Another essential technique that speeds up the comprehension of LLMs in context is prompt engineering, which strategically formulates input queries that include content and instruction. This technique is simpler than pre-training and fine-tuning and allows users to interact with the LLM to control the token datastream.

\begin{figure}[t!]
    \centering    \includegraphics[width=0.95\columnwidth]{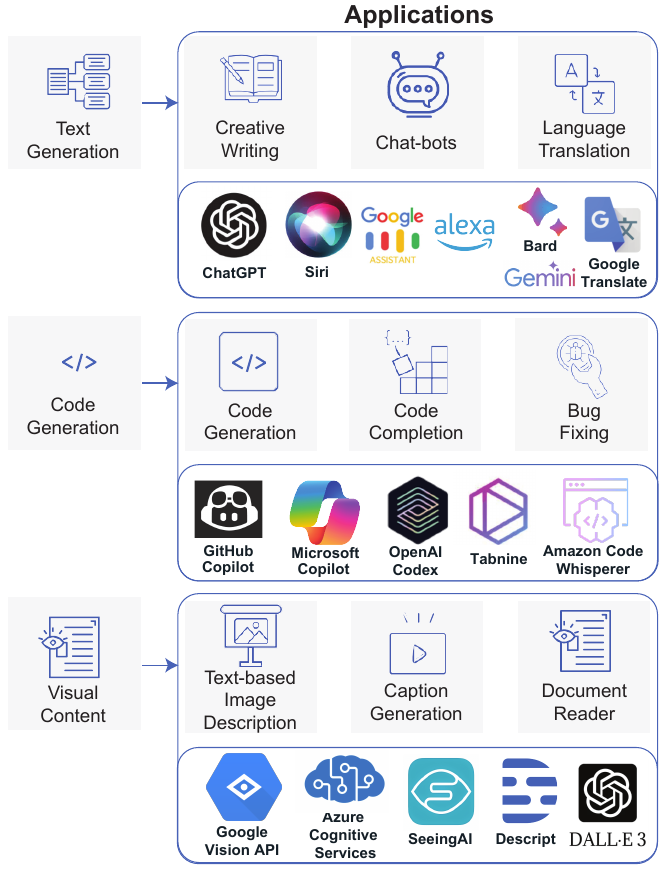}
    \caption{Applications of LLMs and MLLMs across various domains including text generation, code generation, and visual content. These applications showcase the versatility and impact of large language models in enhancing productivity and innovation.
    \label{fig:applications}}
\end{figure}

The following advantages arise from using prompt engineering during LLM inference. Prompt strategies reduce human bias in training data for LLMs by making it easier for users to find relevant results. This is primarily because of the interaction between the user and the system. Second, the model users' prompts have a high information density when compared to the training data used for pre-training and fine-tuning, suggesting that prompting is worth hundreds of data points on average (\cite{scao2021many}). Thirdly, prompt engineering can be customized so that users can achieve excellent accuracy performance even in the absence of training, particularly for downstream tasks. This feature offers prompt engineering unparalleled benefits in addition to training phases. Figure \ref{fig:prompt} is a taxonomy of some popular prompting techniques.



\section{Applications of LLMs}
LLMs have revolutionized various domains by leveraging their capabilities to understand, generate, and manipulate human language. Their applications span a wide array of fields, including visual content creation, audio generation, text generation, code generation, and design automation. While LLMs primarily handle text, they are often integrated with other systems to process and generate multimedia content. The versatility and efficacy of LLMs have led to significant advancements in these areas, with numerous companies developing notable products to harness their potential.

\subsection{Text Generation}

Text generation is one of the most prominent applications of LLMs. It encompasses several tasks such as creative writing, chat-bots, and language translation.

\begin{itemize}
    \item \textbf{Creative Writing}: LLMs based tools like OpenAI's GPT-4 and -4o, Bard, Gemini and other writing assistants help authors generate content, brainstorm ideas, and overcome writer's block.
    \item \textbf{Chat-bots}: Companies like Google, Apple, and Amazon utilize LLMs in their virtual assistants (Google Assistant, Siri, Alexa) to provide intelligent and conversational responses, improving user interaction.
    \item \textbf{Language Translation}: Tools like Google Translate and Microsoft's translation services leverage LLMs to provide accurate and context-aware translations across multiple languages.
\end{itemize}

\subsection{Code Generation}

LLMs have also made significant strides in the field of code generation, aiding developers in writing and optimizing code.

\begin{itemize}
    \item \textbf{Code Generation}: Products like GitHub Copilot and OpenAI Codex assist developers by generating code snippets and automating repetitive tasks, thus speeding up the development process.
    \item \textbf{Code Completion}: Tools such as Tabnine and Microsoft Copilot offer real-time code suggestions and completions, enhancing coding efficiency and reducing errors.
    \item \textbf{Bug Fixing}: Services like Amazon CodeWhisperer and other AI-powered code analysis tools help in identifying and fixing bugs, improving the overall quality of the software.
\end{itemize}

\subsection{Visual Content Understanding}

In addition to text and code, LLMs are instrumental in understanding visual content. This includes tasks like text-based image description, caption generation, and document reading through OCR (Optical Character Recognition).

\begin{itemize}
    \item \textbf{Text-based Image Description}: Tools like Google Vision API and Azure Cognitive Services use LLMs to provide detailed descriptions of images, enhancing accessibility for visually impaired users.
    \item \textbf{Caption Generation}: Applications like SeeingAI and DALL-E 3 generate captions for images, making content more accessible and searchable.
    \item \textbf{Document Reader}: Products such as Descript and other OCR technologies convert scanned documents and images into machine-readable text, facilitating easier access and analysis.
\end{itemize}

The applications of LLMs across various domains demonstrate their vast potential in enhancing productivity and fostering innovation. As these models continue to evolve, we can expect even more groundbreaking applications that will further transform the way we interact with and utilize technology.

\section{Conclusion}
This paper offers an exhaustive review of Large Language Models (LLMs) and their evolution within the domain of Natural Language Processing (NLP). It explores the diverse proficiencies of LLMs across various NLP tasks, including text generation, logical reasoning, machine translation, summarization, and multimodal integration. LLMs are systematically categorized into three primary architectures: encoder-only, decoder-only, and encoder-decoder frameworks. Additionally, the paper highlights the inherent challenges and constraints of LLMs, notably their dependency on statistical patterns as opposed to genuine understanding, and showcases state-of-the-art methodologies in pre-training and fine-tuning. The paper also discusses various model compression techniques, benchmarks, and applications. Overall, this paper offers valuable insights into the capabilities, challenges, and future prospects of LLMs in NLP applications.

\section*{Acknowledgement}
This work was partially supported by the NYUAD Center for Artificial Intelligence and Robotics (CAIR), funded by Tamkeen under the NYUAD Research Institute Award CG010, and the NYUAD Center for CyberSecurity (CCS), funded by Tamkeen under the NYUAD Research Institute Award G1104.

\bibliographystyle{IEEETran}
\bibliography{sample-base}

\newpage
\appendix
\onecolumn
\subsection{Model parameters}
As the evolution of LLMs, the number of parameters is also evolving from the BERT and GPT with only millions level of data to current trillion level of model parameters, the table below shows the models architecture and other metrics.
\LTcapwidth=\textwidth
\begin{table*}[ht]
\centering
\small
\caption{LLMs categorized into their Architectures, parameter count, and the Base LLM from which they are derived. }

\begin{tabular}{m{3cm}m{3cm}m{4cm}m{4cm}}
\label{parametertb} \\
\hline
\textbf{Model}  & \textbf{Architecture} & \textbf{Parameter} & \textbf{Base}  \\ \hline
    BERT & Auto-encoding & 110M, 340M & - \\ \hline
    SpanBERT & Auto-encoding & 110M, 340M & BERT \\ \hline
    RoBERTa & Auto-encoding & 125M, 355M & BERT \\ \hline
    DistilBERT & Auto-encoding & 66M & BERT \\ \hline
    BEiT & Auto-encoding & 86M, 307M & BERT \\ \hline
    Transformer-XL & Auto-encoding & 257M & - \\ \hline
    XLNet & Auto-encoding & 110M, 240M & Transformer-XL \\ \hline
    ERNIE & Auto-encoding & 110M & BERT \\ \hline
    ERNIE 2.0 & Auto-encoding & 110M & ERNIE \\ \hline
    ERNIE 3.0 & Auto-encoding & 260B & ERNIE 2.0 \\ \hline
    ALBERT & Auto-encoding & 125M & BERT \\ \hline
    ELECTRA & Auto-encoding & 14M, 110M, 335M & BERT \\ \hline
    DeBERTA & Auto-encoding & 100M, 350M, 700M & BERT \\ \hline
    BART & Sequence to Sequence & 140M, 400M & - \\ \hline
    T5 & Sequence to Sequence & 60M, 220M, 770M & - \\ \hline
    FLAN & Sequence to Sequence & 60M, 250M, 780M, 3B, 11B & T5 \\ \hline
    Pangu-alpha & Sequence to Sequence & 2.6B, 13B, 200B & - \\ \hline
    Pangu-sigma & Sequence to Sequence & 1.085T & Pangu-alpha \\ \hline
    GLM & Sequence to Sequence & 130B & - \\ \hline
    Minerva & Sequence to Sequence & 540B & - \\ \hline
    GPT & Auto-regressive & 117M & - \\ \hline
    GPT-2 & Auto-regressive & 1.5B & GPT \\ \hline
    GPT-3 & Auto-regressive & 175B & GPT-2 \\ \hline
    GPT-Neo & Auto-regressive & 125M, 1.3B, 2.7B & GPT-3 \\ \hline
    GPT-J & Auto-regressive & 6B & GPT-3 \\ \hline
    GPT-NeoX & Auto-regressive & 20B & GPT-Neo \\ \hline
    PaLM & Auto-regressive & 540B & - \\ \hline
    PaLM-E & Auto-regressive & 562B & PaLM, ViT \\ \hline
    PaLi & Auto-regressive & 3B, 15B, 17B & PaLM \\ \hline
    PaLM-2 & Auto-regressive & 340B & PaLM \\ \hline
    KOSMOS-1 & Auto-regressive & 1.6B & - \\ \hline
    Megatron LM & Auto-regressive & 1.2B, 2.5B, 4.2B, 8.3B & - \\ \hline
    Turing NLG & Auto-regressive & 17B & - \\ \hline
    Megatron-Turing NLG & Auto-regressive & 530B & Megatron LM, Turing NLG \\ \hline
    LLaMA & Auto-regressive & 7B, 13B, 33B, 65B & - \\ \hline
    Alpaca & Auto-regressive & 7B, 13B, 30B, 65B & LLaMA \\ \hline
    Guanaco & Auto-regressive & 7B, 13B, 30B, 65B & LLaMA \\ \hline
    Vicuna & Auto-regressive & 7B, 13B, 30B, 65B & LLaMA \\ \hline
    Dolly & Auto-regressive & 6B & LLaMA \\ \hline
    Dolly v2 & Auto-regressive & 12B & Dolly \\ \hline
    Pythia & Auto-regressive & 70M, 160M, 410M, 1B, 1.4B & - \\ \hline
    FastChat & Auto-regressive & 3B & - \\ \hline
    LLaMA 2 & Auto-regressive & 7B, 13B, 34B, 70B & LLaMA \\ \hline
    Baize & Auto-regressive & 7B, 13B & LLaMA \\ \hline
    LLaVA & Auto-regressive & 13B & - \\ \hline
    Gopher & Auto-regressive & 44M, 117M, 417M, 1.4B, 7.1B & - \\ \hline
    Chinchilla & Auto-regressive & 70B, 280B & Gopher \\ \hline
    Flamingo & Auto-regressive & 80B & Chinchilla \\ \hline
    Jurassic-1 & Auto-regressive & 7.5B, 178B & - \\ \hline
    Claude & Auto-regressive & 52B & - \\ \hline
    Claude 2 & Auto-regressive & 130B & Claude \\ \hline
    Falcon & Auto-regressive & 40B, 180B & - \\ \hline
    
\end{tabular}
\end{table*}

\LTcapwidth=\textwidth
\begin{table*}[t!]
\centering
\small
\begin{tabular}{m{3cm}m{3cm}m{4cm}m{4cm}}
    DALL-E & Auto-regressive & 12B & GPT-3 \\ \hline
    DALLE-E 2 & Auto-regressive & 3.5B & CLIP \\ \hline
    Whisper & Auto-regressive & 74M, 244M, 769M, 1550M & - \\ \hline
    Codex & Auto-regressive & 12B & - \\ \hline
    LaMDA & Auto-regressive & 137B & - \\ \hline
    GaLM & Auto-regressive & 1.2T & - \\ \hline
    mPLUG & Auto-regressive & 14M & - \\ \hline
    mPLUG-Owl & Auto-regressive & 7B & mPLUG, ViT \\ \hline
    AlexaTM & Auto-regressive & 20B & - \\ \hline
    PLATO-2 & Auto-regressive & 1.6B & PLATO \\ \hline
    PLATO-XL\cite{bao2021plato} & Auto-regressive & 11B & PLATO-2 \\ \hline
    OPT & Auto-regressive & 175B & - \\ \hline
    YaLM & Auto-regressive & 100B & - \\ \hline
    BLOOM & Auto-regressive & 176B & - \\ \hline
    Galactica & Auto-regressive & 120B & - \\ \hline
    VILBERT & Auto-regressive & 3B & BERT \\ \hline
    UNITER\cite{chen2020uniter} & Auto-regressive & 303M & - \\ \hline
    Unicoder-VL & Auto-regressive & 195M & BERT \\ \hline
    ERNIE-VILG & Auto-regressive & 10B & ERNIE \\ \hline
    ERNIE-VIL 2.0 & Auto-regressive & 24B & ERNIE-VILG \\ \hline
    CLIP & Auto-regressive & 63M & - \\ \hline
    ViLT & Auto-regressive & 87M & - \\ \hline
    BloombergGPT\cite{wu2023bloomberggpt} & Auto-regressive & 50B & - \\ \hline
    CTRL\cite{keskar2019ctrl} & Auto-regressive & 1.6B & - \\ \hline

\end{tabular}
\end{table*}


\subsection{Multimodal Support}
The table below is the multimodal support for each of the multimodal LLMs mentioned in this survey, the multimodality was classified into the following categories: Text, Image, Video, Audio, Embodied


\begin{table*}[ht]
    \centering
    \small
    \caption{Multimodal support for MLLMs}

    \setlength{\tabcolsep}{0.4mm}{
    \begin{tabular}{lccccc|lccccc}
        \hline
        \textbf{Model} & \textbf{Text} & \textbf{Image} & \textbf{Video} & \textbf{Audio} & \textbf{Embodied} & \textbf{Model} & \textbf{Text} & \textbf{Image} & \textbf{Video} & \textbf{Audio} & \textbf{Embodied} \\
        \hline
        VisualBERT & $\checkmark$ & $\checkmark$ & - & - & - & mPLUG-owl & $\checkmark$ & $\checkmark$ & $\checkmark$ & - & - \\
        \hline
        BEiT & $\checkmark$ & $\checkmark$ & - & - & - & mPLUG-DOCOWL & $\checkmark$ & $\checkmark$ & - & - & - \\
        \hline
        BEiT v2 & $\checkmark$ & $\checkmark$ & - & - & - & WenLan & $\checkmark$ & $\checkmark$ & - & - & - \\
        \hline
        BEiT v3 & $\checkmark$ & $\checkmark$ & - & - & - & VILBERT & $\checkmark$ & $\checkmark$ & - & - & - \\
        \hline
        ERNIE-VilG & $\checkmark$ & $\checkmark$ & - & - & - & B2T2 & $\checkmark$ & $\checkmark$ & - & - & - \\
        \hline
        ERNIE-Vil 2.0 & $\checkmark$ & $\checkmark$ & - & - & - & LXMERT & $\checkmark$ & $\checkmark$ & - & - & - \\
        \hline
        VisualGLM & $\checkmark$ & $\checkmark$ & - & - & - & XLXMERT & $\checkmark$ & $\checkmark$ & - & - & - \\
        \hline
        GPT-4 & $\checkmark$ & $\checkmark$ & - & - & - & UNITER & $\checkmark$ & $\checkmark$ & - & - & - \\
        \hline
        PaLM-E & $\checkmark$ & $\checkmark$ & $\checkmark$ & $\checkmark$ & $\checkmark$ & Unicoder-VL & $\checkmark$ & $\checkmark$ & - & - & - \\
        \hline
        KOSMOS-1 & $\checkmark$ & $\checkmark$ & $\checkmark$ & $\checkmark$ & - & Pixel-BERT & $\checkmark$ & $\checkmark$ & - & - & - \\
        \hline
        PaLi & $\checkmark$ & $\checkmark$ & - & - & - & UNIMO & $\checkmark$ & $\checkmark$ & - & - & - \\
        \hline
        LLaMA adapter & $\checkmark$ & $\checkmark$ & - & - & - & UNIMO 2 & $\checkmark$ & $\checkmark$ & - & - & - \\
        \hline
        LLaMA adapter v2 & $\checkmark$ & $\checkmark$ & - & - & - & BLIP & $\checkmark$ & $\checkmark$ & - & - & - \\
        \hline
        MiniGPT-4 & $\checkmark$ & $\checkmark$ & - & - & - & BLIP 2 & $\checkmark$ & $\checkmark$ & - & - & - \\
        \hline
        LLaSM & $\checkmark$ & $\checkmark$ & $\checkmark$ & $\checkmark$ & - & BLIP 3 & $\checkmark$ & $\checkmark$ & - & - & - \\
        \hline
        Video-LLaMA & $\checkmark$ & - & $\checkmark$ & - & - & WudaoMM & $\checkmark$ & $\checkmark$ & - & - & - \\
        \hline
        LLaVA & $\checkmark$ & $\checkmark$ & - & - & - & CLIP2 & $\checkmark$ & $\checkmark$ & - & - & - \\
        \hline
        VideoChat & $\checkmark$ & - & $\checkmark$ & - & - & ViLT & $\checkmark$ & $\checkmark$ & - & - & - \\
        \hline
        Flamingo & $\checkmark$ & $\checkmark$ & $\checkmark$ & - & - & InterBERT & $\checkmark$ & $\checkmark$ & - & - & - \\
        \hline
        DALL-E & $\checkmark$ & $\checkmark$ & - & - & - & ImageBERT & $\checkmark$ & $\checkmark$ & - & - & - \\
        \hline
        DALL-E 2 & $\checkmark$ & $\checkmark$ & - & - & - & Med-PaLM & $\checkmark$ & $\checkmark$ & - & $\checkmark$ & - \\
        \hline
        CLIP & $\checkmark$ & $\checkmark$ & - & - & - & Med-PaLM 2 & $\checkmark$ & $\checkmark$ & - & $\checkmark$ & - \\
        \hline
        Whisper & $\checkmark$ & - & - & $\checkmark$ & - & OSCAR & $\checkmark$ & $\checkmark$ & - & $\checkmark$ & - \\
        \hline
        ALIGN & $\checkmark$ & $\checkmark$ & - & - & - & Virtex & $\checkmark$ & $\checkmark$ & - & - & - \\
        \hline
        mPLUG & $\checkmark$ & $\checkmark$ & $\checkmark$ & - & - & VILLA & $\checkmark$ & $\checkmark$ & - & $\checkmark$ & - \\
        \hline
        mPLUG 2 & $\checkmark$ & $\checkmark$ & $\checkmark$ & - & - & BARD & $\checkmark$ & $\checkmark$ & - & $\checkmark$ & - \\
        \hline
        SLIP & $\checkmark$ & $\checkmark$ & - & - & - & FLIP & $\checkmark$ & $\checkmark$ & - & - & - \\
        \hline
    \end{tabular}}
    \label{table:multimodal_support}
\end{table*}

\onecolumn
\newpage
\subsection{Training approaches of Multimodal Large Language Models}
The table below outlines the commonly used training approaches for multimodal large language models, highlighting techniques for several models discussed in this survey paper. It compares the data inputs, integration algorithms, and training objectives across different models.
\begin{table*}[!htbp]
    \centering
    \footnotesize
     \setlength{\tabcolsep}{0.4mm}{
    \caption{ Multimodal training approaches for MLLMs }
    \begin{tabular}[!htp]{lccccccccccccccccccccc}
    \hline
\textbf{Model}  & \textbf{LM} & \textbf{MLM} & \textbf{ITM}  & \textbf{MRC} & \textbf{MRM} &\textbf{WRA} & \textbf{Seq2Seq} & \textbf{CMCL} & \textbf{VQA} & \textbf{MRFR} & \textbf{SGP} & \textbf{VLC} & \textbf{MTL} & \textbf{WPA} & \textbf{MVM} & \textbf{VLM} & \textbf{MSM} & \textbf{ITC} & \textbf{ITG} & \textbf{PrefixLM} & \textbf{MOC} \\ \hline
VisualBERT & - & $\checkmark$ & $\checkmark$ & - & - & - & - & - & - & - & - & - & - & - & - & - & - & - & - & - & - \\
\hline
BEiT & - & $\checkmark$ & - & - & - & - & - & - & - & - & - & - & - & - & - & - & - & - & - & - & - \\
\hline
BEiT v2 & - & $\checkmark$ & - & - & - & - & - & - & - & - & - & - & - & - & - & - & - & - & - & - & - \\
\hline
BEiT v3 & - & $\checkmark$ & - & - & - & - & - & - & - & - & - & - & - & $\checkmark$ & - & - & - & - & - & - & - \\
\hline
ERNIE-VilG & - & $\checkmark$ & $\checkmark$ & - & - & - & - & - & $\checkmark$ & $\checkmark$ & - & - & - & - & - & - & - & - & - & - & - \\
\hline
ERNIE-Vil 2.0 & - & - & - & - & - & - & - & - & - & - & - & - & $\checkmark$ & - & - & - & - & - & - & - & - \\
\hline
KOSMOS-1 & $\checkmark$ & - & - & - & - & - & - & - & - & - & - & - & - & - & - & - & - & - & - & - & - \\
\hline
MiniGPT-4 & - & - & $\checkmark$ & - & - & - & - & - & - & - & - & - & - & - & - & - & - & $\checkmark$ & $\checkmark$ & - & - \\
\hline
LLaVA & - & - & - & - & - & - & - & - & - & - & - & $\checkmark$ & - & - & - & - & - & - & - & - & - \\
\hline
Flamingo & - & - & - & - & - & - & - & - & - & - & - & - & $\checkmark$ & - & - & - & - & - & - & - & - \\
\hline
CLIP & - & - & - & - & - & - & - & - & - & - & - & $\checkmark$ & - & - & - & - & - & - & - & - & - \\
\hline
Whisper & - & - & - & - & - & - & - & - & - & - & - & - & $\checkmark$ & - & - & - & - & - & - & - & - \\
\hline
ALIGN & - & - & $\checkmark$ & - & - & - & - & - & - & - & - & - & - & - & - & - & - & - & - & - & - \\
\hline
mPLUG & - & $\checkmark$ & $\checkmark$ & - & - & - & - & - & - & - & - & - & - & - & - & - & $\checkmark$ & - & $\checkmark$ & - \\
\hline
mPLUG 2 & - & $\checkmark$ & - & - & - & - & - & - & - & - & $\checkmark$ & - & - & - & $\checkmark$ & - & - & - & - & - \\
\hline
mPLUG-owl & - & - & - & - & - & - & - & - & - & - & - & - & - & - & - & - & - & - & - & $\checkmark$ & - \\
\hline
VILBERT & - & $\checkmark$ & $\checkmark$ & $\checkmark$ & - & - & - & - & - & - & - & - & - & - & - & - & - & - & - & - & - \\
\hline
B2T2 & - & $\checkmark$ & $\checkmark$ & - & - & - & - & - & - & - & - & - & - & - & - & - & - & - & - & - & - \\
\hline
LXMERT & - & $\checkmark$ & $\checkmark$ & $\checkmark$ & $\checkmark$ & - & - & $\checkmark$ & - & - & - & - & - & - & - & - & - & - & - & - & - \\
\hline
UNITER & - & $\checkmark$ & $\checkmark$ & $\checkmark$ & $\checkmark$ & $\checkmark$ & - & - & $\checkmark$ & - & - & - & - & - & - & - & - & - & - & - & - \\
\hline
Unicoder-VL & - & $\checkmark$ & $\checkmark$ & $\checkmark$ & $\checkmark$ & - & - & - & - & - & - & - & - & - & - & - & - & - & - & - & - \\
\hline
Pixel-BERT & - & $\checkmark$ & $\checkmark$ & - & - & - & - & - & - & - & - & - & - & - & - & - & - & - & - & - & - \\
\hline
UNIMO\cite{li2020unimo} & - & $\checkmark$ & $\checkmark$ & - & - & - & $\checkmark$ & - & - & - & - & - & - & - & - & - & - & - & - & - & - \\
\hline
UNIMO 2\cite{li2022unimo} & - & $\checkmark$ & $\checkmark$ & - & - & - & - & - & - & - & $\checkmark$ & - & - & - & - & - & - & - & - & - & - \\
\hline
BLIP & $\checkmark$ & - & - & - & - & - & - & - & - & - & - & - & - & - & - & - & - & $\checkmark$ & - & - & - \\
\hline
BLIP 2 & - & - & $\checkmark$ & - & - & - & - & - & - & - & - & - & - & - & - & - & - & $\checkmark$ & $\checkmark$ & - & - \\
\hline
CLIP$^2$ & - & - & - & - & - & - & - & - & - & - & $\checkmark$ & - & - & - & - & - & - & - & - & - & - \\
\hline
ViLT & - & $\checkmark$ & $\checkmark$ & - & - & - & - & - & - & - & - & - & - & - & - & - & - & - & - & - & - \\
\hline
InterBERT & - & - & $\checkmark$ & - & $\checkmark$ & - & - & - & - & - & - & - & - & - & - & $\checkmark$ & - & - & - & - & - \\
\hline
ImageBERT & - & $\checkmark$ & - & - & - & - & - & - & $\checkmark$ & - & - & - & - & - & - & - & - & - & - & - & $\checkmark$ \\
\hline
OSCAR & - & - & - & - & - & - & - & - & - & - & $\checkmark$ & $\checkmark$ & - & - & - & - & - & - & - & - & - \\
\hline
Virtex & - & - & - & - & - & - & - & - & - & - & - & - & - & - & - & - & - & $\checkmark$ & - & - & - \\
\hline
FLIP & - & $\checkmark$ & - & - & $\checkmark$ & - & - & - & - & - & $\checkmark$ & - & - & - & - & - & - & - & - & - & - \\
\hline
COCA\cite{yu2022coca} & - & $\checkmark$ & - & - & - & - & - & - & - & - & $\checkmark$ & - & - & - & - & - & - & - & - & - & - \\
\hline
LSeg & - & - & - & - & - & - & - & - & - & - & $\checkmark$ & - & - & - & - & - & - & - & - & - & - \\
\hline
VL-BERT & - & $\checkmark$ & - & $\checkmark$ & $\checkmark$ & - & - & - & - & - & - & - & - & - & - & - & - & - & - & - & - \\
\hline
VideoBERT & - & $\checkmark$ & - & - & - & - & - & - & - & - & - & - & - & - & - & - & - & - & - & - & - \\
\hline
ALBEF\cite{li2021align} & - & $\checkmark$ & - & - & - & - & - & - & - & - & $\checkmark$ & - & - & - & $\checkmark$ & - & - & - & - & - & - \\
\hline
SimVLM\cite{wang2021simvlm} & - & $\checkmark$ & - & - & - & - & - & - & - & - & - & - & - & - & - & - & - & - & - & - & - \\
\hline
FILIP & - & - & - & - & - & - & - & - & - & - & $\checkmark$ & - & - & - & - & - & - & - & - & - & - \\
\hline
VLMo\cite{bao2022vlmo} & - & $\checkmark$ & - & - & - & - & - & - & - & - & $\checkmark$ & - & - & - & $\checkmark$ & - & - & - & - & - & - \\
\hline
SOHO\cite{huang2021seeing} & - & $\checkmark$ & - & - & - & - & - & - & - & - & - & - & - & $\checkmark$ & $\checkmark$ & - & - & - & - & - & - \\
\hline
MAP\cite{lin2023map} & - & $\checkmark$ & - & - & - & - & - & - & - & - & $\checkmark$ & - & - & - & $\checkmark$ & - & - & - & - & - & - \\
\hline
    \end{tabular}}
\end{table*}

\EOD
\end{document}